\documentclass[10pt, letterpaper]{article}

\usepackage[margin=1in]{geometry}

\usepackage[numbers]{natbib} 
\usepackage{amsmath,amssymb} 
\usepackage{graphicx} 
\usepackage{caption,subcaption}
\usepackage{algorithm,algorithmic}
\usepackage{listings,newfloat}
\usepackage{color,xcolor,textcomp} 
\usepackage{mathtools} 
\usepackage{arydshln} 
\usepackage{BOONDOX-uprscr}
\usepackage{times,helvet,courier}
\usepackage{relsize}    
\usepackage{amsopn}    
\usepackage{soul}
\usepackage{booktabs}

\usepackage{amsopn}

\DeclareMathOperator*{\argmax}{arg\,max}
\DeclareMathOperator*{\argmin}{arg\,min}

\newcommand{\sa}{\mathsf a}

\renewcommand{\vec}[1]{\mathbf{#1}}

\usepackage{algorithm}
\usepackage{textcomp}
\usepackage{xcolor}
\usepackage{graphics}
\usepackage{caption}
\usepackage{subcaption}
\usepackage{arydshln}
\usepackage{mathtools}
\usepackage{relsize}
\usepackage{BOONDOX-uprscr}
\usepackage{dirtytalk}
\DeclareMathAlphabet\mathbfcal{OMS}{cmsy}{b}{n}

\usepackage{color}
\definecolor{darkgreen}{rgb}{0,0.5,0}

\newcommand{\edit}[1]{\textcolor{black}{#1}}
\usepackage{amsmath}
\usepackage{amssymb}

\title{Dynamic Risk Assessment for Autonomous Vehicles from Spatio-Temporal Probabilistic Occupancy Heatmaps}

\author{
Han Wang$^{1,a}$, Yuneil Yeo$^{1,a}$, Antonio R. Paiva$^{2}$,\\
Jack P. Goodman$^{3}$, Jean Utke$^{2}$, and Maria Laura Delle Monache$^{1}$\vspace{0.5em}\\
\small $^{1}$Department of Civil and Environmental Engineering, University of California, Berkeley, CA, USA\\
\small $^{2}$D3 Analytics Innovation, Allstate Insurance Company, Northbrook, IL, USA\\
\small $^{3}$Department of Electrical Engineering and Computer Sciences, University of California, Berkeley, CA, USA\\
\small {\tt hanw@berkeley.edu, yuneily@berkeley.edu, mldellemonache@berkeley.edu,}\\
\small {\tt antonio.paiva@allstate.com, jutke@allstate.com, jgoodman4@berkeley.edu}\\[0.5em]
\small $^{a}$These authors contributed equally.\\
\small Corresponding author: Yuneil Yeo ({\tt yuneily@berkeley.edu})
}
\date{}

\begin{document}
\maketitle
\thispagestyle{empty}
\pagestyle{empty}

\begin{abstract}
Accurately assessing collision risk in dynamic traffic scenarios is a crucial requirement for trajectory planning in autonomous vehicles~(AVs) and enables a comprehensive safety evaluation of automated driving systems. To that end, this paper presents a novel probabilistic occupancy risk assessment~(PORA) metric. It uses spatiotemporal heatmaps as probabilistic occupancy predictions of surrounding traffic participants and estimates the risk of a collision along an AV's planned trajectory based on potential vehicle interactions. The use of probabilistic occupancy allows PORA to account for the uncertainty in future trajectories and velocities of traffic participants in the risk estimates. The risk from potential vehicle interactions is then further adjusted through a Cox model\edit{,} which considers the relative \edit{motion} between the AV and surrounding traffic participants. We demonstrate that the proposed approach enhances the accuracy of collision risk assessment in dynamic traffic scenarios, resulting in safer vehicle controllers, and provides a robust framework for real-time decision-making in autonomous driving systems. From evaluation in Monte Carlo simulations, PORA is shown to be more effective at accurately characterizing collision risk compared to other safety surrogate measures. Keywords: Dynamic Risk Assessment, Autonomous Vehicle, Probabilistic Occupancy, Driving Safety
\end{abstract}


\section{Introduction}

Safety is a primary concern in the design and operation of autonomous vehicles~(AVs), requiring robust methods to assess potential collision risks in dynamic and complex traffic environments. Traditional risk assessment methods are based on models of human perception and cognition and rely on subjective interpretation on \edit{the} observed driving environment to understand human drivers' behavioral patterns~\cite{darwish2020learning}. Although these methods are crucial for understanding human decision-making while driving, the introduction of AVs has underscored the need for more generally applicable data-driven approaches to safety evaluation~\cite{pradhan2021driver}.

There are two interrelated perspectives on AV safety. On one hand, AV systems must be designed to ensure that driving decisions remain robust across a variety of dynamic scenarios and traffic participant behaviors, thereby preventing crashes. On the other hand, AV safety evaluation also involves comparing the risk incurred in AV actions to those that a human driver would consider acceptable~\cite{roesener2017comprehensive}. Both perspectives require the ability to accurately assess the risks of a potential course of action given the state of the environment and other vehicles. In response to this need, this paper proposes a novel metric to assess these risks in dynamic environments. Despite many attempts to evaluate AV safety, significant challenges remain due to the inherent dynamic risks and the rarity of \edit{crashes}. Traditional aggregation of \edit{crash} statistics and experimental hypothesis testing techniques have encountered significant challenges when assessing AV safety \cite{de2020procedure, zhou2024scalable}. This is primarily because traditional experimental methods require either overly large datasets or long periods of observation to yield statistically significant results. As a result, a number of studies have been conducted to develop indirect safety measures that estimate collision risk without direct reliance on extensive crash data \cite{das2023surrogate, el2021using}.

One widely adopted approach is the use of traffic-conflict surrogate safety measures~(SSMs), such as Time-to-Collision (TTC) and Time-to-Stop (TTS)~\cite{arun2021systematic, li2021risk}. These measures provide a useful approach for comparing the relative safety of different driving environments, behaviors, or systems~\cite{nikolaou2023exploiting, ka2020implementing}. However, traditional SSMs still possess significant limitations for complex, dynamic environments as they are based on the restrictive assumptions about vehicle trajectories and interactions~\cite{vogel2003}. Specifically, traditional SSMs often rely on deterministic predictions of vehicle trajectories, which cannot account for the uncertainty and variability inherent in traffic participant behaviors in real\edit{-}world traffic scenarios~\cite{ward2015}. This limitation motivates the need for a more robust method that can account for these uncertainties.

Accordingly, for an accurate assessment of driving risk, one must characterize the uncertainty resulting from the dynamics and variability in traffic participants' behaviors. One approach is by representing uncertainty as a collection of potential trajectory samples for each participant~\cite{kim2022diverse, chai2019multipath, neumeier2021variational, ivanovic2020multimodal}. Trajectory sampling methods, such as Gaussian Mixture Models~(GMMs) and Variational Autoencoders~(VAEs), ensure temporal consistency. However, they face challenges in dynamic traffic environments with unpredictable participant interactions~\cite{wiederer2023joint, lu2024quantifying}. Most importantly, their representations are inefficient at distinguishing between vehicle interactions with a low probability. And yet those are the crucial interactions for risk assessment because they have the highest potential to result in crash events. 

\edit{Recent advances, including transformer-based models~\cite{yuan2021agentformer, ngiam2021scene, nayakanti2022wayformer}, diffusion-based frameworks~\cite{jiang2023motiondiffuser, gu2022stochastic, janner2022planning}, and diffusion transformers (DiTs)~\cite{peebles2023scalable}, have improved the representation of multi-agent dynamics and the generation of diverse, high-fidelity trajectories. However, even these methods still face challenges in reliably capturing low-probability but safety-critical events, which are essential for robust risk assessment. A key limitation is that their learned distributions are typically dominated by high-probability, nominal driving patterns, while rare but safety-critical maneuvers are significantly underrepresented~\cite{wiederer2023joint, lu2024quantifying}. This imbalance arises from both data scarcity and likelihood-based training objectives that favor frequent behaviors~\cite{chai2019multipath, ivanovic2020multimodal}. As a result, this imbalance can result in an incomplete characterization of the outcome space, particularly in the low-probability tails of the distribution where aggressive cut-ins, sudden braking, or multi-vehicle interactions occur~\cite{neumeier2021variational, gu2022stochastic}. In the context of our work, the objective is not to model individual trajectories or discrete events per se, but to capture the uncertainty and spatial distribution of possible future occupancies arising from the entire range of potential behaviors, including those rare yet safety-critical cases.}

Heatmap-based occupancy predictors have emerged as a promising solution to capture the probabilistic distribution of future traffic participants' positions~\cite{lange2023self}. Within such an approach, at each time step, the model generates a bird’s-eye view heatmap, a discretized spatial grid where each cell indicates the probability of occupancy by surrounding traffic participants. The set of heatmaps over several future timesteps provides an uncertainty characterization of the trajectories of traffic participants and environmental context. \edit{This representation can capture low-probability but high-impact events while avoiding inefficient or biased heuristic trajectory sampling. This makes heatmap-based predictions an efficient means to represent the full range of potential behaviors, including safety-critical situations, thereby strengthening the reliability of risk estimates for motion planning and collision avoidance. While we adopt heatmaps in this work, the modular framework is compatible with any model capable of producing accurate occupancy distributions.}

Based on occupancy heatmaps, previous approaches proposed a dynamic risk assessment considering the uncertainty in vehicle movements \cite{eidehall2008} by using advanced techniques like Bayesian Occupancy Filters \cite{tay2008efficient} and Dynamic Lambda-Fields~\cite{lutzow2023density}. Alternatively, other approaches utilized machine learning and deep learning techniques \cite{feth2018dynamic, mcallister2017, shi2022real} in the dynamic risk assessments. However, these approaches overlook critical factors like the vehicles' dimensions and the impact of relative \edit{motions} on interaction risk, which are key for accurate collision risk assessment.

\begin{figure}[!ht]
    \centering
    \includegraphics[width=\linewidth]{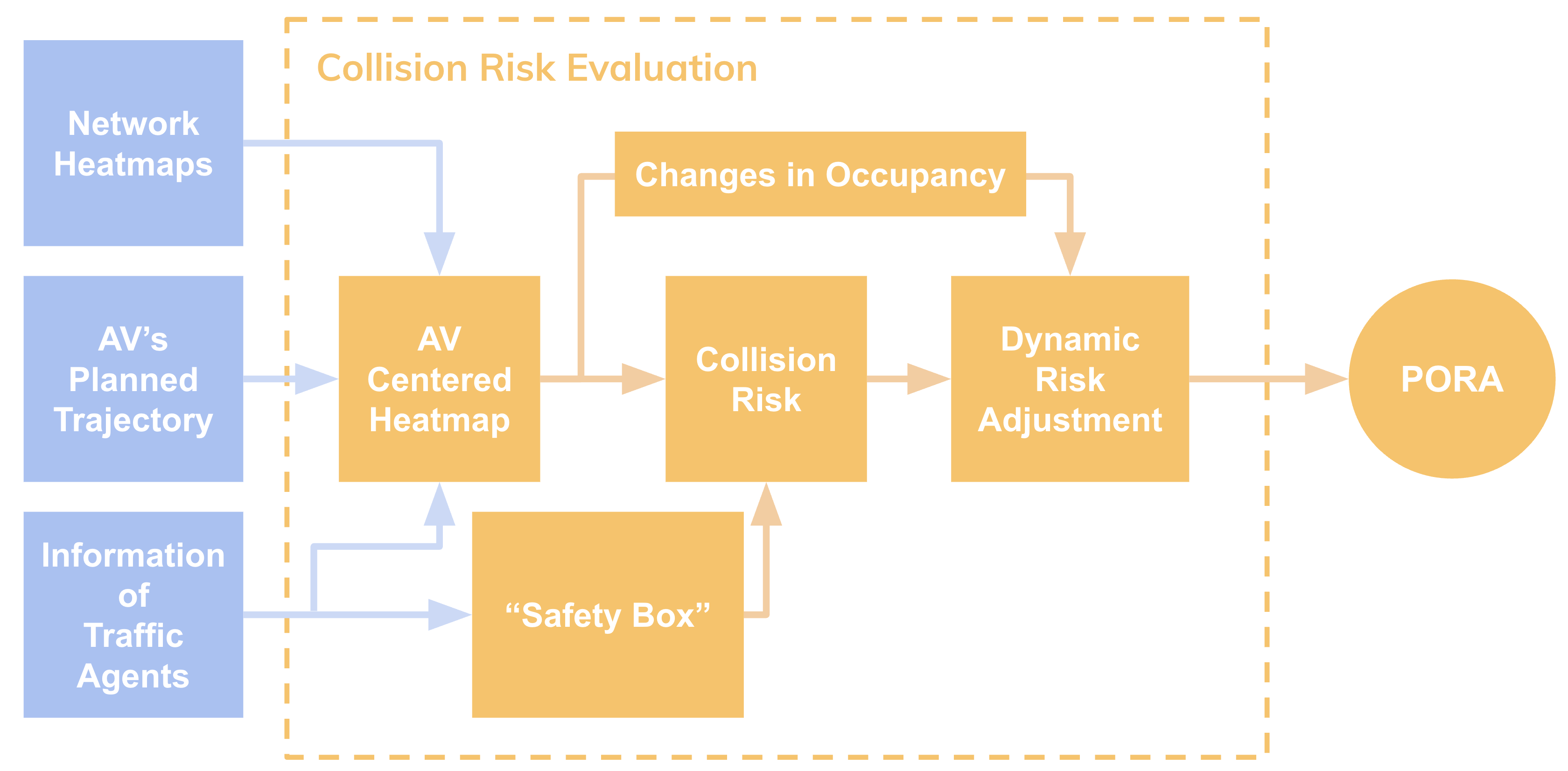}
    \caption{General Approach to Collision Risk Evaluation through PORA.}
    \label{fig: general_framework_PORA}
\end{figure}

To address these limitations, we propose a novel approach to Probabilistic Occupancy Risk Assessment (PORA), as shown in Figure~\ref{fig: general_framework_PORA}. By utilizing heatmaps that represent the short-term predicted occupancy distribution of vehicles in a discretized bird’s-eye-view grid, PORA captures the uncertainties in vehicles' trajectories and evaluates the overlap between predicted vehicle occupancies and their associated probabilities, an essential process for effective collision risk assessment. Furthermore, PORA is designed to incorporate vehicles’ dimensions and account for the relative dynamics among traffic participants, thereby overcoming the limitations of previous heatmap-based measures.

While any heatmap-based occupancy predictor can be utilized to compute PORA, we propose a new heatmap-based occupancy prediction model due to the limitations of the existing models\edit{, including insufficient integration of contextual information from surrounding traffic participants and the environment, as well as inadequate temporal resolution to capture rapidly evolving scenarios.} While a few recent models like GOHOME~\cite{gilles2022gohome} have started to incorporate contextual information, most state-of-the-art heatmap generation models still do not fully integrate contextual information from the surrounding traffic environment effectively. This limitation results in predictions that may be spatially coherent but contextually irrelevant, such as assigning high probabilities to non-drivable areas or neglecting critical cues from the trajectories of other road users~\cite{yang2024generalized}. Furthermore, the temporal resolution of these predictors often does not align well with the dynamic nature of real-time traffic, leading to delayed or overly smoothed predictions that hinder the ability to respond to rapidly evolving scenarios~\cite{meyer2023deep}.

Additionally, the lack of flexibility of many existing trajectory prediction methods has motivated us to propose a modular framework for heatmap generation. These methods rely on the monolithic encoder-decoder framework trained end-to-end to maximize overall performance. However, their lack of modularity limits the adaptability of individual components to task-specific requirements, deployment in new environments, and incorporation of additional data modalities like HD maps or updated traffic behavior models~\cite{rasouli2020deep}, making them impractical in scenarios requiring frequent updates or real-time adaptation. 

\begin{figure}
    \centering
    \includegraphics[width=\linewidth]{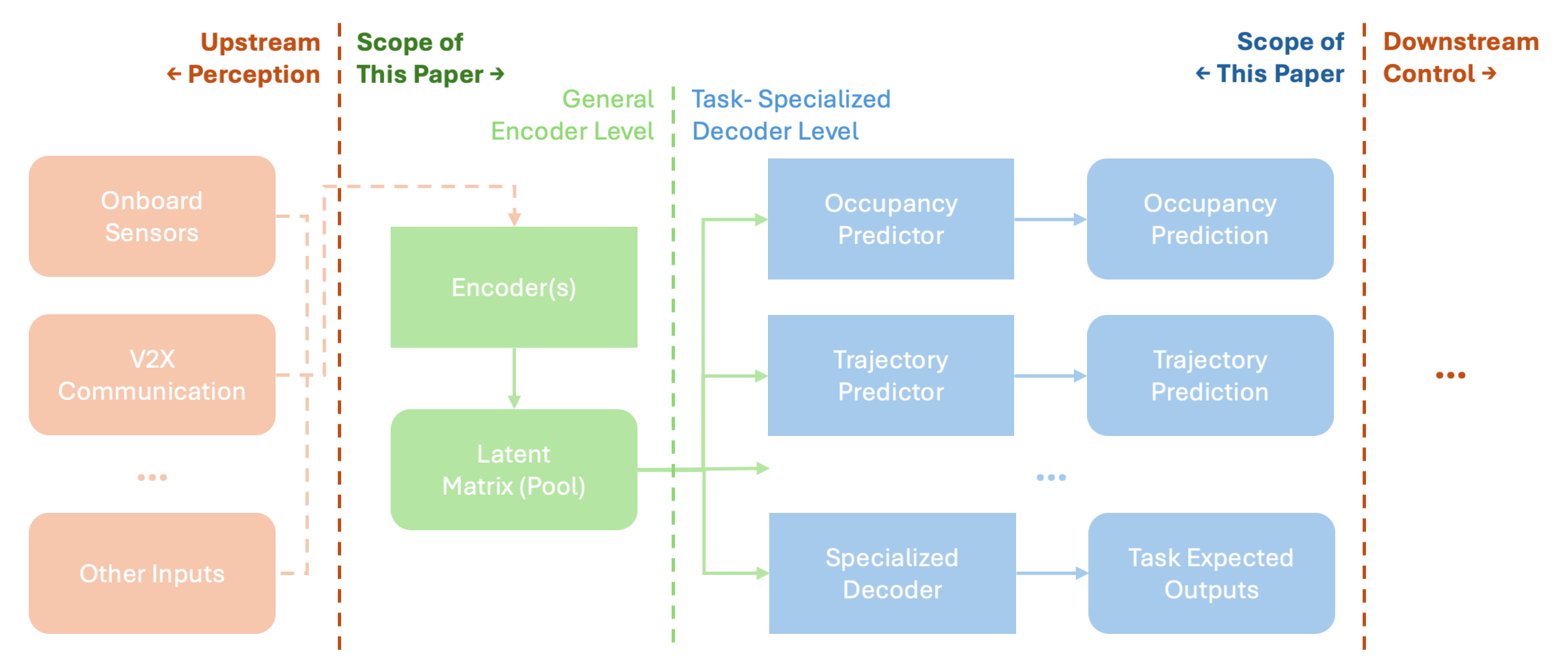}
    \caption{Proposed Modular Framework for Autonomous Vehicle Prediction Tasks.}
    \label{fig:architecture_overview}
\end{figure}

To overcome these limitations, we propose a modular prediction framework that generates a probabilistic occupancy heatmap while accounting for multi\edit{-}participant interactions. The general overview of the modular framework is illustrated in Figure~\ref{fig:architecture_overview}. The framework consists of an upstream perception layer, a general encoder, and task-specific decoders. The perception layer collects data from onboard sensors, V2X, and other sources. The encoder transforms this input into a latent matrix representing spatiotemporal context. This latent matrix is shared across decoders that perform specialized tasks such as occupancy or trajectory prediction. The modular design allows separate optimization of encoder and decoder components, supporting task-specific adaptation while maintaining end-to-end differentiability.

Experimental evaluations using real-world (Argoverse~2) and simulated datasets (SUMO) demonstrate that the proposed framework significantly improves spatial alignment and predictive accuracy, achieving state-of-the-art performance in uncertainty representation. By producing context-aware and uncertainty-informed heatmaps, the proposed generation model enables PORA to deliver more accurate and reliable collision risk assessment.

To validate that PORA provides a more informative risk assessment with the modular heatmap generation framework, we used a scenario-generation-based microsimulation platform, OSM2SUMO, designed for Monte Carlo testing with real-world trajectory datasets and road geometry data. This platform enables us to evaluate the effectiveness of PORA compared to traditional safety surrogate measures in various dynamic traffic scenarios. By using different safety surrogate measures as the negative reward to be minimized by a reinforcement learning controller, we show that PORA achieves significantly better performance in complex traffic scenarios, reducing the number of collisions and conflicts more effectively compared to when traditional safety surrogate measures are used. This means that PORA is more effective at distinguishing between safe situations and potential crash events, providing a more accurate assessment of the risks present to the AV, and thereby leading to safer actions.

The organization of this paper is as follows. In Section~\ref{sec: collision_risk_evaluation}, we outline the process of computing the proposed heatmap-based metric PORA. The overview of the scenario-generation-based microsimulation platform to conduct Monte-Carlo testing and evaluate the performance is presented in Section~\ref{sec:experiments}. The simulation experiment setup for assessing PORA is detailed in Section~\ref{sec:experiments_PORA}, where its results and comparison with other metrics in the simulation platform are shown in Section~\ref{sec: results}.

\section{Collision Risk Evaluation}
\label{sec: collision_risk_evaluation}

In this section, we introduce PORA, a novel heatmap-based metric that provides a more effective characterization of collision risk for autonomous vehicles (AVs). In contrast to previous surrogate safety measures, PORA incorporates data from all surrounding traffic participants and accounts for the variability and uncertainty in their trajectories to provide a more comprehensive collision risk assessment. The design of PORA facilitates collision risk evaluation in scenarios where an AV interacts with multiple surrounding participants. We first give an overview of PORA, and then we will go into the mathematical details in the remainder of the section.

PORA relies on three key pieces of information:
\begin{enumerate}
    \item \textbf{Occupancy heatmaps $\{H_{t_k}: k = 1, \ldots, K\}$} represent the probability of occupancy by surrounding traffic participants on a spatial grid at each future timestep $t_k$\edit{, where each heatmap may correspond to either an immediate next timestep or a longer-term future timestep, depending on the chosen prediction horizon.}
    \item \textbf{The AV planned trajectory}, specifying the expected positions and velocities at each future timestep $t_k$, and 
    \item \textbf{Data on surrounding traffic participants}, including their types (e.g.\edit{,} vehicle, bicycle, pedestrian) and physical dimensions.
\end{enumerate}

Based on these inputs, PORA is computed in two steps. In the first step, we compute a collision risk along the AV planned trajectory based on the predicted occupancy heatmaps $H_{t_k}$. \edit{This is achieved by partitioning $H_{t_k}$ into sub-heatmaps centered on the AV's position and sized according to vehicles' dimensions.}

Then, in the second step, the collision risk is adjusted by incorporating the effect of the velocities and the changes in the occupancy probabilities within the AV-centered heatmaps over time. The adjustment accounts indirectly for the relative \edit{motions} between participants without \edit{the need to explicitly determine} the velocities of the surrounding traffic participants.

Through these steps, PORA addresses the limitations of collision risk estimation based solely on location and provides a representation of inter-participant conflicts and collision risks.

Formally, PORA can be represented as follows: 
\begin{equation*}
\begin{aligned}
    \text{PORA} = \mathtt P = \mathcal{F}(\mathbf{T}_{\text{AV}}, {\mathbfcal V}_{\text{AV}}, H_{t_k}, l_{\text{AV}}, w_{\text{AV}}, \{l_{n}\}, \{w_{n}\})
\end{aligned}
\end{equation*}
where $\mathcal{F}$ is the function that takes the following variables as inputs:
\begin{itemize}
    \item Planned trajectory data of the AV, denoted as $\mathbf{T}_{\text{AV}} = \{\mathbf{x}_{t_k}\}_{k=1}^K$, where $\mathbf{x}_{t_k} \in \mathbb{R}^2$ represents the position coordinates at time $t_k$,
    \item Planned velocities of the AV, denoted as ${\mathbfcal V}_{\text{AV}} = \{\boldsymbol{v}_{t_k}\}_{k=1}^K$, where $\boldsymbol{v}_{t_k} \in \mathbb{R}^2$ represents the velocity vector at time $t_k$, 
    \item Heatmaps $H_{t_k}$,
    \item Length and width of the AV, denoted as $l_{\text{AV}}$ and $w_{\text{AV}}$, respectively,
    \item Length and width of the surrounding traffic participants, denoted as $\{l_{n}\}$ and $\{w_{n}\}$ for each surrounding traffic participant $n$ among the total $N$ surrounding traffic participants. \edit{These are generally estimated by the AV's perception but, as an approximation, average dimensions for each vehicle type can be used.}
\end{itemize}

\begin{figure}
    \centering
    \includegraphics[width=\linewidth]{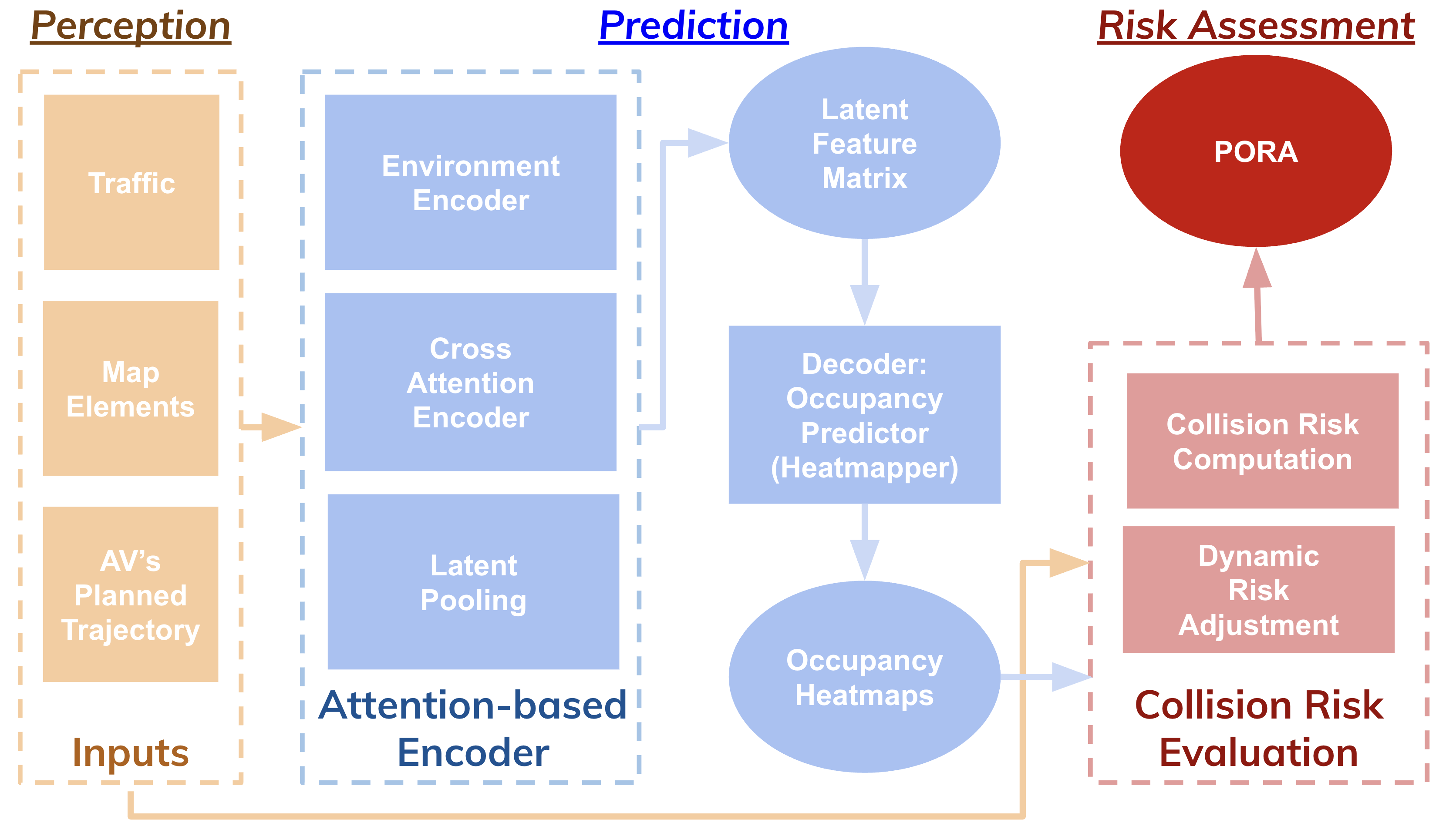}
    \caption{\edit{Data Flow across Perception, Prediction, and Risk Assessment Modules in the PORA Framework. This diagram illustrates the modular encoder-decoder architecture and its integration into the PORA risk assessment pipeline. Perception inputs are processed by an attention-based encoder. A task-specific decoder then generates occupancy heatmaps, which are evaluated through a risk assessment module to compute the PORA metric.}}
    \label{fig:modular_pipeline}
\end{figure}

The following subsections describe the overall process of computing these factors and PORA, including the modular heatmap generation framework and the collision risk with adjustment based on dynamic traffic interactions. \edit{Figure~\ref{fig:modular_pipeline} summarizes the data flow among perception, prediction, and risk assessment components within the PORA computation pipeline.}

\edit{\subsection{Key Definitions}}
\edit{We begin by defining the notation and the key terms that will be used in the remainder of the paper.}

\edit{\begin{itemize}
    \item \textbf{Traffic Participants} refer to all dynamic agents in the traffic environment that may influence the AV's safety. Traffic participants include, but are not limited to, pedestrians, bicyclists, motorcyclists, passenger automobiles, buses, and trucks. 
    \item \textbf{Target Participant} refers to the traffic participant relative to which future motion is being predicted by the occupancy prediction model during the training. This is significant when training the model from data because we incorporate predictions from the perspective of different traffic participants for diversity and completeness. This term is used only to describe the training process of the heatmap generator; after training, when calculating PORA, the AV predicts heatmaps accounting for all surrounding participants rather than a single participant. 
    \item \textbf{Global Occupancy Heatmap $H_{t_k}$} is a grid-based probabilistic representation of the spatial distribution of where all surrounding traffic participants are predicted to be at a future time step $t_k$. In the global occupancy heatmap $H_{t_k}$, the occupancy probabilities are represented over the road network of interest, incorporating road geometry and map information to align predictions with the drivable areas.
    \item \textbf{Safety Box $\Phi$} represents a spatially bounded region around the AV within which the presence or entry of surrounding traffic participants would compromise the AV's safety, potentially requiring modifications to the planned trajectory such as velocity reduction, lane change, emergency braking, and/or evasive steering to avoid or mitigate risk. The size of the safety box is determined by the dimensions of the AV and surrounding traffic participants, in addition to the AV's dynamics, including the AV's current velocity, maximum deceleration capability, and reaction time. 
    \item \textbf{AV-centered heatmaps $\mathtt h_{t_k}$} encode the safety-critical interactions between the AV and surrounding traffic participants through localized probabilistic representations. They are the sub-heatmaps cropped from global occupancy heatmaps $H_{t_k}$, sized according to the safety box $\Phi$. 
    \item \textbf{Occupancy Probability $P_{(\mathtt i,\mathtt j),{t_k}}(O)$} is the probability that grid cell $(\mathtt i,\mathtt j)$ in the AV-centered heatmap $\mathtt h_{t_k}$ is occupied by surrounding traffic participant at future time $t_k$. Higher occupancy probabilities indicate greater interaction potential and possible increased risk to the AV, but do not guarantee a collision.
    \item \textbf{Conditional Probability of a Collision given Occupancy $P_{(\mathtt i,\mathtt j),{t_k}}(\mathcal C|O)$} is the probability of a collision occurring at grid cell $(\mathtt i,\mathtt j)$ given that the cell is occupied by a traffic participant. Multiplying this value by the occupancy probability $P_{(\mathtt i,\mathtt j),{t_k}}(O)$ gives the collision probability at that grid cell based on the multiplication rule for independent events. 
    \item \textbf{Subarea $\varphi \subset \Phi$} represents the region within the AV-centered heatmaps where any interaction between the AV and another traffic participant results in a guaranteed collision. 
    \item \textbf{Relative Motion} refers to the rate at which the AV and surrounding traffic participants approach or separate from each other, serving as the multi-dimensional analogue of relative velocity in one-dimensional space. In this project, due to the challenges in extracting future trajectories for surrounding traffic participants, it is incorporated indirectly through the change in occupancy probability $\Delta P_{(\mathtt i,\mathtt j),{t_k}}(O)$. 
    \end{itemize}}

\subsection{Heatmap Generation - Modular Framework}

Heatmaps are generated as probabilistic occupancy grids that represent the likelihood of \edit{traffic participants} occupying specific grid cells over time. \edit{The set of occupancy heatmaps $\{H_{t_k}: k = 1, \ldots, K\}$ is generated over a prediction horizon of $K$ next timesteps. These timesteps may be a lower temporal resolution than that used in AV control to incorporate predictions several seconds into the future with fairly small values of $K$.}

While any probabilistic occupancy predictor can be used to compute PORA, we introduce a modular, end-to-end differential framework designed to support flexible and scalable occupancy prediction. By decoupling the encoder and decoder, this framework enables independent component training, robust uncertainty estimation, and seamless integration with risk assessment metrics like PORA.

Let us discretize the two-dimensional space in grid cells, indexed by $i \in [0, \mathcal I]$ and $j \in [0, \edit{\mathcal J}]$ with $\mathcal{I}$ and $\edit{\mathcal J}$ being the number of discretization steps along length and width dimensions, respectively. Then, let $H_{t_k}$ denote the heatmap at future time $t_k$, where each grid cell $(i, j)$ represents 
\begin{equation*}
    H_{t_k}(i, j) = \text{P}(\text{occupancy at grid cell } (i, j) \mid t_k).
\end{equation*}

\begin{figure}
    \centering
    \includegraphics[width=\linewidth]{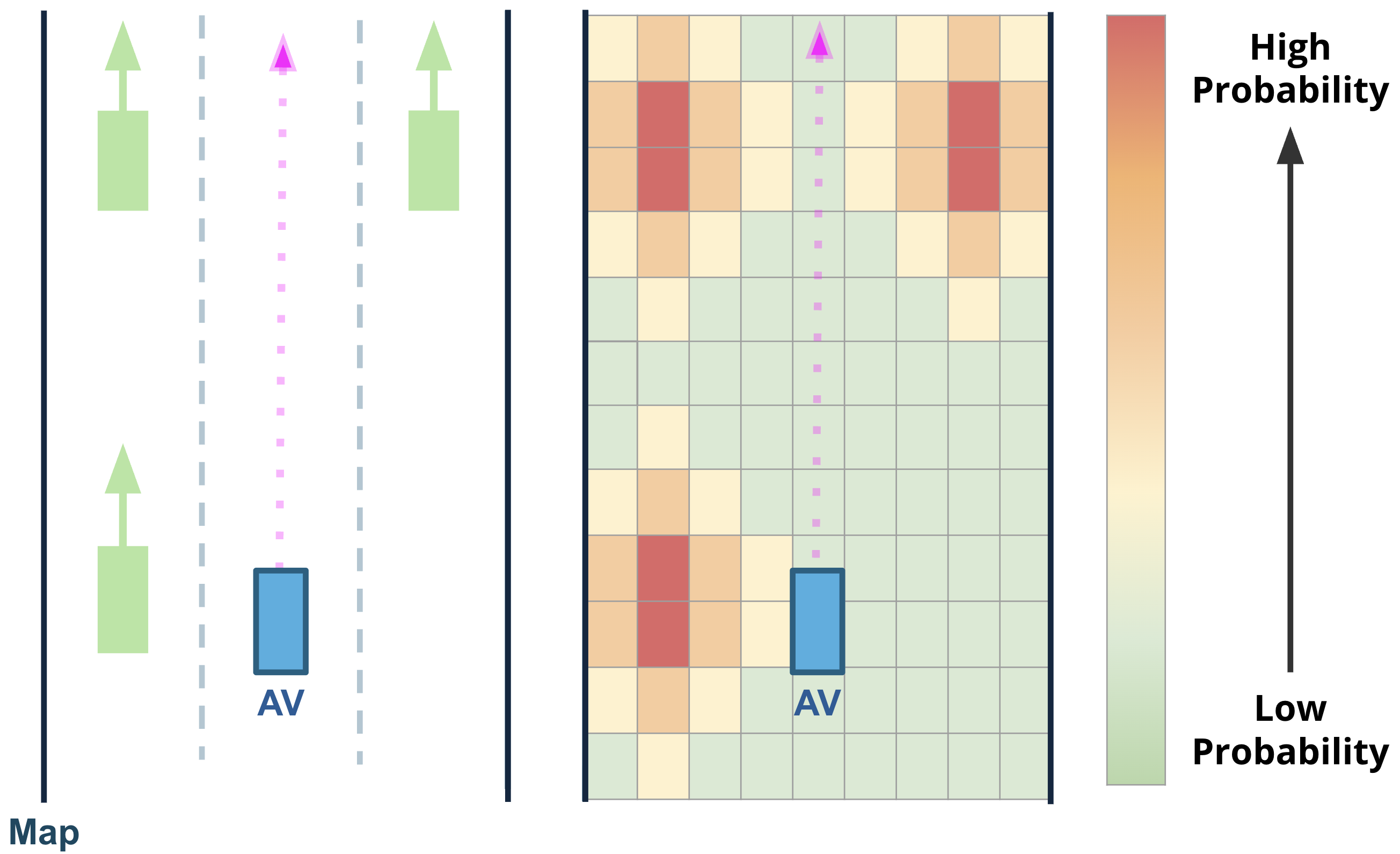}
    \caption{Illustration of the elements in the \edit{example} traffic scenario and its heatmap $H_{t_k}$: traffic \edit{participants} (including AV) and map. \edit{The pink dotted line represents the AV's planned trajectory in this example scenario with the planned direction of motion.} The leftmost part of the figure shows a bird's\edit{-}eye view of a scenario with 4 vehicles in a 3 lane road with multiple traffic participants (green) whose future motion is to be predicted by the AV (blue). The rightmost part of the figure shows the corresponding predicted occupancy probability heatmap, where red indicates higher occupancy probability and green indicates lower probability. \edit{The prediction shown in the heatmap corresponds to the immediate next timestep, but the framework also supports longer prediction horizons.} The spatial grid resolution is finer in practice; a coarser resolution is used in these examples so that the grid can be clearly visualized.}
    \label{fig:scn_elements}
\end{figure}

Figure~\ref{fig:scn_elements} shows an example scene and its corresponding heatmaps. The objective is to estimate a probabilistic heatmap that characterizes the future positions likely to be occupied by \edit{surrounding traffic participants (green)}. To do so, the autonomous vehicle (blue, labeled as ``AV") incorporates contextual information from \edit{past trajectories of traffic participants} and static road features such as lane boundaries.

The heatmap in Figure~\ref{fig:scn_elements} thus shows the predicted spatial distribution at a future timestep. By utilizing this information, the AV can anticipate the future motion\edit{s} of \edit{surrounding traffic participants} and adjust its planned trajectory accordingly.

We now present the details regarding the modular framework to generate occupancy heatmaps, as shown in Figure~\ref{fig:architecture_overview}. \edit{The training objective is to estimate a probabilistic heatmap that characterizes the future positions likely to be occupied by a target \edit{participant}, defined as the traffic participant whose motion is the focus of the prediction during training. We note that the target \edit{participant} is not necessarily an AV. It may be any participant whose behavior the AV seeks to predict, and this designation is used to diversify and enrich the training data by incorporating perspectives from different traffic participants.} At any time $t_0$, the framework predicts heatmaps over $K$ future time steps using observations from $I$ past time steps:
\begin{equation*}
    H_{t_k} = F(\mathbf{T}_{\text{target}}, \mathbf{T}_{\text{traffic}}, \mathbf{M}; \theta_e, \theta_d),
\end{equation*}
where $F$ is an end-to-end differentiable function with encoder parameters $\theta_e$ and decoder parameters $\theta_d$.

The inputs are:
\begin{itemize}
    \item $\mathbf{T}_{\text{target}}$: Past trajectory of the target \edit{participant}, given by $\mathbf{T}_{\text{target}} = \{\mathbf{x}_{n_\text{target}, t_k}\}_{k=-I}^{0}$, where $n_\text{target}$ is the index of the target.
    \item $\mathbf{T}_{\text{traffic}}$: Trajectories of other surrounding \edit{participants}, $\mathbf{T}_{\text{traffic}} = \{\mathbf{T}_n\}$ for $n \in \{1,2, \dots, N\} \setminus \{n_\text{target}\}$, with each $\mathbf{T}_n = \{\mathbf{x}_{n,t_k}\}_{k=-I}^{0}$.
    \item $\mathbf{M}$: Vectorized map data containing static elements such as road boundaries and lane markings.
\end{itemize}

The subsequent sections describe the encoder-decoder architecture and the modular training procedure in detail.

\subsubsection{Encoder: Transformer-based Context Extractor}
\label{sec:encoder_heatmap}

\begin{figure*}
    \centering
    \includegraphics[width=1\linewidth]{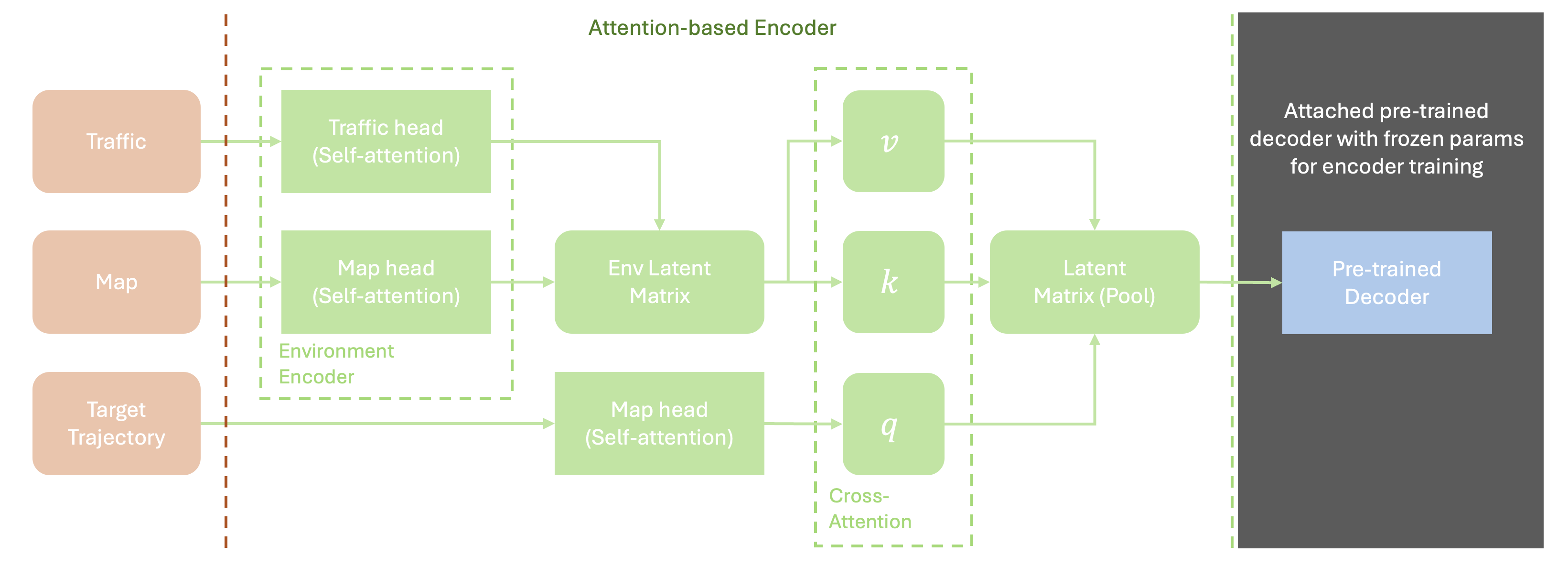}
    \caption{Attention-Based Encoder for Modular Feature Extraction in Autonomous Vehicle Prediction: In the heatmap prediction setting, the encoder processes multiple input streams, including traffic data, map information, and target trajectory, to generate a shared latent representation that captures spatial and temporal dependencies. Each input type is encoded independently through self-attention heads, such as the traffic and map heads, which extract relevant features into an environment latent matrix. A cross-attention mechanism refines this matrix by integrating information across input sources, resulting in a unified latent matrix (pool) that serves as the contextual embedding for downstream prediction tasks. During encoder training, a pre-trained decoder with fixed parameters is attached, allowing the encoder to learn representations aligned with the decoder's input requirements.}
    \label{fig:hm_encoder}
\end{figure*}

With the architecture shown in Figure~\ref{fig:hm_encoder}, the encoder extracts spatial-temporal features from $\mathbf{T}_{\text{target}}$, $\mathbf{T}_{\text{traffic}}$, and $\mathbf{M}$, producing a latent matrix $\mathbf{Z} \in \mathbb{R}^{d_z}$. Each self-attention layer $\text{SA}(\cdot)$ maps the corresponding input into an embedding space of dimension $d_z$: 
\begin{align*}
    \mathbf{E}_{\text{target}} &= \text{SA}(\mathbf{T}_{\text{target}}) \in \mathbb{R}^{d_z}, \\
    \mathbf{E}_{\text{traffic}} &= \{\text{SA}(\mathbf{T}_n)\}_{n \in \{1,2,\dots,N\}\setminus\{n_{\text{target}}\}} \in \mathbb{R}^{N-1 \times d_z}, \\
    \mathbf{E}_{\text{map}} &= \text{SA}(\mathbf{M}) \in \mathbb{R}^{d_z}.
\end{align*}

These embeddings are then combined through a cross-attention module with the query $\mathbf{Q}$, key $\mathbf{K}$, and value matrices $\mathbf{V}$: 
\begin{align*}
    \mathbf{Q} &= \text{Linear}(\mathbf{E}_{\text{target}}), \\
    \mathbf{K} &= \text{Linear}(\mathbf{E}_{\text{traffic}} \oplus \mathbf{E}_{\text{map}}), \\
    \mathbf{V} &= \text{Linear}(\mathbf{E}_{\text{traffic}} \oplus \mathbf{E}_{\text{map}}),
\end{align*}
where $\oplus$ denotes concatenation. The cross-attention output is then given by:
\begin{equation}
    \text{Attention}(\mathbf{Q}, \mathbf{K}, \mathbf{V}) = \text{softmax}\left(\frac{\mathbf{Q} \mathbf{K}^\top}{\sqrt{d_{\text{k}}}}\right) \mathbf{V},
\end{equation}
producing the latent matrix $\mathbf{Z}$, which encapsulates temporal vehicle dynamics and static spatial context from the map. This matrix is used as input for downstream prediction tasks.

\begin{figure}
    \centering
    \includegraphics[width=1\linewidth]{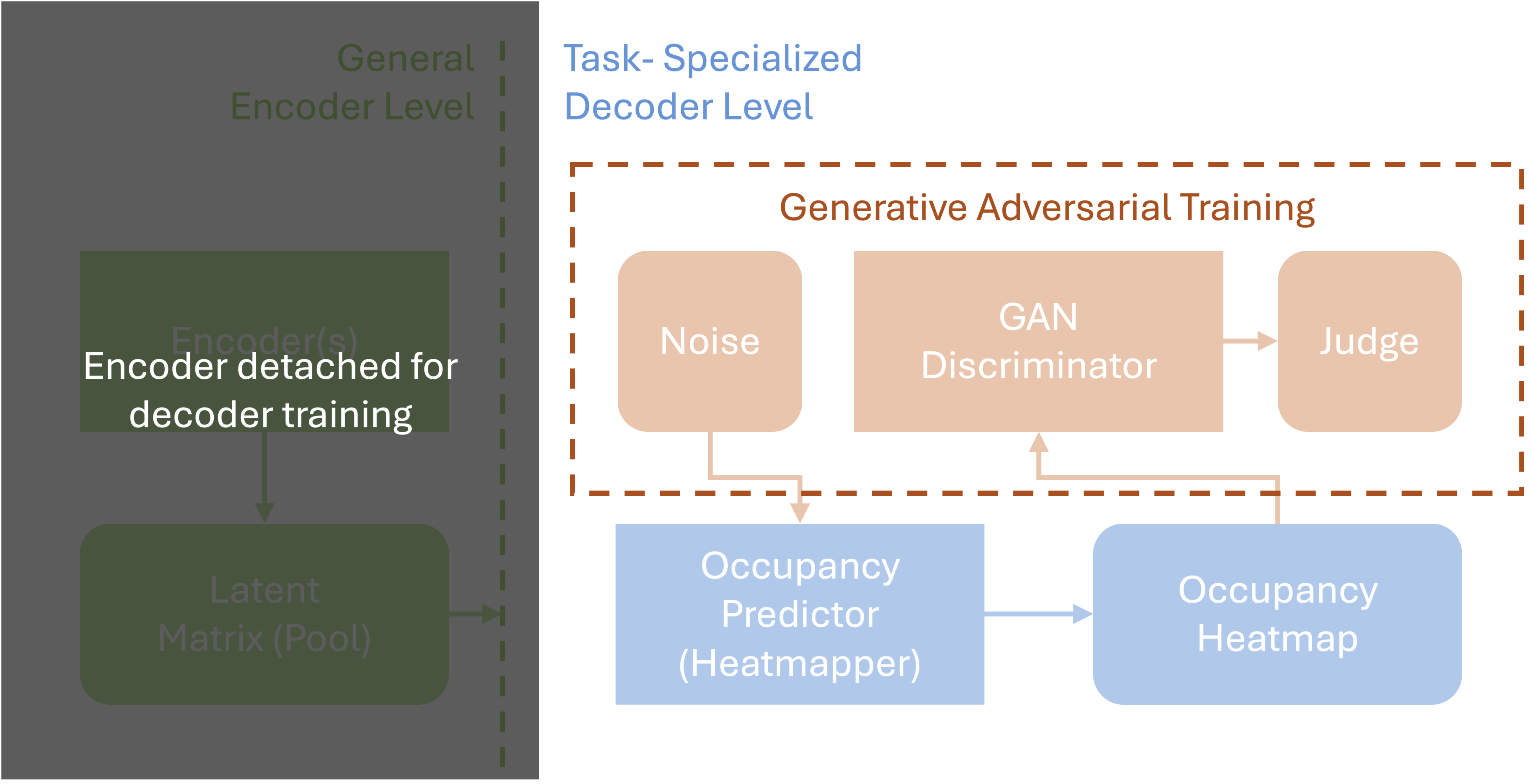}
    \caption{Decoder Training Process with Generative Adversarial Network (GAN) for Occupancy Prediction: The encoder is detached to allow independent training of the task-specific decoder. The decoder uses a GAN architecture, where the generator (occupancy predictor or 'heatmapper') takes input noise and produces probabilistic occupancy heatmaps representing the likelihood of vehicle presence over a spatial grid. The discriminator evaluates the realism of the generated heatmaps by distinguishing them from real samples, while an additional component ('judge') assesses output quality based on task-specific evaluation metrics.}
    \label{fig:hm_decoder}
\end{figure}

\subsubsection{Decoder: GAN-based Occupancy Heatmap Generator}
\label{sec:decoder_heatmap}

Figure~\ref{fig:hm_decoder} illustrates the decoder structure used in the heatmap prediction framework. The decoder, based on a Generative Adversarial Network (GAN) generator $G$, maps the latent matrix $\mathbf{Z}$ to a sequence of probabilistic occupancy heatmaps $\{H_{t_k}\}_{k=1}^K$. Each heatmap $H_{t_k}$ is generated using a latent vector $\mathbf{z}_{t_k}$, extracted from $\mathbf{Z}$: 
\begin{equation*}
    H_{t_k} = G(\mathbf{z}_{t_k}; \theta_d).
\end{equation*}

In this framework, the decoder is pre-trained independently using a GAN setup before integration with the encoder. The details of the training procedure are provided in Section~\ref{sec:modular_training}.

A discriminator $D$ is also trained to distinguish between real and generated heatmaps. The adversarial loss function is defined as:
\begin{equation}\label{E:adversarial_loss_func}
\begin{split}
\min_{G} \max_{D} \mathcal{L}(D, G) = 
& \ \mathbb{E}_{\mathbf{H} \sim p_{\text{data}}} [\log D(\mathbf{H})] \\
& + \mathbb{E}_{\mathbf{z} \sim p_z} [\log (1 - D(G(\mathbf{z})))],
\end{split}
\end{equation}
where $\mathbf{H}$ denotes real heatmaps sampled from the true data distribution $p_{\text{data}}$, and $\mathbf{z}$ is sampled from the latent distribution $p_z$. This adversarial setup encourages the generator to produce heatmaps that closely resemble real occupancy distributions and their dynamics.

\subsubsection{Modular Training Approach}
\label{sec:modular_training}

The modular training procedure separates encoder and decoder training into distinct phases. 

\begin{figure}[h]
    \centering
    \includegraphics[width=1\linewidth]{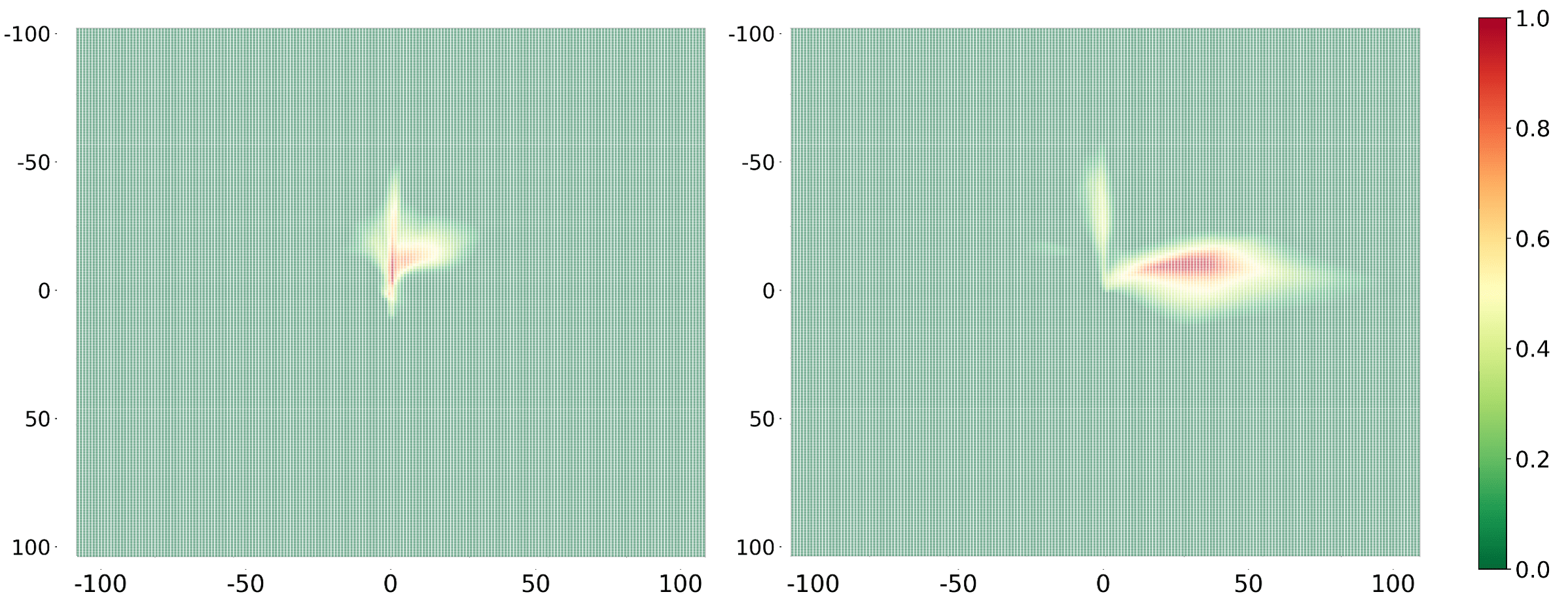}
    \caption{Initial Heatmap Outputs Generated by the GAN Decoder Without Encoder Attached: These heatmaps are generated by the decoder using only noise inputs. At this stage, they do not encode predictive content.}
    \label{fig:decoder_only_hm}
\end{figure}

\paragraph{Independent Decoder Training} The decoder is initially trained independently using the adversarial loss described in Section~\ref{sec:decoder_heatmap}.

The dataset for decoder pre-training is constructed from simulated vehicle trajectories recorded at the "trigger points" (e.g., intersections, merging zones, and straight segments). Then, these trajectories are converted into a sequence of $K$ heatmaps, representing the predicted occupancy distribution for a corresponding future prediction horizon. Figure~\ref{fig:decoder_only_hm} presents two sample heatmaps generated by the decoder from noise inputs during this initial phase.

The objective of the independent decoder training phase is to enable the generator to produce spatially coherent, realistic heatmaps, establishing a foundation for subsequent encoder training.

\paragraph{Encoder Training with Fixed Decoder} With the weights of \edit{the} pre-trained decoder held fixed, the encoder is trained to produce latent representations that the decoder can map to accurate occupancy heatmaps.

With the ground-truth heatmaps $\{H_{t_k}\}_{k=1}^K$ constructed from recorded vehicle trajectories, the encoder is optimized using a negative cross-entropy loss
\begin{equation}
\label{eq:encoder_loss}
    L = -\frac{1}{C} \sum_{c=1}^{C} \left[ y_c \log(\hat{y}_c) + (1 - y_c) \log(1 - \hat{y}_c) \right],
\end{equation}
where $C$ is the number of grid cells, $y_c$ is the binary ground-truth occupancy value for cell $c$, and $\hat{y}_c$ is the predicted probability from the decoder.

\paragraph{End-to-End Fine-Tuning} Finally, an optional fine-tuning phase jointly optimizes both the encoder and decoder parameters by minimizing prediction loss:
\begin{equation}
    \min_{\theta_e, \theta_d} L(\mathbf{H}, \hat{\mathbf{H}}),
\end{equation}
where $\hat{\mathbf{H}}$ denotes the predicted heatmaps. This phase preserves the modular structure while allowing the encoder and decoder to be co-optimized for improved performance.

\edit{After training, the resulting occupancy prediction model is applied by the AV at inference time to generate heatmaps that account for all surrounding traffic participants over the prediction horizon, unlike during training where the model is conditioned on a single designated ``target participant''.}

\subsection{Collision Risk}

With the proposed modular framework generating probabilistic occupancy heatmaps \edit{accounting for all surrounding traffic participants}, we begin computing PORA by first calculating the collision risk using the AV  planned location and the heatmaps $H_{t_k}$.

The computation of collision risk for PORA involves three key sub-steps. First, we define the ``safety box'' $\Phi$, a dynamic spatial region around the AV determined by both the dimensions of the surrounding traffic participants and the AV's dynamic behavior. Once $\Phi$ is established, we extract the AV-centered heatmaps $\mathtt h_{t_k}$ from the global heatmaps $H_{t_k}$ to focus on the conflicts between the AV and surrounding traffic participants in this critical region. The collision risk is then computed using $\mathtt h_{t_k}$ and the conditional probability of a collision, given that a surrounding traffic participant occupies a grid within $\Phi$.

\subsubsection{Safety Box}

While potential conflicts may occur across various locations in the network, our focus is specifically on ensuring the safety of the AV. To this end, we consider only the critical area most directly \edit{relevant to} the AV's safety. We call this critical area, the safety box $\Phi$. The safety box $\Phi$ characterizes the range of all possible collision locations based on the vehicles' dimensions and the AV's absolute speed. 

Here, we emphasize that the AV predicts probabilistic occupancy heatmaps for multiple surrounding \edit{participants} for the purpose of evaluating its own safety. \edit{At inference time, when calculating PORA, the AV applies the trained heatmap generator to produce single set of heatmaps capturing the future behavior of all surrounding participants, thereby enabling overall collision risk assessment that considers interactions with every participant, without requiring separate heatmaps from the perspective of each participant.}

We denote the width of the safety box of an AV by $\mathtt w$:
\begin{equation*}
     \mathtt w = w_{\text {AV}} + \max_{n \in [1,N]} {l_n} 
\end{equation*} 
where $w_{\text {AV}}$ and $l_n$ are the width of the AV and the lengths of all the other traffic participants, respectively. The width of the safety box is designed to detect vehicles in an orthogonal crash path (i.e., a ``T-bone" collision). 

Unlike the width of the safety box, the length of the safety box will frequently change over time as we consider the absolute velocity of the AV through the stopping sight distance. The American Association of State Highway and Transportation Officials (AASHTO) established that the stopping sight distance (SSD) considering the braking time and the reaction time of a driver \cite{hancock2013policy} is given by
\begin{equation*}
    \text{SSD}({t_k}) = \left(0.278 \textbf{v}({t_k})\mathsf r + \frac{\textbf{v}({t_k})^2}{\edit{(254 / 9.81)}\sa}\right)
\end{equation*}
where $\edit{\textbf{v}}$ is the absolute speed of the vehicle (km/h), $\mathsf r$ is the perception-reaction time (sec), and $\sa$ is the deceleration rate (m/$\text{sec}^2$). 

The SSD is considered in defining the dimension of the safety box to account for the fact that the AV cannot stop completely instantaneously and continue traveling by the stopping sight distance even under maximum braking. Consequently, the range of all possible collision locations depends on the stopping distance and, ultimately, the absolute velocity of the AV. Since the vehicle primarily moves in the direction of the heading when braking is applied, the stopping sight distance $\text{SSD}({t_k})$ extends only the length of the safety box in that direction.

Accordingly, the length of the safety box of an AV, $\mathtt l (t_k)$, is defined as: 
\begin{equation*}
     \mathtt l (t_k) = l_{\text {AV}} + \max_{n \in [1,N]} {l_n} + \text{SSD}({t_k})
\end{equation*}
where $l_{\text {AV}}$ is the length of an AV. 

\begin{figure} [!ht]
    \begin{subfigure}[b]{0.49\linewidth}
        \centering
        \includegraphics[width=0.9\linewidth]{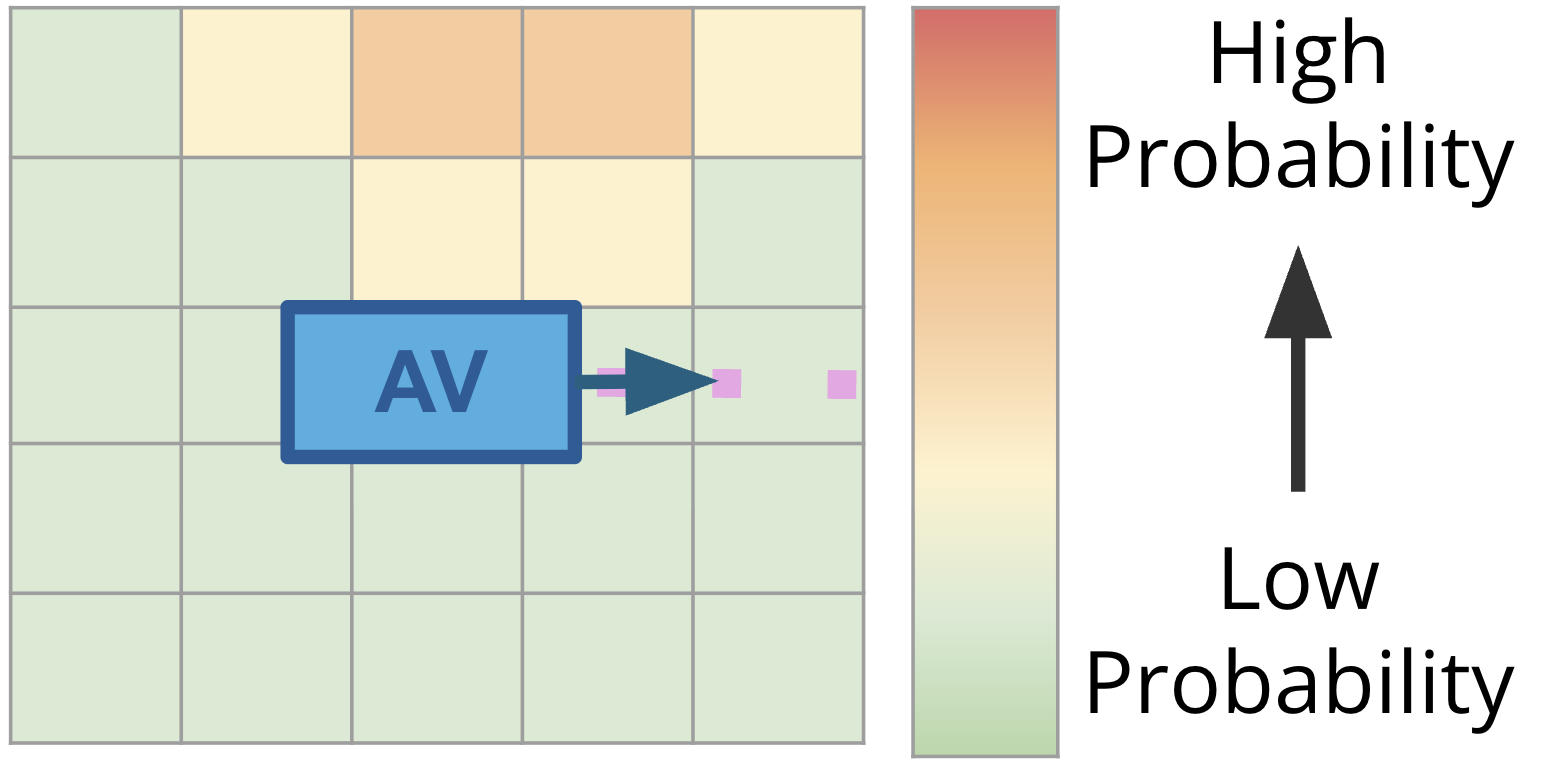}
        \caption{\edit{Low Speed (i.e., Small $\lvert \vec{v}({t_k}) \rvert$)}}
        \label{sec:heatmap_v_0}
    \end{subfigure}
    \begin{subfigure}[b]{0.49\linewidth}
        \centering
        \includegraphics[width=0.995\linewidth]{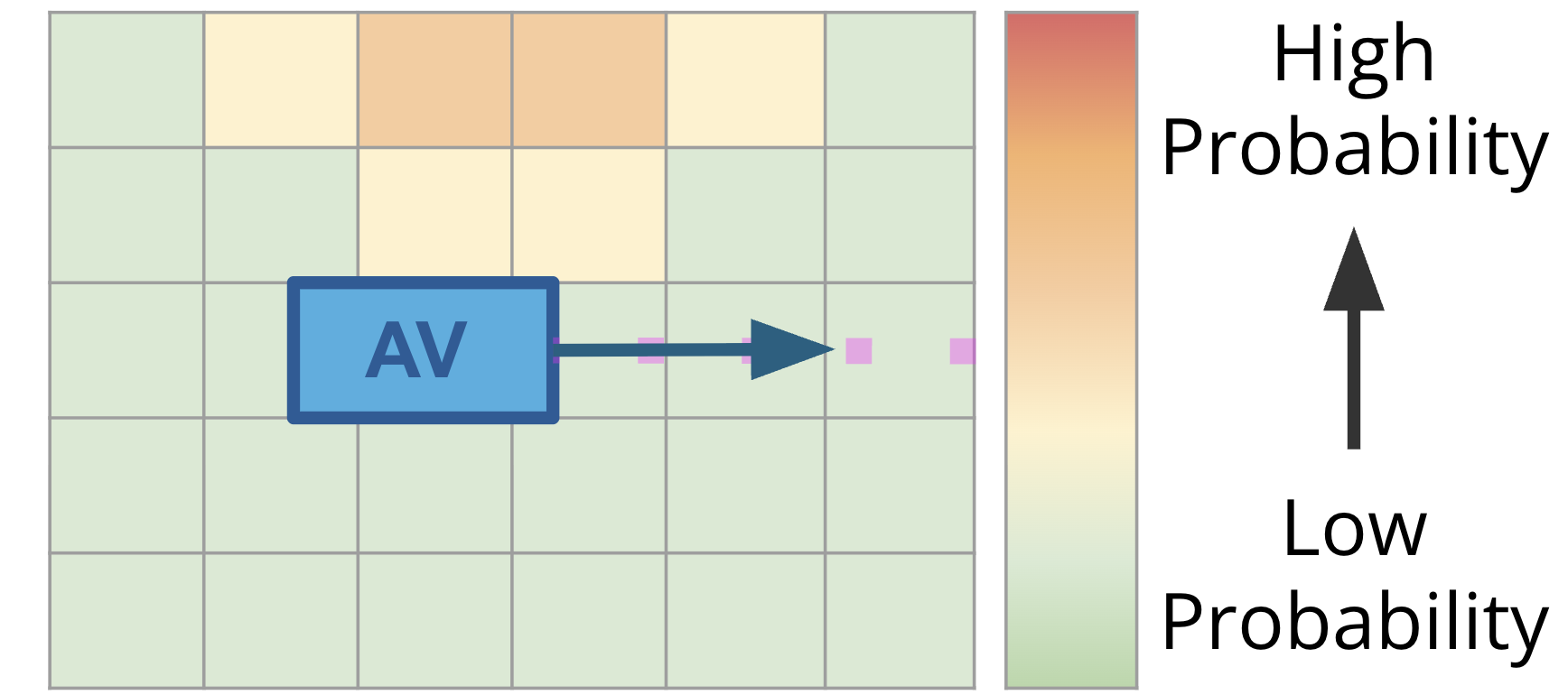}
        \caption{\edit{High Speed (i.e., Large $\lvert \vec{v}({t_k}) \rvert$)}}
        \label{sec:heatmap_v_no_0}
    \end{subfigure}
    \caption{Example of a safety box $\Phi$ depicting how its dimension depends on the magnitude of the absolute velocity $\vec{v}({t_k})$ of the AV (i.e., the ``ego'' vehicle) \edit{as illustrated in the example scenario shown in Figure~\ref{fig:scn_elements}}. The blue arrow indicates the direction of movement\edit{, where its length represents the magnitude of the absolute velocity $\vec{v}({t_k})$. The pink dotted line shows the AV's planned trajectory, as in Figure~\ref{fig:scn_elements}.} The heatmaps in this figure are at a coarser grid resolution for better visualization.}
    \label{fig: h_t}
\end{figure}

Figure \ref{fig: h_t} shows an example of a safety box $\Phi(t_k)$ with dimensions $\mathtt l (t_k) \times \mathtt w$ and how its length changes with the absolute velocity of an AV.

\subsubsection{AV-Centered Heatmaps}

\begin{figure} [!ht]
    \centering
\includegraphics[width=0.9\linewidth]{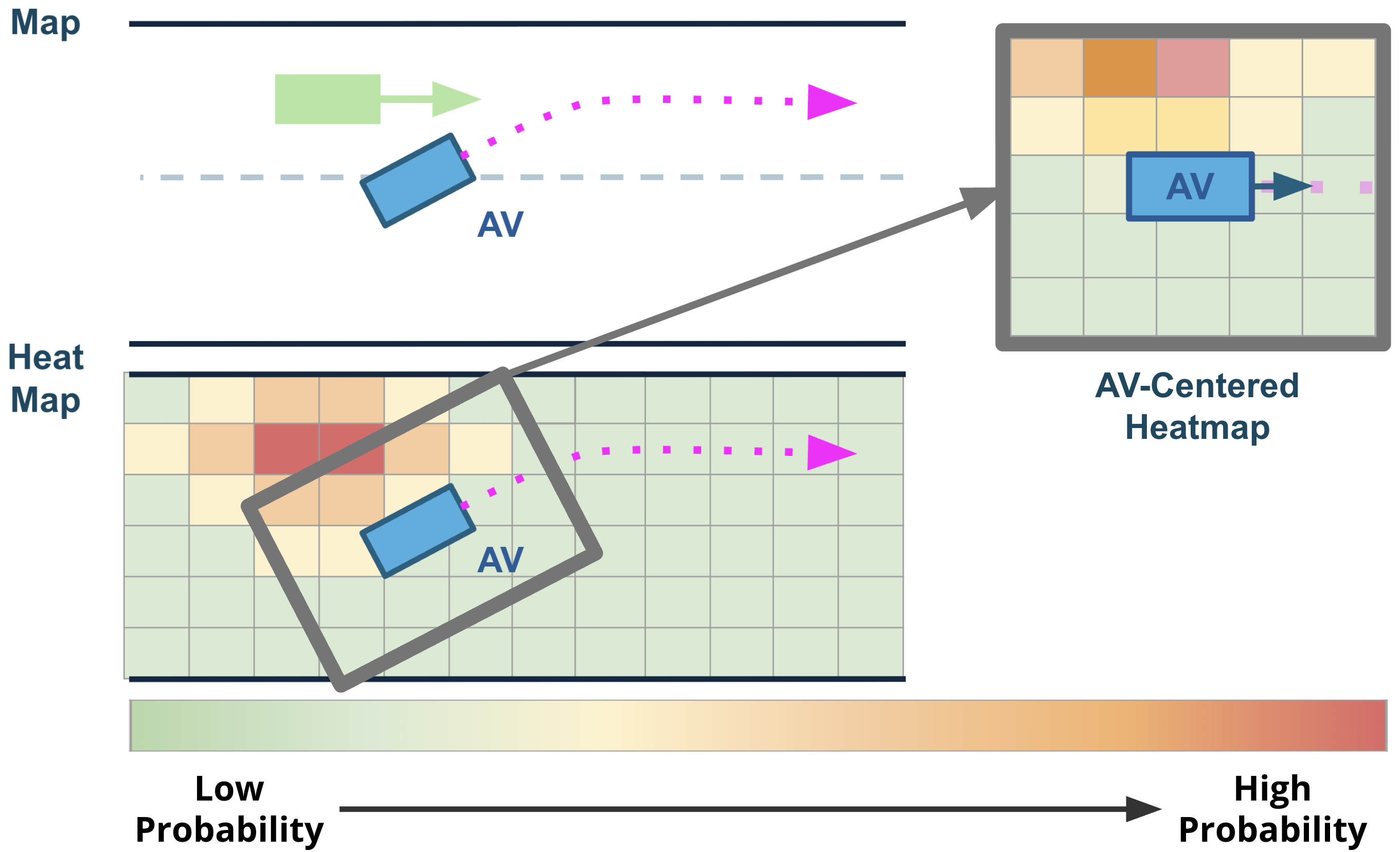}
    \caption{\edit{Example of an AV-centered heatmaps $\mathtt h_{t_k}$, derived from the global heatmaps $H_{t_k}$ reflecting the predicted occupancy at immediate next timestep. The top-left shows a bird's-eye view of a two-lane scenario with the AV and another vehicle (green). The bottom-left displays the predicted global occupancy heatmap $H_{t_k}$, where red indicates high and green indicates low occupancy probability. Higher probabilities represent grid cells that other agents are more likely to occupy. Based on the definition of the safety box, high occupancy in the AV-centered heatmap suggests greater potential interaction, which may pose increased risk for the AV, whereas lower probabilities indicate less likelihood of interaction. The AV's planned trajectory is shown in pink. The AV-centered heatmap, shown in the top-right, is extracted by translating, rotating, and cropping $H_{t_k}$ around the AV to the size of the safety box (shown as the rectangle with gray boundary), aligning the AV's direction of travel. Heatmaps in this figure use a coarser grid resolution for ease of interpretation.}}
    \label{fig: sim_CP}
\end{figure}

After computing the length and the width of the safety box $\Phi$, we generate the AV-centered heatmaps $\mathtt{h}_{t_k}$, whose dimension matches those of $\Phi$. Specifically, we translate, rotate, resample, and trim the global heatmaps $H_{t_k}$ into $\mathtt{h}_{t_k}$ using the AV's velocity vector $\mathbf{v}_{t_k}$ to align them with the AV's direction of travel along its planned path. This transformation—incorporating the AV's position, velocity, and heading—ensures that the AV-centered heatmaps capture the region of potential interactions with other traffic participants accurately. Figure \ref{fig: sim_CP} illustrates the concept of the AV-centered heatmaps $\mathtt{h}_{t_k}$.

After we construct the AV-centered heatmaps $\mathtt h_{t_k}$ for each time ${t_k}$ within the prediction horizon [${t_1}$, ${t_K}$], we use them to estimate the probability of spatial conflicts between the AV and another traffic participant at a given location. Let $\mathtt i$ and $\mathtt j$ denote the spatial indices representing the rows and columns of each AV-centered heatmap $\mathtt h_{t_k}$. The pair $(\mathtt i,\mathtt j)$ then corresponds to a given spatial location within the AV-centered heatmaps. That is,
\[
\mathtt h_{\edit{t_k}}(\mathtt i,\mathtt j) = \text{Prob}(\text{occupancy }(\mathtt i, \mathtt j) \text{ at time } {t_k}) = P_{(\mathtt i,\mathtt j),{t_k}}(O). 
\]
As a result, the AV-centered heatmap $\mathtt h_{\edit{t_k}}$ represents the probability that each grid cell $(\mathtt i,\mathtt j)$ is occupied by a participant at each time step $t_k$, denoted by $P_{(\mathtt i,\mathtt j),{t_k}}(O)$. \edit{Higher probability of occupancy within the AV-centered heatmap reflects a greater chance of interaction with other agents, which may signal increased risk to the AV, while lower probability indicates reduced potential of interaction.}

\subsubsection{Collision Risk: Probability of Collision}

While the AV-centered heatmap $\mathtt h_{\edit{t_k}}$ may predict that a traffic participant will be within the safety box $\Phi$ of an AV, it does not necessarily imply a collision. The collision event depends on the traffic participant's heading and position. To accurately assess the AV's collision risk, we evaluate the conditional probability of a collision given the occupancy for each grid cell $P_{(\mathtt i,\mathtt j),{t_k}}(\mathcal C|O)$. 

Within the safety box $\Phi$, we define the subarea $\varphi \subset \Phi$ where a collision always occurs between the AV and a traffic participant. The length $\mathtt L$ and the width $\mathtt{W}$ of the subarea $\varphi$ is defined as 
\begin{equation*}
    \begin{cases}
        \mathtt L = l_{\text{AV}} + \min_{n \in [1, N]}w_n,\\
        \mathtt{W} = w_{\text{AV}} + \min_{n \in [1, N]}w_n,
    \end{cases}
\end{equation*}
where $w_n$ represents the widths of all traffic participants. The width and the length of the subarea are fixed over time.

The probability of collision given that a surrounding traffic participant occupies a grid located at $(\mathtt i,\mathtt j)$, $P_{(\mathtt i,\mathtt j),{t_k}}(\mathcal C|O)$, is described as \begin{equation*}
    P_{(\mathtt i,\mathtt j),{t_k}}(\mathcal{C}|O) = 
    \begin{cases}
         1, \quad \text{if } (\mathtt i,\mathtt j) \in \varphi, \\
         0, \quad \text{if } (\mathtt i, \mathtt j) \notin \Phi, \\
         f((\mathtt i, \mathtt j)), \quad \text{otherwise}.
    \end{cases}
\end{equation*}
where $f(\mathtt i, \mathtt j)$ is the spatial weighting function that computes $P_{(\mathtt i,\mathtt j),{t_k}}(\mathcal C|O)$ based on the grid cell's proximity to the center of the vehicle. Specifically\edit{,} $f$ decreases as the grid cell approaches the edge of the safety box $\Phi$, reflecting reduced collision likelihood. \edit{This spatial weighting function $f$ can be derived through calibration using real-world data, ensuring that the resulting estimates provide accurate collision risk assessment.}

\begin{figure} [!ht]
    \centering
    \includegraphics[width=0.7\linewidth]{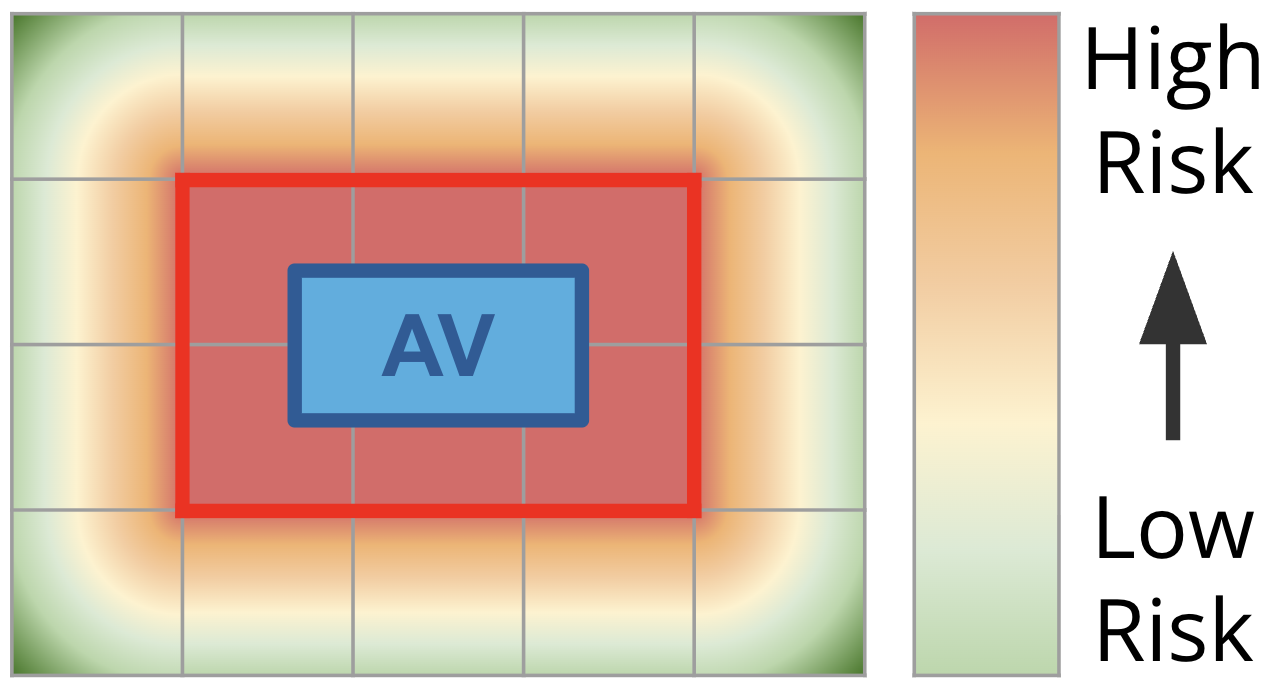}
    \caption{Probability of Collision Happening within the example safety box given the occupancy $P_{(\mathtt i,\mathtt j),{t_k}}(\mathcal C|O)$. \edit{Importantly, this probability in each grid cell reflects the magnitude of the risk given that the corresponding grid cell is occupied. Please note that the probability of collision does not depend on the occupancy predictions represented in heatmaps. Here, the red region with the red-colored boundary, representing the highest risk (i.e., Definite Guarantee of Collision), is the subregion $\varphi$ within the safety box $\Phi$. The probability of collision decreases as one moves from the boundary of the subregion $\varphi$ toward the edge of the safety box $\Phi$.} Coarser resolution is used for the heatmap here to enhance clarity and to serve an illustrative purpose.}
    \label{fig: condition_collision_prob}
\end{figure}

Figure~\ref{fig: condition_collision_prob} shows $P_{(\mathtt i,\mathtt j),{t_k}}(\mathcal C|O)$ in the example safety box. 

With $P_{(\mathtt i,\mathtt j),{t_k}}(\mathcal C|O)$ and $P_{(\mathtt i,\mathtt j),{t_k}}(O)$, the probability of collision at a grid location, $P_{(\mathtt i,\mathtt j),{t_k}}(\mathcal C)$, is given by $P_{(\mathtt i,\mathtt j),{t_k}}(\mathcal C) = P_{(\mathtt i,\mathtt j),{t_k}}(\mathcal C|O) \cdot P_{(\mathtt i,\mathtt j),{t_k}}(O)$. Through this process, we compute the collision probability of each grid location in the safety box and for each timestep ${t_k}$ along the planned trajectory. 

\subsection{Dynamics-adjusted collision risk} \label{sec: limitation}

Assessing the collision risk solely based on the collision probability has limitations, as dynamics and kinematic motions of moving participants play a vital role in dynamic risk assessment \cite{jiao2023autonomous}. To improve the accuracy of collision risk assessment, both the absolute velocities of traffic participants and the relative \edit{motion} \edit{--- that is, the rate at which they approach or separate from each other ---} should be considered \cite{jiao2023autonomous}. 

The AV's absolute velocity was already incorporated in the risk assessment by dynamically adjusting the length of the AV's safety box. This also accounted for the AV's stopping sight distance, which directly depends on its absolute velocity, and its impact on the likelihood of a collision. Still, one should further consider the velocity of the other traffic participants in estimating the risk arising from those interactions.

However, incorporating the relative \edit{motion} between vehicles directly is more challenging. Due to the stochastic nature of the occupancy prediction, the future trajectories and absolute velocities of surrounding traffic participants are unknown in the prediction horizon. The estimation of the planned trajectory and absolute velocity of surrounding traffic participants from the most probable location of surrounding traffic participants at each time in the prediction horizon might not be realistic due to the inherent uncertainties. As a result, relative \edit{motions} between AVs and other surrounding traffic participants can not be directly accessed.

To address this issue, we account for the relative \edit{motion} indirectly by computing $\Delta P_{(\mathtt i,\mathtt j),{t_k}}(O)$, the change in the occupancy probability at grid location $(\mathtt i,\mathtt j)$ from the previous timestep $t_{k-1}$ to the current timestep $t_k$. That is, 
\begin{equation*}
    \Delta P_{(\mathtt i,\mathtt j),{t_k}}(O) = P_{(\mathtt i,\mathtt j),{t_k}}(O) - P_{(\mathtt i,\mathtt j),{t_{k-1}}}(O).
\end{equation*}
The change in the occupancy probability reflects how fast surrounding traffic participants approach the AV, how fast the AV approaches the surrounding traffic participants, or both. 
As such, $\Delta P_{(\mathtt i,\mathtt j),{t_k}}(O)$ contains the same information as the relative \edit{motion} without needing explicit knowledge of the exact trajectories of surrounding traffic participants\edit{, and avoids computing the relative motion between each surrounding \edit{participant} individually.} Therefore, using  $\Delta P_{(\mathtt i,\mathtt j),{t_k}}(O)$ for the adjustment of risk serves the same role as considering relative \edit{motions} for dynamic collision risk assessment.

\edit{To empirically validate the relationship between the change in occupancy probability, $\Delta P_{(\mathtt i,\mathtt j),{t_k}}(O)$, and the relative motion, we perform a correlation analysis. Here, the relative motion is defined as the change in the center-to-center distance (i.e., Euclidean distance between geometric centers) between the two vehicles in consecutive time steps. For each scenario, we compute correlation coefficients separately, and then aggregate the results across scenarios using scenario-weighted Fisher z-transformation for Pearson and Spearman correlations, and weighted means for Kendall's tau, with $n-3$ weighting based on the number of observations per scenario. Note that correlation coefficients range between -1 and +1, where -1 indicates perfect negative correlation and +1 indicates perfect positive correlation. Across scenarios, the number of samples ranged from 14,045 to 163,887 (median: 87,922), totaling 704,261 samples. The aggregated results are summarized in the following table.}

\begin{table}[!ht]
\centering
\resizebox{\textwidth}{!}{\begin{tabular}{lcccc}
\toprule
\textbf{Correlation Measure} & \textbf{Scenario-Weighted Overall Correlation Coefficient} \\
\midrule
Pearson & -0.5115 \\
Spearman & -0.5092 \\
Kendall & -0.4465 \\
\bottomrule
\end{tabular}}
\caption{\edit{Correlation Results Across All Evaluated Scenarios}}
\label{table:correlation_analysis}
\end{table}

\edit{As shown in Table~\ref{table:correlation_analysis}, all three correlation coefficients are consistently negative, indicating that larger increases in center-to-center distance are generally associated with larger decreases in predicted occupancy probability in consecutive timesteps. This supports the interpretation that the change in occupancy probability inherently reflects the relative motion between the AV and surrounding traffic participants. That is, $\Delta P_{(\mathtt i,\mathtt j),{t_k}}(O)$ captures the same essential information as relative motions without requiring explicit knowledge of surrounding participants' trajectories, thereby serving the same functional role as relative motion in dynamic collision risk assessment.}

\edit{The moderate magnitudes of the correlation coefficients, which are lower than expected, motivate examination of the underlying causes. This is likely due to several reasons: decreased heatmap accuracy at longer prediction horizons, maneuver variability across scenarios (e.g., turning vs. going straight at intersections), and discretization errors introduced by the gridded occupancy representation. In a later section, we further assess the usefulness of the risk-adjustment based on the change in occupancy probability by comparing PORA against alternative heatmap-based risk metrics that do not apply risk-adjustment based on the change in occupancy probability.}

To incorporate dynamic information $\Delta P_{(\mathtt i,\mathtt j),{t_k}}(O)$ into the collision risk assessment, we employ the Cox proportional hazards model \cite{cox1972regression}. This statistical tool adjusts the baseline hazard by accounting for the impact of various factors, including time-dependent variables and complex variables \cite{fisher1999time, abd2021methods, kumar1994proportional}. Due to the flexibility and robustness of the model, the Cox model is widely used in the medical and financial fields to evaluate the survival rates and investment risk\edit{,} respectively \cite{zhang2018time, bellera2010variables, lane1986application}. 

Given the properties of the Cox model, we adopt approaches similar to those utilized in medical and financial fields to implement the dynamic collision risk assessment in our application. We set the risk as the baseline risk and adjust it based on the changes in the occupancy probability $\Delta P_{(\mathtt i,\mathtt j), t_{k}}(O)$. 

Equation~\eqref{eq:cox_model} adjusts the collision probability to get the updated collision risk ${\mathcal P}^*_{(\mathtt i,\mathtt j), t_{k}}(\mathcal{C})$.
\begin{equation}
{\mathcal P}^*_{(\mathtt i,\mathtt j), t_{k}} \\ = 
\begin{cases}
    P_{(\mathtt i,\mathtt j), t_{k}}(\mathcal{C}), & \text{if } k = 1 \\
    P_{(\mathtt i,\mathtt j), t_{k}}(\mathcal{C}) \cdot \exp(\vartheta), & \text{otherwise}
\end{cases} 
\label{eq:cox_model}
\end{equation}
where $\vartheta = \beta \cdot \Delta P_{(\mathtt i,\mathtt j), t_{k}}(O)$. Equation \eqref{eq:cox_model} shows that ${\mathcal P}^*_{(\mathtt i,\mathtt j), t_{k}}$ is less than $P_{(\mathtt i, \mathtt j), t_{k}}(\mathcal{C})$ when $\Delta P_{(\mathtt i,\mathtt j), t_{k}}(O) < 0$. This reduction in collision risk aligns with the principle that the likelihood of a collision decreases accordingly if a traffic participant rapidly disappears from an AV's perspective. Conversely, ${\mathcal P}^*_{(\mathtt i,\mathtt j), t_{k}}$ is greater than $P_{(\mathtt i,\mathtt j), t_{k}}(\mathcal{C})$ if $\Delta P_{(\mathtt i,\mathtt j), t_{k}}(O) > 0$, reflecting an increased risk of collision when vehicles appear suddenly.

Since $P_{(\mathtt i,\mathtt j),{t_k}}(\mathcal C) \in [0,1]$ and $\Delta P_{(\mathtt i,\mathtt j),{t_k}}(\mathcal C) \in [-1, 1]$, ${\mathcal P}^*_{(\mathtt i,\mathtt j),{t_k}}$ is in $[0, \exp(\beta)]$. To facilitate the interpretation and enable direct comparability with  $P_{(\mathtt i,\mathtt j),{t_k}}(\mathcal C)$, we perform the normalization to make the updated risk be in $[0,1]$: 
\begin{equation}
    \hat{\mathcal P}^*_{(\mathtt i,\mathtt j),{t_k}} =  \begin{cases}
P_{(\mathtt i,\mathtt j),{t_k}}(\mathcal C), & \text{if } {k} = 1\\
\frac{1}{\exp(\beta)}{\mathcal P}^*_{(\mathtt i,\mathtt j),{t_k}}, & \text{otherwise}.
\end{cases} 
\label{eq: norm_cox_model}
\end{equation}

Through normalization, it is also possible to capture the case where the participant is in the same location with respect to the AV, $P_{(\mathtt i,\mathtt j),{t_k}} = P_{(\mathtt i,\mathtt j),{t_{k+1}}}$ (either moving at the same speed or stopping at the intersection). In this case, the AV is well aware of the behavior of participants. Therefore, the risk of collision is reduced, but not entirely eliminated, and also not reduced as much as in the case where $\Delta P_{(\mathtt i,\mathtt j),{t_k}}(O) < 0$.

To have accurate dynamic collision risk assessment through $\hat{\mathcal P}^*_{(\mathtt i,\mathtt j),{t_k}}(\mathcal C)$, the parameter $\beta$ needs to be calibrated. The calibration can be performed in two ways. One method is based on a training dataset with known collision times $\check {t}$. In this case, $\beta$ is calibrated so that 
\begin{equation*}
\begin{aligned}
    \beta^* = & \argmax_{\beta} \hat{\mathcal P}^*_{(\mathtt i,\mathtt j),\check {t}}\\
    \textrm{s.t. } & \hat{\mathcal P}^*_{(\mathtt i,\mathtt j),{t_k}} \in [0,1] \quad \forall k \in [1,K] \\
    & \hat{\mathcal P}^*_{(\mathtt i,\mathtt j),\check {t}} \geq \hat{\mathcal P}^*_{(\mathtt i,\mathtt j),{t_k}} \quad \forall k \in [1,K]. \\
\end{aligned}
\end{equation*}
In other words, $\beta$ is calibrated so that it maximizes the collision risk at the known collision time for different scenarios while making sure that the collision risk within the prediction horizon is between 0 and 1.

Since \edit{real-world trajectory} datasets \edit{involving} collisions are rare \edit{and/or} \edit{largely} inaccessible \edit{to the public}, an alternate method is to calibrate the parameter $\beta$ through simulations to avoid and mitigate collisions as much as possible. That is, $\beta$ can be calibrated using simulations so that 
\begin{equation*}
    \beta^* = \argmin_{\beta} \edit{\mathtt{J}}(\beta) 
\end{equation*}
where $\edit{\mathtt{J}}(\beta)$ represents the collision cost function in simulations. 

As the maximum collision probability over a spatial grid yields a more conservative and effective measure of risk \cite{rummelhard2016probabilistic}, we define the collision risk at each timestep $t_k$ as follows:
\begin{equation}
    \text{PORA} = \mathtt P  = \max_{(\mathtt i,\mathtt j)} \hat{\mathcal P}^*_{(\mathtt i,\mathtt j),{t_k}}
\end{equation}
This formulation shows that PORA aggregate\edit{s} the collision risk of every grid such that it reflects the worst-case risk scenario, aligning with the objective of avoiding a collision regardless of where they may occur. 

\edit{Since PORA provides a normalized risk score between 0 and 1, it can be directly integrated into real-time decision-making for AV trajectory planning. A low PORA value (e.g., near 0) indicates that the planned trajectory is considered safe, allowing the AV to proceed as intended. If PORA exceeds a predefined design threshold, the AV may abort the current maneuver and explore alternative trajectories. In situations where PORA is close to 1, the AV may need to initiate braking to avoid an imminent collision. These decision thresholds can be determined through empirical calibration or statistical analysis of PORA values across various scenarios and their safety outcomes. Moreover, thresholds can be made context-dependent, adapting to traffic congestion, maneuver type, or environmental conditions. Being context-dependent enables more flexible and risk-sensitive planning. This integration bridges the gap between quantitative risk assessment and practical AV behavior planning.}

\edit{Alternatively, PORA can be used as a negative reward component for reinforcement learning of safer policies. This use case is employed in the simulations for evaluation of PORA, as detailed in Section~\ref{sec:av.controller.training}. Even after training, PORA can be used as a safety metric for monitoring the policy's performance and identifying situations for augmenting training.}

In conclusion, PORA provides a comprehensive dynamic collision risk assessment by computing and adjusting risk based on an AV's planned trajectory, the probabilistic occupancy of surrounding traffic participants (represented in heatmaps), and dynamic factors such as absolute and relative vehicle motions.

\section{Experimental Evaluation of the Proposed Heatmap Generation Framework}
\label{sec:experiments}

In order to evaluate PORA, we need to first train and validate the underlying modular framework responsible for generating the probabilistic occupancy heatmaps. This section describes the experimental setup, training procedure, and performance evaluation of the proposed modular framework for probabilistic occupancy heatmap generation. The goal is to assess the accuracy of the heatmaps, their spatial alignment with drivable areas, and the ability to capture uncertainties relevant to autonomous vehicle operations.

\subsection{Experimental Setup}
\label{sec:experimental_setup}

The proposed platform was evaluated using both real-world and simulated datasets:

\begin{itemize}
    \item \textbf{Real-World Data:} The Argoverse 2 Motion Forecasting Dataset~\cite{Argoverse2} was used to evaluate the model on real-world urban traffic scenarios. This dataset provides annotated trajectories of traffic participants captured from diverse city environments. It includes:
    \begin{itemize}
        \item 3D tracking annotations across more than 11{,}000 real-world scenarios,
        \item Trajectories of vehicles, pedestrians, and cyclists,
        \item High-definition (HD) maps with detailed urban road geometry and lane-level annotations.
    \end{itemize}

    \edit{While the Argoverse 2 Motion Forecasting Dataset provides real-world trajectory data across a variety of scenarios, including critical situations such as near-misses useful for risk analysis, it exhibits the following limitations:}
    \edit{\begin{enumerate}
        \item \textbf{Potential Inconsistencies Between Position and Speed Data:} Some raw trajectories contain inconsistencies between the reported positions and speeds, which can introduce errors in downstream analysis~\cite{li2023comparative}.
        \item \textbf{No Real Collision Event Trajectories:} Similar to many real-world trajectory datasets, Argoverse 2 does not include trajectories leading to actual collisions. While it contains near-misses and high-risk interactions, it lacks sequences with real crash events. 
    \end{enumerate}}

    The prediction task involves forecasting future participant locations over a 5-second horizon based on observed sequences. The trajectory for each participant is represented as:
    \begin{equation*}
        \mathbf{T} = \left\{ (x_1, y_1, t_1), (x_2, y_2, t_2), \ldots, (x_K, y_K, t_K) \right\},
    \end{equation*}
    where $(x_k, y_k)$ denotes the spatial coordinates at time $t_k$, and $K$ is the number of observed time steps.

    \item \textbf{Simulated Data:} The SUMO (Simulation of Urban MObility) platform was used to generate synthetic traffic scenarios. Real-world road layouts were reconstructed for six U.S. cities from the Argoverse dataset: Austin (TX), Detroit (MI), Miami (FL), Pittsburgh (PA), Palo Alto (CA), and Washington, D.C. SUMO enabled the modeling of diverse scenarios, including intersections, highways, and mixed traffic environments. For each simulation, trigger points were defined to extract vehicle trajectories and generate occupancy heatmaps for \(K\) future frames.
\end{itemize}

This combination of real-world and simulated datasets was designed to support generalization across heterogeneous traffic conditions and enhance the platform’s applicability to real-world AV scenarios.

\begin{figure*}
    \centering
    \includegraphics[width=1\linewidth]{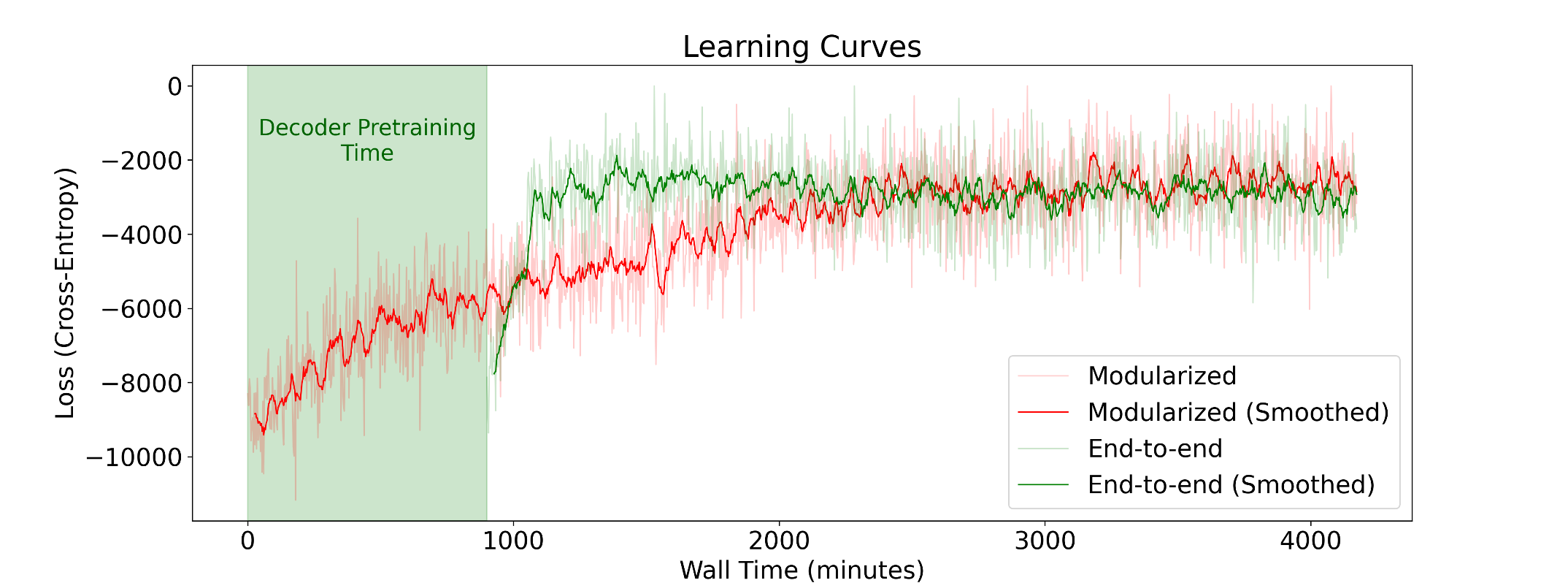}
    \caption{Comparison of Learning Curves for Modular and End-to-End Training Approaches: The green-shaded region indicates the decoder pretraining phase in the modular framework. After pretraining, the modular training curve resumes with a horizontal offset corresponding to the pretraining duration. The modular approach exhibits faster convergence.}
    \label{fig:benchmark_training_curve}
\end{figure*}

\subsection{Training Procedure}
\label{sec:training_procedure}

The training process was organized into distinct modular phases to enable flexible and interpretable model development.

\subsubsection{Initial Decoder Training Using GAN}
\label{sec:decoder_training}

The initial decoder (\textbf{Decoder 0}) was trained using a Generative Adversarial Network (GAN) framework to produce realistic occupancy heatmaps.

The GAN components were defined as follows:
\begin{itemize}
    \item \textbf{Generator (\(G\)):} The decoder received noise vectors \(\mathbf{z} \sim p_z\) and generated predicted heatmaps \(H_{t_k}\).
    \item \textbf{Discriminator (\(D\)):} Trained to distinguish between real heatmaps from the dataset (\(\mathbf{H}\)) and generated samples (\(\hat{\mathbf{H}} = G(\mathbf{z})\)).
\end{itemize}

The adversarial loss was optimized using Equation~\eqref{E:adversarial_loss_func}. During this phase, the decoder generated heatmaps with realistic spatial structure but without meaningful alignment to scene context or road constraints.

\subsubsection{Encoder-Decoder Integration and Encoder Training}
\label{sec:encoder_training}

Following decoder pretraining, the encoder was integrated with the decoder, whose parameters were fixed. The encoder was trained to map input data \((\mathbf{T}_{\text{target}}, \mathbf{T}_{\text{traffic}}, \mathbf{M})\) into latent representations \(\mathbf{Z}\) that the decoder could use to produce accurate heatmaps:
\begin{equation*}
    \mathbf{Z} = f_{\theta_e}(\mathbf{T}_{\text{target}}, \mathbf{T}_{\text{traffic}}, \mathbf{M}),
\end{equation*}
where \(f_{\theta_e}\) denotes the encoder with parameters \(\theta_e\). The encoder was optimized using the cross-entropy loss in Equation~\eqref{eq:encoder_loss}.

Two encoder variants were developed:
\begin{itemize}
    \item \textbf{Encoder 0:} Trained on Argoverse data only,
    \item \textbf{Encoder 1:} Trained on both Argoverse and SUMO datasets to improve generalization across real and synthetic domains.
\end{itemize}

\subsubsection{Decoder Fine-Tuning}
\label{sec:decoder_fine_tuning}

To reduce false-positive predictions in non-drivable areas, the decoder was fine-tuned under two constraints:

\begin{itemize}
    \item \textbf{Filtered Dataset (\(\mathcal{D}_{\text{filtered}}\)):} Samples with non-vehicular target participants (e.g., pedestrians or cyclists) were excluded to constrain learning to vehicle-specific motion on drivable areas.
    \item \textbf{Drivable Area Penalty:} An additional loss term penalize\edit{s} predictions outside drivable regions:
    \begin{equation*}
        \mathcal{L}_{\text{penalty}} = \lambda \sum_{(i, j) \notin \text{Drivable}} H_{t_k}(i, j),
    \end{equation*}
    where \(\lambda\) is a hyperparameter that controls the penalty strength, and \((i, j)\) indexes grid cells outside the drivable area.
\end{itemize}

Two additional decoder variants were produced through fine-tuning:
\begin{itemize}
    \item \textbf{Decoder 1:} Trained using \(\mathcal{D}_{\text{filtered}}\),
    \item \textbf{Decoder 2:} Trained using \(\mathcal{D}_{\text{filtered}}\) with the drivable area penalty \(\mathcal{L}_{\text{penalty}}\).
\end{itemize}

\begin{figure*}[h!]
    \centering
    \begin{subfigure}[b]{0.32\textwidth}
        \includegraphics[width=\textwidth]{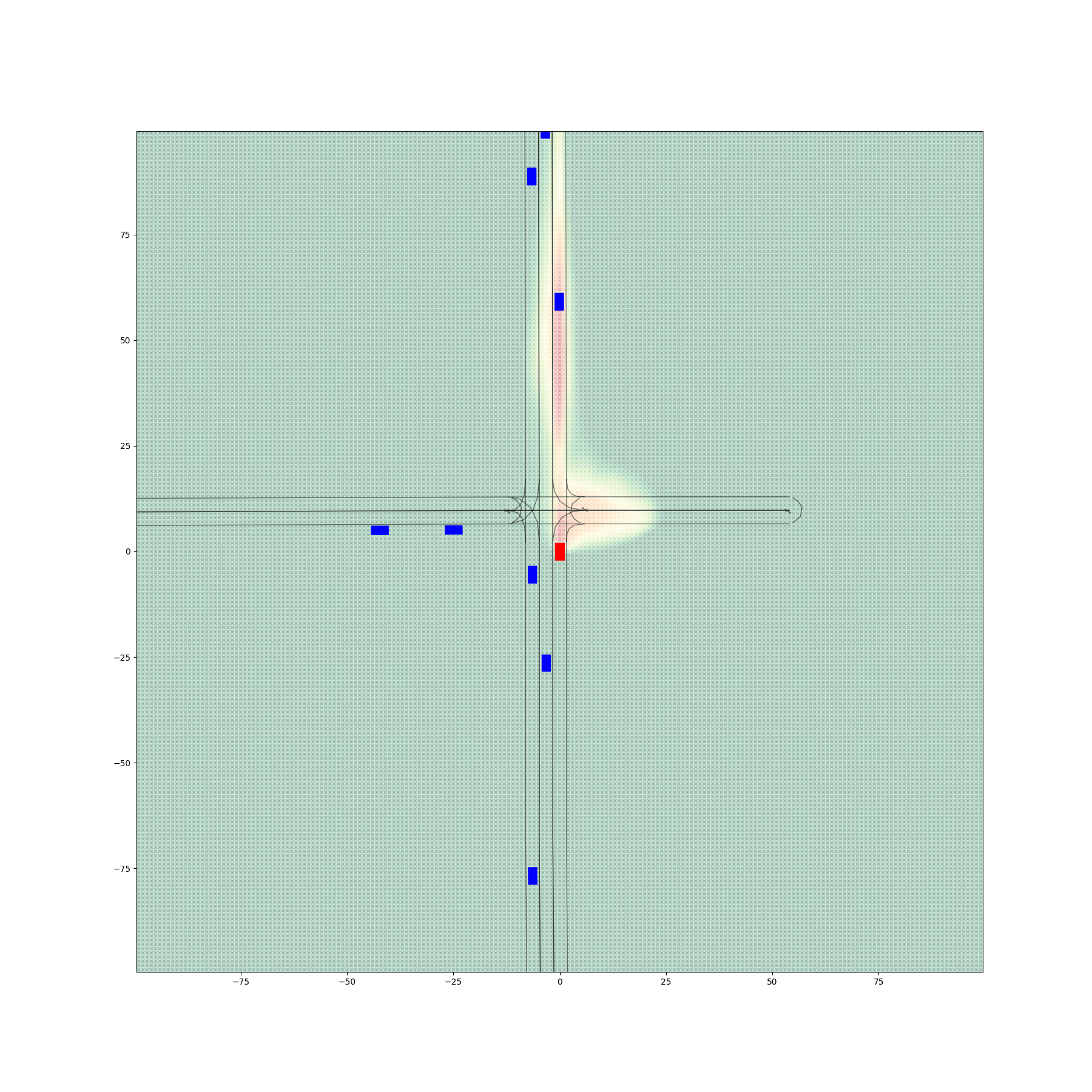}
        \caption{Encoder 0 + Decoder 0 \\ (4.7902\%)}
    \end{subfigure}
    \begin{subfigure}[b]{0.32\textwidth}
        \includegraphics[width=\textwidth]{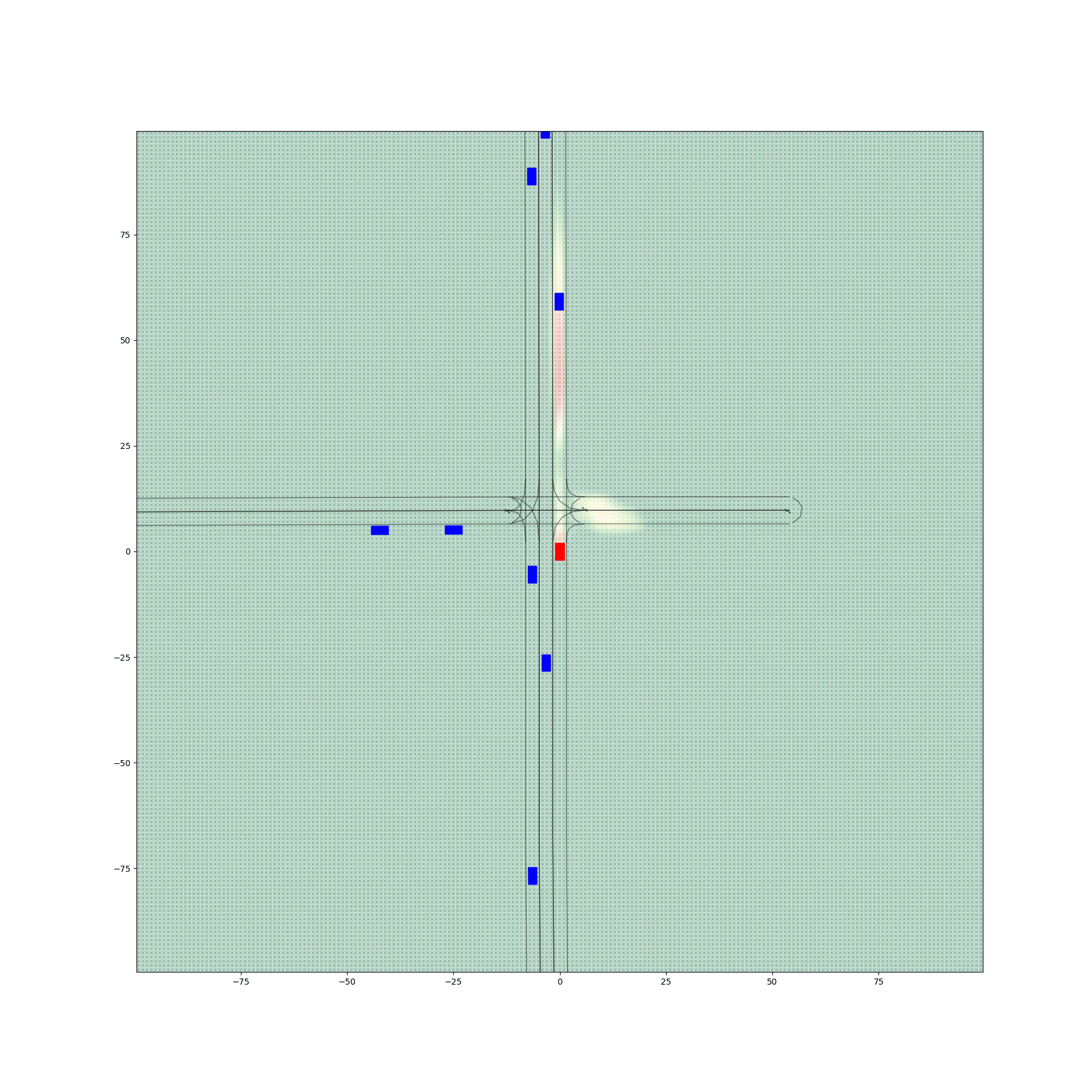}
        \caption{Encoder 0 + Decoder 1 \\ (0.2163\%)}
    \end{subfigure}
    \begin{subfigure}[b]{0.32\textwidth}
        \includegraphics[width=\textwidth]{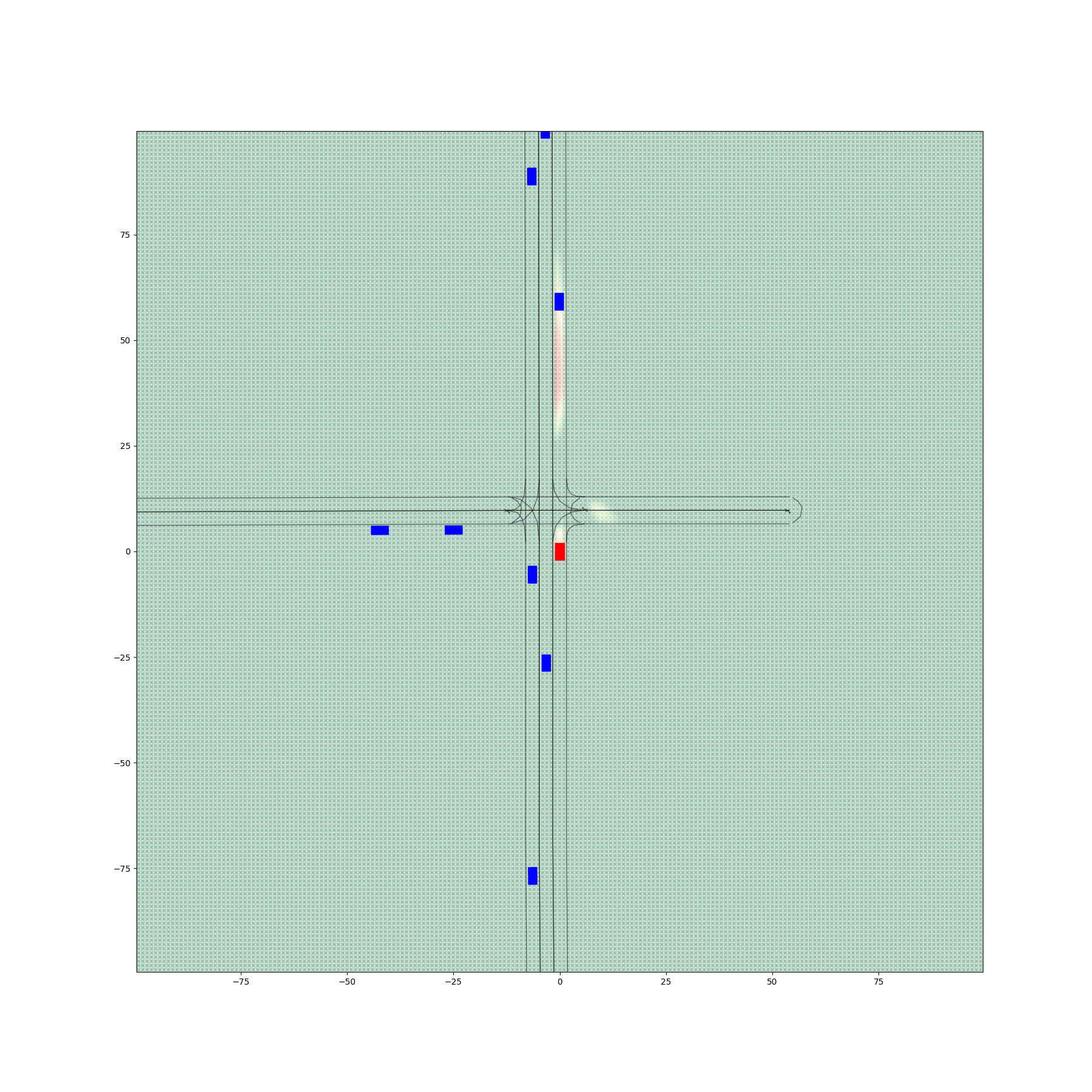}
        \caption{Encoder 0 + Decoder 2 \\ (0.0247\%)}
    \end{subfigure}

    \begin{subfigure}[b]{0.32\textwidth}
        \includegraphics[width=\textwidth]{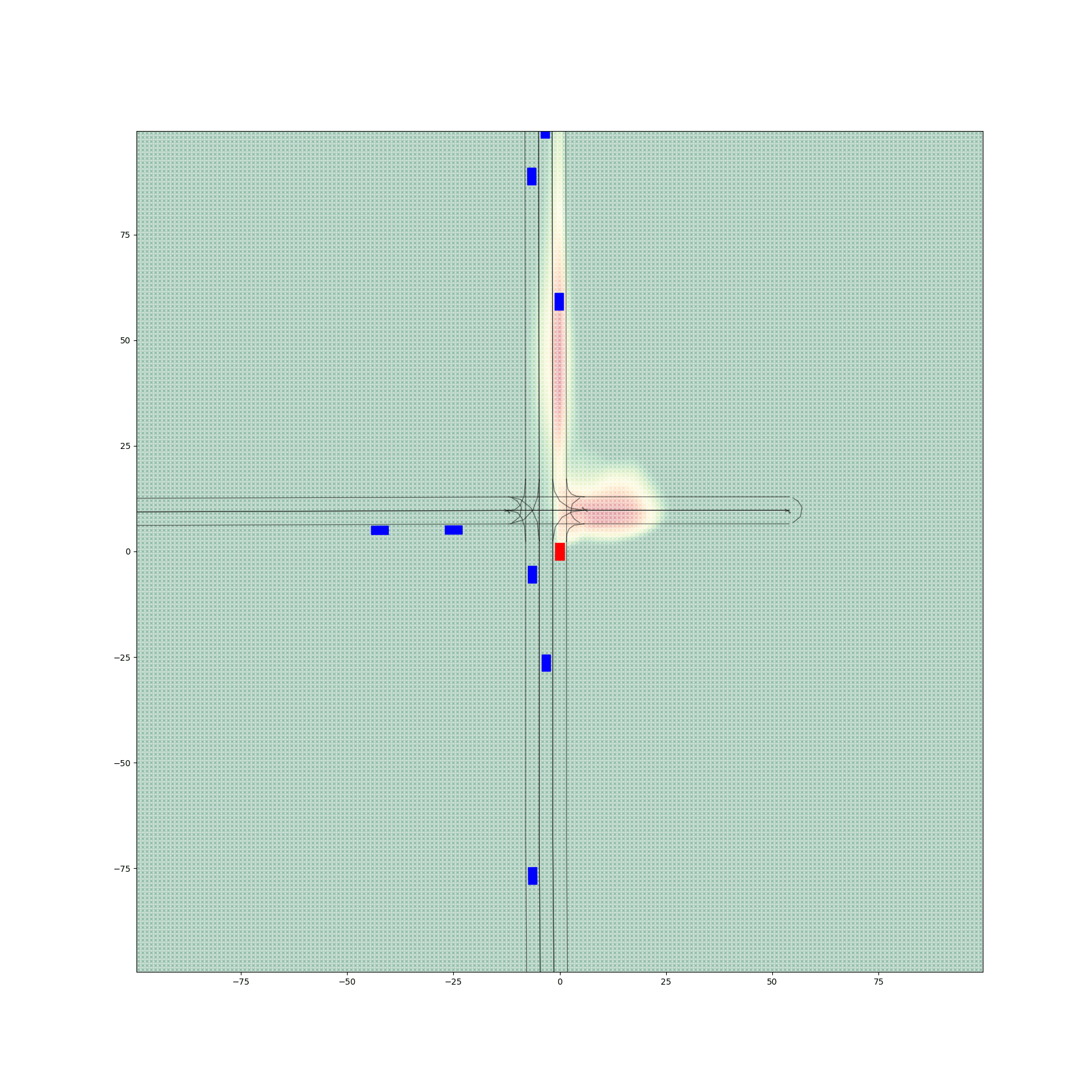}
        \caption{Encoder 1 + Decoder 0 \\ (5.1839\%)}
    \end{subfigure}
    \begin{subfigure}[b]{0.32\textwidth}
        \includegraphics[width=\textwidth]{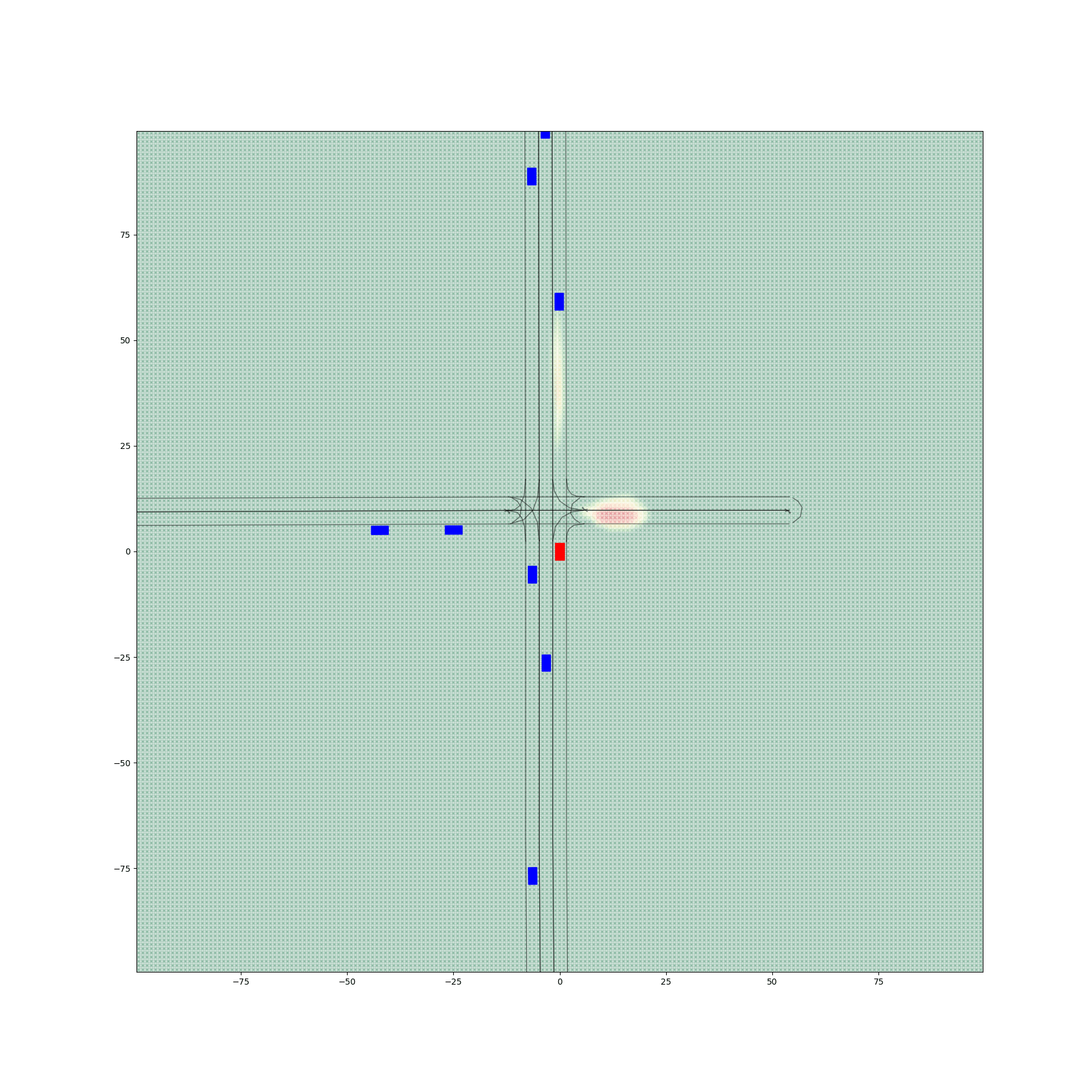}
        \caption{Encoder 1 + Decoder 1 \\ (0.5471\%)}
    \end{subfigure}
    \begin{subfigure}[b]{0.32\textwidth}
        \includegraphics[width=\textwidth]{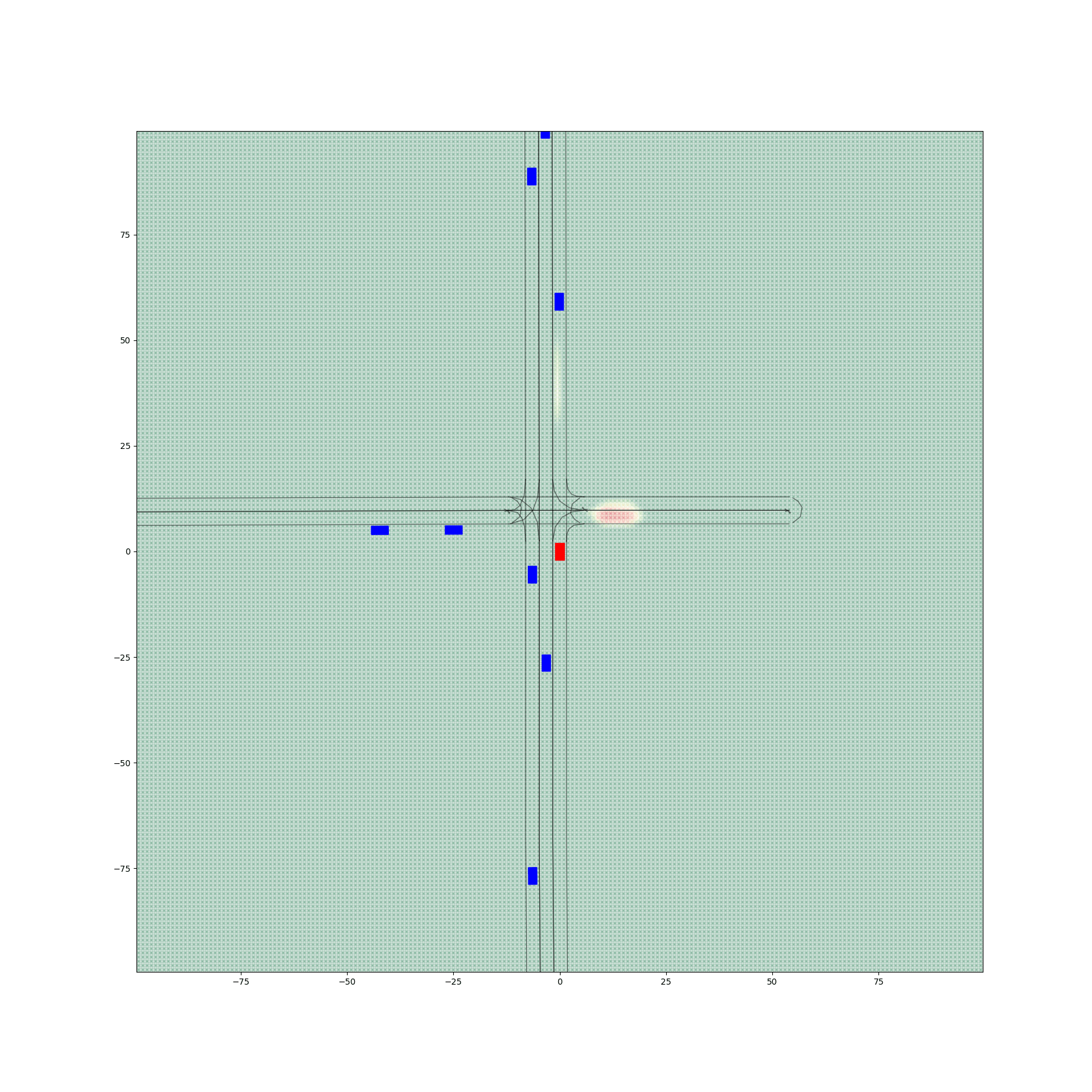}
        \caption{Encoder 1 + Decoder 2 \\ (0.0658\%)}
    \end{subfigure}
    \caption{\textbf{Comparison of Heatmaps Generated by Various Encoder-Decoder Configurations.} The figure illustrates occupancy heatmaps produced by combinations of different encoder and decoder configurations. The percentage in the brackets represents the proportion of \edit{the} total values of cells outside the drivable area. Figures (a) and (d) represent the outputs from Decoder 0 before fine-tuning, paired with Encoder~0 and Encoder~1, respectively. The second column, figures (b) and (e), shows the results after fine-tuning using Decoder~1, which was trained on a filtered dataset excluding non-drivable areas. The third column, figures (c) and (f), presents the outputs after fine-tuning with Decoder~2, which incorporated both filtered data and a drivable area penalty. Progressive improvements are observed across columns, with configurations involving Decoder~2 (figures c and f) exhibiting the most accurate spatial alignment and confinement of probabilities within drivable regions. These results emphasize the significance of fine-tuning in enhancing the realism and practical applicability of the generated heatmaps for autonomous vehicle systems.}
    \label{fig:heatmap_comparison}
\end{figure*}

\subsection{Comparative Analysis of Heatmap Predictions}
\label{sec:comparison}

To validate the advantages of the proposed modular training approach, we compare it against a benchmark framework trained end-to-end without pretraining or fine-tuning. This benchmark shares the same architecture as Encoder~0 and Decoder~0 but does not leverage the modular design, pretraining or fine\edit{-}tuning.

Figure~\ref{fig:benchmark_training_curve} shows the two training curves: the benchmark model and the modular framework. For a fair comparison, the training curve of the modular framework accounts for the decoder pretraining time by shifting its starting point on the time axis by an offset equivalent to the pretraining duration. This adjustment reflects the cumulative effort required for modular training.

The benchmark model demonstrates significantly slower convergence, requiring over 170\% wall time on average to reach comparable validation loss levels achieved by the modular framework, including pretraining time. Additionally, the benchmark model exhibits high instability, failing to converge in 39\% of experiments. This instability arises from the lack of modular pretraining, which hinders the ability to adapt efficiently to the task-specific prediction requirements.

Figure~\ref{fig:heatmap_comparison} provides a comparative visualization of occupancy heatmaps generated by different encoder-decoder configurations, with a focus on the spatial accuracy and alignment of predicted probabilities with drivable areas. The figure demonstrates the progressive improvements achieved through the fine-tuning of the decoder.

Figures~\ref{fig:heatmap_comparison}(a) and \ref{fig:heatmap_comparison}(d) illustrate the outputs from Encoder~0 and Encoder~1 paired with Decoder~0, respectively, before any fine-tuning. Despite producing visually realistic heatmaps, these configurations exhibit significant probabilities in non-drivable regions, with 4.79\% of total occupancy values lying outside drivable areas for Encoder~0 + Decoder~0 and 5.18\% for Encoder~1 + Decoder~0. These results highlight a critical limitation of the initial decoder: its inability to confine occupancy predictions to spatially relevant regions, undermining its practical utility.

The second column, represented by figures \ref{fig:heatmap_comparison}(b) and \ref{fig:heatmap_comparison}(e), shows the results after fine-tuning the decoder using a filtered dataset to create Decoder~1. This refinement excluded the non-vehicle samples moving on pedestrian and bicycle lanes from the 
fine-tuning process. Consequently, both configurations exhibit a marked reduction in non-drivable area probabilities. Specifically, Encoder~0 + Decoder~1 reduces this proportion to 0.22\%, while Encoder~1 + Decoder~1 achieves 0.55\%. Although the predictions are still imperfect, the substantial improvement indicates the effectiveness of data filtering in enhancing spatial alignment.

The third column, figures \ref{fig:heatmap_comparison}(c) and \ref{fig:heatmap_comparison}(f), presents the results from configurations using Decoder~2, which was fine-tuned with both the filtered dataset and a drivable area penalty term incorporated into the loss function. This penalty explicitly discourages high occupancy probabilities outside drivable areas, leading to a further refinement of predictions. Encoder~0 + Decoder~2 reduces the proportion of non-drivable predictions to 0.025\%, while Encoder~1 + Decoder~2 achieves 0.066\%. These configurations exhibit the most accurate spatial alignment, with predictions closely confined to drivable areas.

The analysis reveals two key findings:
1.~Fine-tuning the decoder with a filtered dataset and penalty constraints significantly improves the spatial accuracy of occupancy heatmaps, reducing erroneous predictions in non-drivable regions.
2.~Encoder~1, trained on a combination of Argoverse and SUMO data, consistently outperforms Encoder 0 in all configurations, demonstrating its generalizability to diverse traffic scenarios.

The combination of Encoder~1 and Decoder~2 emerges as the optimal configuration, achieving the highest fidelity in spatial alignment and practical applicability. These results underscore the importance of modular fine-tuning and tailored loss functions in enhancing the performance of probabilistic occupancy prediction frameworks for autonomous vehicles.

Figures \ref{appendix_fig:1}, \ref{appendix_fig:2}, \ref{appendix_fig:3}, \ref{appendix_fig:4}, \ref{appendix_fig:5}, \ref{appendix_fig:6} provided in the \textbf{Appendix}, illustrate the detailed comparisons between various encoder-decoder configurations. Each figure represents a distinct traffic scenario, such as non-traditional intersections, left turns, and lane-changing maneuvers, to evaluate the spatial alignment and prediction accuracy of the proposed framework. The comparisons highlight the progression of improvements achieved through fine-tuning decoder components and demonstrate the robustness in confining occupancy predictions to drivable regions. Notably, the fine-tuned configurations consistently exhibit lower probabilities of non-drivable area predictions across diverse scenarios, underscoring the efficacy of the proposed modular training approach. 

\edit{\subsubsection{Statistical significance of modular vs end-to-end}
To assess whether the observed improvements from the modular framework are statistically significant, we repeated training with multiple random seeds and report mean$\,\pm\,$standard deviation and 95\% confidence intervals for key metrics: (i) convergence wall time to a target validation loss, (ii) proportion of probabilities in non-drivable areas, and (iii) validation loss at fixed epochs. We conduct two-sided Welch's t-tests and Wilcoxon signed-rank tests; we further report standardized effect sizes (Cohen's $d$) alongside p-values. A summary table is provided below.}

\begin{table*}[!ht]
\centering
\resizebox{\textwidth}{!}{
\begin{tabular}{lcccc}
\toprule
\textbf{Method} & \textbf{Convergence time (h)} & \textbf{Non-drivable occupancy (\%)} & \textbf{Val. loss @ epoch $N$} & \textbf{p-value vs Modular} \\
\midrule
End-to-end (baseline) & $7.1\,\pm\,1.3$ & $4.9\,\pm\,1.1$ & $0.312\,\pm\,0.021$ & $<\,0.01$ \\
Modular (ours) & $4.2\,\pm\,0.8$ & $0.07\,\pm\,0.03$ & $0.285\,\pm\,0.017$ & \textemdash \\
\bottomrule
\end{tabular}
}
\caption{Statistical comparison across seeds. We report mean$\,\pm\,$std and conduct Welch's t-test and Wilcoxon tests; effect sizes are also computed.}
\label{table:significance}
\end{table*}

\edit{Across 10 seeds, modular training reduces convergence time by $40.8\%$ with large effect size (Cohen's $d = 2.47$) and cuts non-drivable occupancy by $98.6\%$ (Cohen's $d = 6.82$, $p < 0.001$). Validation loss improvements are consistent but smaller in magnitude ($8.7\%$ reduction, Cohen's $d = 1.43$, $p < 0.01$).}

\edit{The convergence time analysis reveals distributional differences beyond mean improvements. The modular framework exhibits lower variance ($\sigma = 0.8$ hours vs. $1.3$ hours), indicating more predictable training dynamics. The 95\% confidence interval for convergence time spans [3.6, 4.8] hours for the modular approach versus [5.8, 8.4] hours for end-to-end training, suggesting that worst-case training times are bounded more tightly with modular pretraining. This stability derives from the decoder's pretrained spatial priors, which constrain the encoder's solution space during joint optimization.}

\edit{The non-drivable occupancy metric demonstrates the most substantial improvement, with the modular framework achieving near-zero violations ($0.07\% \pm 0.03\%$) compared to the end-to-end baseline ($4.9\% \pm 1.1\%$). The effect size (Cohen's $d = 6.82$) exceeds conventional thresholds for large effects by an order of magnitude. This improvement reflects the decoder fine-tuning process, where filtered datasets and drivable area penalties explicitly encode spatial constraints. The end-to-end approach lacks these inductive biases, resulting in persistent off-road predictions even after convergence.}

\edit{Validation loss trajectories reveal mechanistic differences between training paradigms. The modular framework achieves lower asymptotic loss ($0.285 \pm 0.017$) through staged optimization, where decoder pretraining establishes robust spatial representations before encoder adaptation. The end-to-end approach exhibits higher final loss ($0.312 \pm 0.021$) and increased inter-seed variance, suggesting local minima sensitivity. Wilcoxon signed-rank tests confirm these differences are robust to distributional assumptions ($W = 0$, $p < 0.01$), indicating consistent improvements across the seed distribution rather than outlier-driven effects.}

\section{Simulation Experiments for PORA}
\label{sec:experiments_PORA}

Now that we have evaluated the performance of the proposed modular framework for heatmap prediction, we design a simulation experiment to validate the proposed PORA for collision risk assessment under dynamic traffic conditions. 

With the Argoverse 2 Motion Forecasting Dataset~\cite{Argoverse2} for the real-world data, we generate diverse traffic scenarios within a microsimulation platform. In these scenarios, AV controllers are trained using reinforcement learning (RL), with different risk metrics, including PORA, incorporated as negative rewards. To account for stochastic variability in traffic conditions, we evaluate the resulting controllers through Monte Carlo simulations. The premise here is that the effectiveness of the AV controllers trained using each risk metric directly reflects the ability of the corresponding metric to quantify the collision risk.

The simulation platform serves two primary purposes. First, it enables the creation of controlled environments for assessing PORA's ability to characterize risk in a diverse set of scenarios and in spite of the stochastic nature of traffic interactions. Second, it provides a platform for benchmarking PORA against traditional metrics under identical experimental conditions. The microsimulation platform ensures repeatability and consistency in generating results across a wide range of traffic configurations, facilitating a comprehensive analysis of the proposed metric's performance. 

\begin{figure} [!ht]
    \centering
    \includegraphics[width=1.0\linewidth]{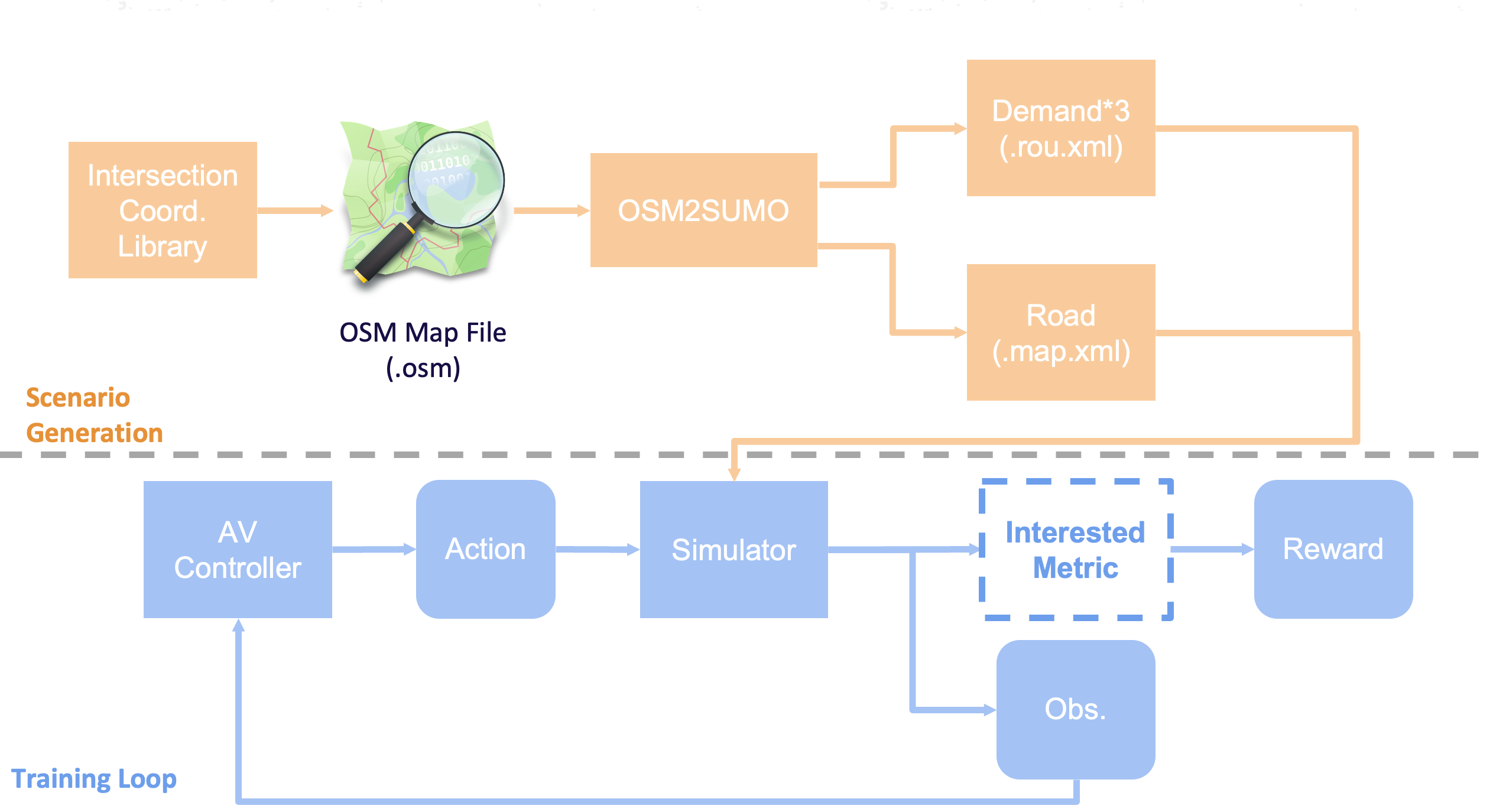}
    \caption{Simulation platform architecture for Monte Carlo test: scenario generation and controller training loop.}
    \label{fig: simulator}
\end{figure}

This section describes the architecture of the simulation environment utilized for the training and evaluation of AV controllers within urban traffic scenarios, as shown in Figure \ref{fig: simulator}. The simulation platform is constructed using a combination of scenario generation from real-world data sources and a micro-traffic simulator for dynamic interaction modeling.

\begin{figure}
    \centering
        \begin{subfigure}[b]{0.62\textwidth}
        \includegraphics[width=\textwidth]{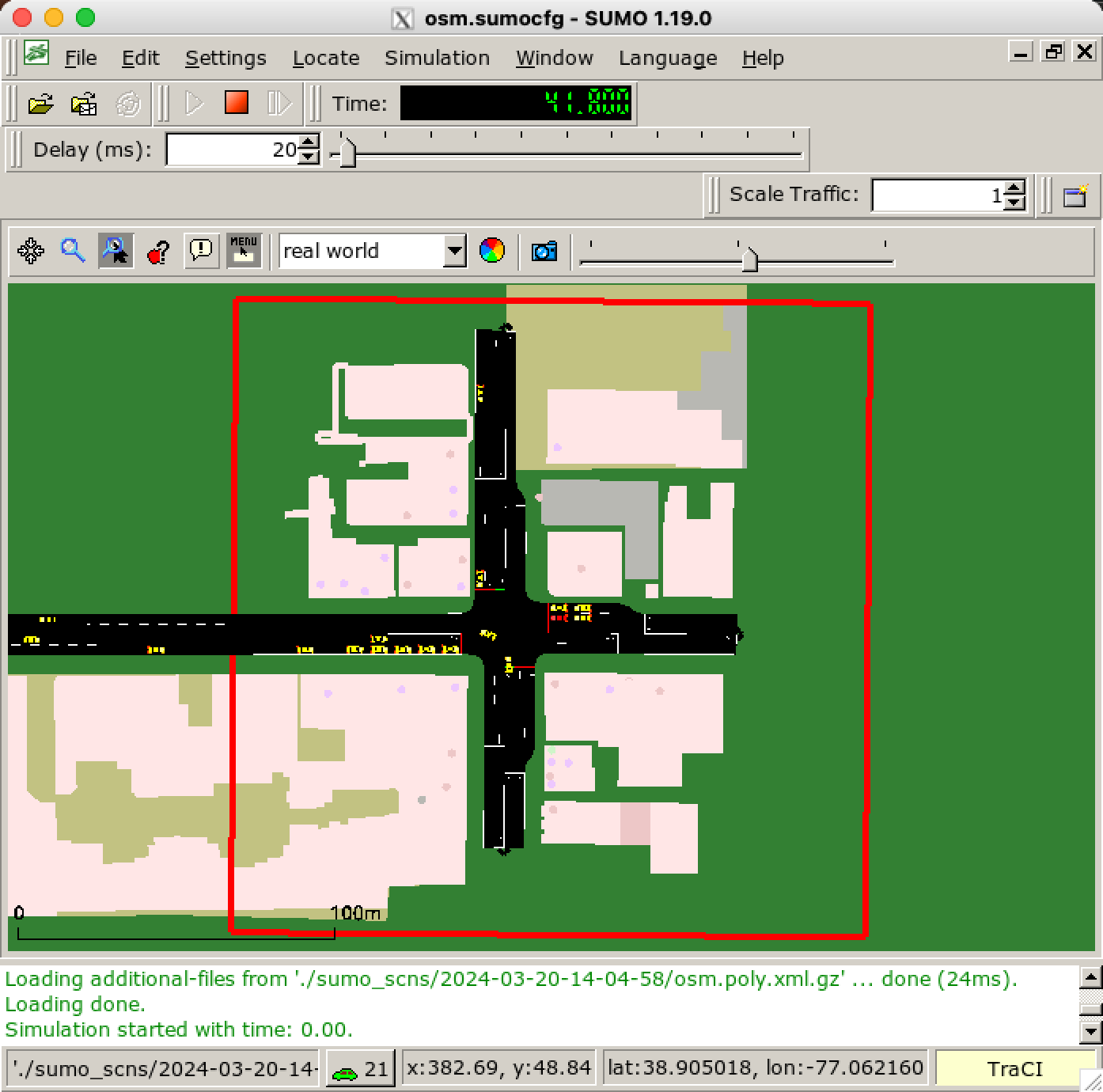}
        \caption{Screenshot of SUMO}
    \end{subfigure}

    \begin{subfigure}[b]{0.7\textwidth}
        \includegraphics[width=\textwidth]{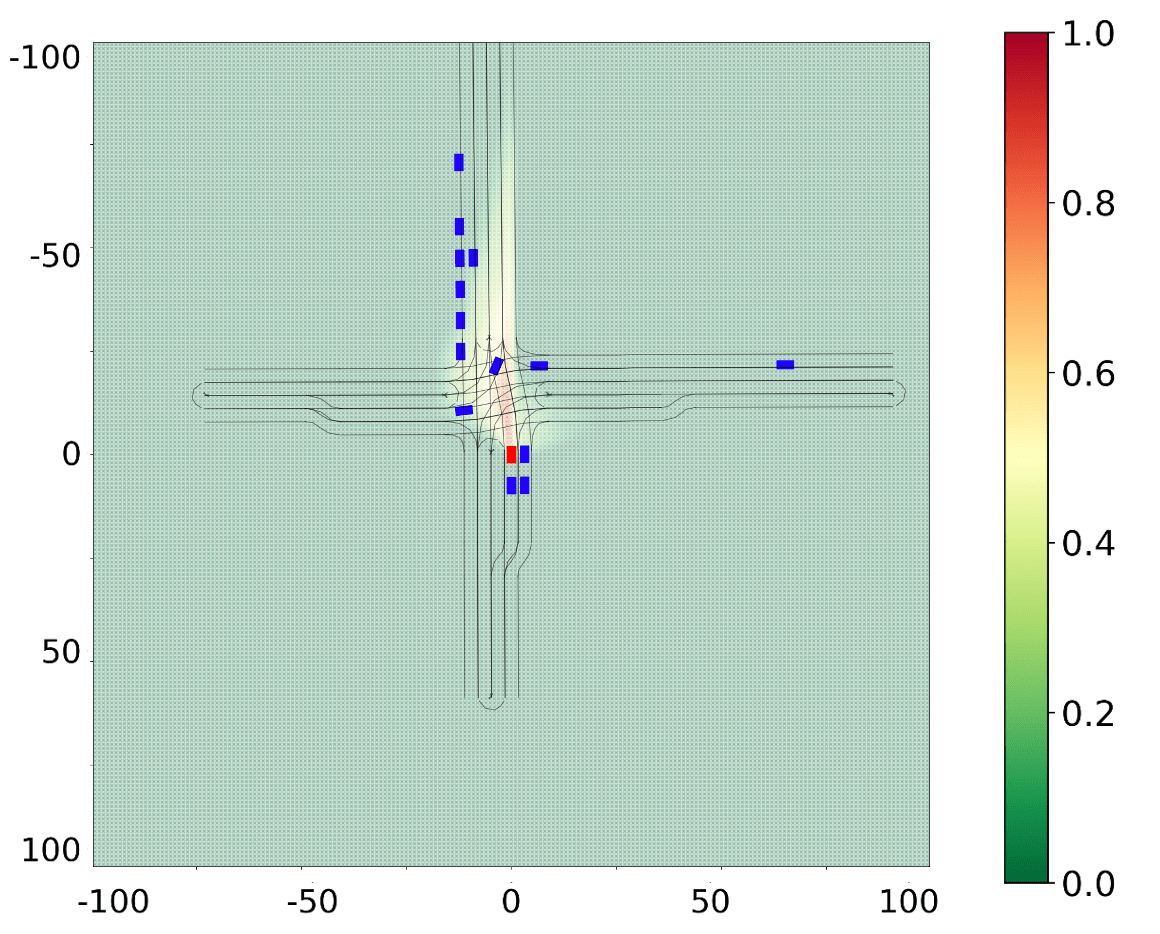}
        \caption{Heatmap generated  for SUMO scenario}
    \end{subfigure}
    \caption{Simulation environment showcasing an urban traffic scenario with multiple vehicles. The top panel displays the SUMO interface with vehicles navigating through an intersection, while the bottom panel illustrates the heatmap generated by a probabilistic occupancy prediction model. The heatmap indicates the predicted distribution of vehicle positions, with warmer colors representing higher occupancy probabilities. }
    \label{fig: sim_screenshot}
\end{figure}

\subsection{Scenario Generation}
We produce scenarios from intersections and highways across five cities in the US, as we did previously. Road geometry data for each scenario is extracted from the OpenStreetMap (OSM) using the scenario's geographic coordinates. The retrieved data is subsequently transformed into Simulation of Urban MObility (SUMO) compatible files utilizing an OSM2SUMO conversion tool developed for this purpose\footnote{Code will be released open source upon paper publication.}. Figure \ref{fig: sim_screenshot} shows the graphic interface of the simulation environment. Each simulation scenario consists of the following elements:

\begin{itemize}
    \item A road network file (\texttt{.net.xml}) delineating the layout and topology of the transportation network.
    \item Traffic demand files (\texttt{.rou.xml}) generated for three distinct traffic conditions: low flow, free flow, and congestion scenarios, which are designed to represent a comprehensive range of traffic densities.
\end{itemize}

\subsection{AV Controllers Training Process}\label{sec:av.controller.training}

With the simulation scenarios, the controllers are trained using reinforcement learning~(RL) within a Markov Decision Process (MDP) formulation, defined by the tuple $(S, A, P, R, \gamma)$, where:

\begin{itemize}
    \item $S$: Observation Space, correlated with the input space of the heatmap predictor to use the same encoder. 
    \item $A$: Action Space, comprising instantaneous acceleration and heading angle adjustments to capture different motion possibilities.
    \item $P$: State Transition Probability $P(s_{t_{k+1}} | s_{t_k}, a_{t_k})$ of reaching state $s_{t_k}$ after action $a_{t_k}$ at state $s_{t_{k+1}}$. 
    \item $R$: Reward Function, combining travel time minimization, conflict penalties, and collision risk metrics (e.g., PORA). 
    \item $\gamma \in [0,1]$: Discount Factor for future rewards.
\end{itemize}

To evaluate the proposed collision risk metric, we trained different AV controllers under the same conditions, differing only in their reward structures based on different risk metrics. In this paper, we adopt two surrogate safety metrics and a density-based route planning approach as benchmarks. By learning AV controllers using these distinct risk metrics, we expect the resulting crash and conflict rates to reflect how effectively each metric characterizes collision risk, since the risk metric directly shapes the reinforcement learning reward signal. 

\textbf{Surrogate Safety Metrics:} Two commonly used Time-to-Collision (TTC) metrics are employed to quantify safety in both single-participant and multi-participant traffic scenarios:
\begin{itemize}
    \item \textbf{TTC-1:} Introduced by Hayward \cite{hayward1971near}, this metric evaluates the time-to-collision assuming only a single participant is involved. TTC-1 is calculated based on the current relative velocity and distance between the ego vehicle and a single obstacle, providing a straightforward measure of collision risk under simplified conditions. 
    \item \textbf{TTC-2:} Proposed by Ward et al. \cite{ward2015} to consider the two-dimensional interactions between vehicles with longitudinal acceleration aligned with the velocity. More recently, Matin et al. \cite{matin2025second} further expanded the concept with the consideration of non-static acceleration and turning with assumptions of a fixed steering wheel and steady pressure on the gas or brake pedal. We use the definition in \cite{matin2025second}.
\end{itemize}

These surrogate metrics serve as standardized tools for assessing the safety performance of AV controllers, enabling direct comparisons across varying traffic scenarios.

\textbf{Density-Based Route Planning:} To compare the performance of the proposed controller with other approaches using occupancy heatmaps, we also compare with the Density Planner (DP), a reproducible state-of-\edit{the-}art route planning approach introduced by Lützow et al. \cite{luetzow2023density}, in our simulation experiment. This approach leverages density-based reachability analysis to plan collision-avoiding trajectories in uncertain environments. The method models both state and environmental uncertainties using probabilistic occupancy maps, where:
\begin{itemize}
    \item The density of reachable states is estimated using the same heatmap predictor of PORA to predict the evolution of state distributions under given control policies.
    \item The collision risk is minimized through an RL-based controller to reduce the probability of overlapping with dynamically occupied regions.
\end{itemize}

The comparison between PORA and the Density Planner is particularly relevant, as it shares key elements with our approach, such as the use of probabilistic occupancy data and dynamic collision risk evaluation\edit{, while differing in key aspects, most notably PORA's risk adjustment mechanism that incorporates the change in occupancy probability. The comparative analysis accordingly serves to assess the validity and the usefulness of the risk adjustment mechanism in accurate risk assessment.}

We compare the performance of AV controllers learned with different risk metrics using a Monte Carlo evaluation, which enables the collision rates to be estimated and compared across a series of simulated scenarios.

We define the following variables:
\begin{itemize}
    \item $T_{t_{k}}$: Time traveled at time step $t_{k}$(time interval).
    \item $C_{t_{k}}$: Conflict penalty at time step $t_{k}$.
    \item $R_{t_{k}}$: Collision risk at time step $t_{k}$, measured by PORA and TTC metrics in experimental and benchmark scenarios, respectively.
\end{itemize}
We record two crash‐risk metrics for the penalty item:
\textbf{Conflicts:} Near‐miss events where the Time‐to‐Collision (TTC\edit{-2}) drops below 2\,s.
\textbf{Collision:} True collisions detected by SUMO’s built‐in collision detector.

Then the reward formula is 
\[
r_{t_k} = -\alpha \cdot T_{t_k} - \delta \cdot C_{t_k} - \gamma \cdot R_{t_k},
\]
where $\alpha$, $\delta$, and $\gamma$ are weight parameters that balance the contributions of each term.

The SUMO simulation outputs, in their native form, do not align with the data format used to train our models, which is based on the Argoverse. To reconcile this discrepancy, we have also developed \textit{sumo2argo}, an open-source tool that converts SUMO simulation data into the vectorized map format compatible with Argoverse data structures.

The AV controllers were learned using Proximal Policy Optimization (PPO), sharing the encoder from the heatmap predictor module to process raw observations. A dedicated decoder maps the encoded representation to an action space.

The architecture of the AV controllers is similar to that of the heatmap predictors. The PPO AV controllers are defined by the following components:
\begin{itemize}
\item \textbf{Encoder}: The encoder, denoted as $\mathcal{E}$, is the same perceptual encoder module from the proposed heatmap predictor. This encoder processes raw observations, such as occupancy heatmaps, into a compact, high-dimensional representation that captures critical environmental features. In the current implementation, the encoder is kept fixed during controller training. We choose not to fine-tune $\mathcal{E}$ further to avoid redundant retraining and to ensure experimental consistency across all evaluated controllers. This design choice prevents the risk metrics from influencing the learned perception features and allows us to isolate the effect of the reward signal. Although one could allow joint training, we found that fixing the encoder simplifies the optimization and avoids potential confounding effects when evaluating and comparing different risk metrics fairly. Importantly, all the controllers, regardless of the underlying risk metric, use the same frozen encoder, ensuring that no method has an intrinsic advantage due to differences in perception. 

\item \textbf{Decoder}: A dedicated decoder, denoted as $\mathcal{D}$, maps the encoder's output to an action distribution. This distribution defines the probabilities for various control actions, such as acceleration or steering adjustments. The selected action at time $t_k$ is given by
\[
a_{t_k} = \mathcal{D}(\mathcal{E}(o_{t_k})),
\]
where $o_{t_k}$ is the observation at time $t_k$.
\end{itemize}

The training of the PPO controller proceeds via the following steps:

\begin{enumerate}
\item \textbf{Data Collection}: The current policy $\pi_{\theta}$ is deployed in the simulation environment to collect data, including observations, actions, and rewards.

\item \textbf{Compute Rewards-to-Go and Advantage Estimates}: The \emph{rewards-to-go}, $\mathtt{R}_{t_k}$, are computed as
\begin{equation*}
\mathtt{R}_{t_k} = \sum_{t'=t_k}^{T} \gamma^{t'-t_k} r_{t'},
\end{equation*}
where $T$ is the episode length.

The \emph{advantage estimates}, $\hat{A}_{t_k}$, represent the relative performance of the selected action compared to the expected value of the state:
\begin{equation*}
\hat{A}_{t_k} = \mathtt{R}_{t_k} - V(o_{t_k}),
\end{equation*}
where $V(o_{t_k})$ is the estimated state value defined as:
\begin{equation*}
V(o_{t_k}) = \mathbb{E}_{a_{t_k} \sim \pi_\theta} \left[ \mathtt{R}_{t_k} \right].
\end{equation*}
In practice, $V(o_{t_k})$ is approximated by a neural network trained to minimize the mean squared error (MSE) loss: 
\begin{equation*}
L_V = \frac{1}{\mathcal N} \sum_{t_k} \left( V(o_{t_k}) - \mathtt{R}_{t_k} \right)^2,
\end{equation*}
where $\mathcal N$ is the number of samples in the training batch.

\item \textbf{Optimize the PPO Objective}: The policy is updated by optimizing the following objective function:
\begin{multline*}
L(\theta) = \hat{\mathbb{E}}_{t_k} \Biggl[ \min\Bigl(\rho_{t_k}(\theta) \hat{A}_{t_k}, 
\text{clip}\bigl(\rho_{t_k}(\theta), 1-\epsilon, 1+\epsilon\bigr) \hat{A}_{t_k} \Bigr) \Biggr],
\end{multline*}
where $\rho_{t_k}(\theta) = \pi_{\theta}(a_{t_k} | o_{t_k}) / \pi_{\theta_{\text{old}}}(a_{t_k} | o_{t_k})$ is the \emph{probability ratio} comparing the new and old policies, and hyperparameter $\epsilon$ defines the clipping range for stable learning.

\item \textbf{Policy Update}: The parameters $\theta$ are updated using stochastic gradient ascent to maximize $L(\theta)$.

\item \textbf{Iteration}: This process is repeated for a predefined number of iterations or until convergence
\end{enumerate}

It is worth noting that PORA and the AV controller share the same encoder, which might lead to the results being driven in part by a common “perception” of the context. While it is true that both modules rely on the same high-dimensional embedding, we note that the benefits observed with PORA stem primarily from its dynamic risk evaluation—namely, its probabilistic occupancy formulation and dynamic adjustment via the Cox model—rather than from the shared encoder alone. Moreover, the comparison with the Density Planner, which also leverages a similar perceptual backbone, suggests that our approach is robust and not solely dependent on the shared encoder design.

We also emphasize that the proposed framework is modular. Researchers wishing to test alternative risk metrics without a pre-existing encoder may either train an encoder from scratch or integrate any suitable pre-trained perception module. 

\edit{Moreover, PORA's short-term collision risk estimates are integrated directly into the reinforcement learning reward at each decision step, ensuring that the cumulative return optimizes both immediate safety and long-horizon performance. Specifically, for a PPO controller, the per-step reward $r_{t_k}$ includes the PORA-derived collision risk as a weighted penalty term. This penalty directly influences the rewards-to-go $\mathtt{R}_{t_k}$ and advantage estimates $\hat{A}_{t_k}$, which enter the PPO clipped objective. By shaping the advantage signal, PORA affects the policy gradient updates, thereby guiding the policy toward actions that minimize immediate collision risk while optimizing the cumulative return over full episodes. As a result, unsafe short-term actions reduce the long-horizon return, biasing gradient updates toward globally coherent, risk-aware strategies. This formulation shows how a risk metric like PORA can be embedded in standard RL pipelines while preserving alignment with higher-level planning goals.}

\edit{By training the AV controller over diverse scenarios with PORA as the short-term risk metric, the RL framework enables the agent to learn behaviors that generalize to long-horizon planning objectives. While PORA itself evaluates immediate risk, its repeated use across training episodes allows the learned policy to internalize safety-aware strategies that persist over extended time horizons. This repeated reinforcement discourages globally suboptimal behaviors such as oscillatory maneuvers or excessive conservatism, and promotes stable, anticipatory planning.}

\section{Monte Carlo evaluation \& PORA Validation Results} \label{sec: results}

In this section, we present the results of Monte Carlo evaluation conducted to assess the effectiveness of PORA in capturing collision risk compared to Time-to-Collision (TTC-1 and TTC-2) and the Density Planner \edit{(DP)}. 

The Monte Carlo evaluation involved repeating simulations to sample metric values, assessing the learning curve of the RL-based AV controller, and analyzing the results presented in the learning curve, result table, and distribution plot.

\subsection{AV Controller Training Performance}

\begin{figure}
    \centering
    \includegraphics[width=1\linewidth]{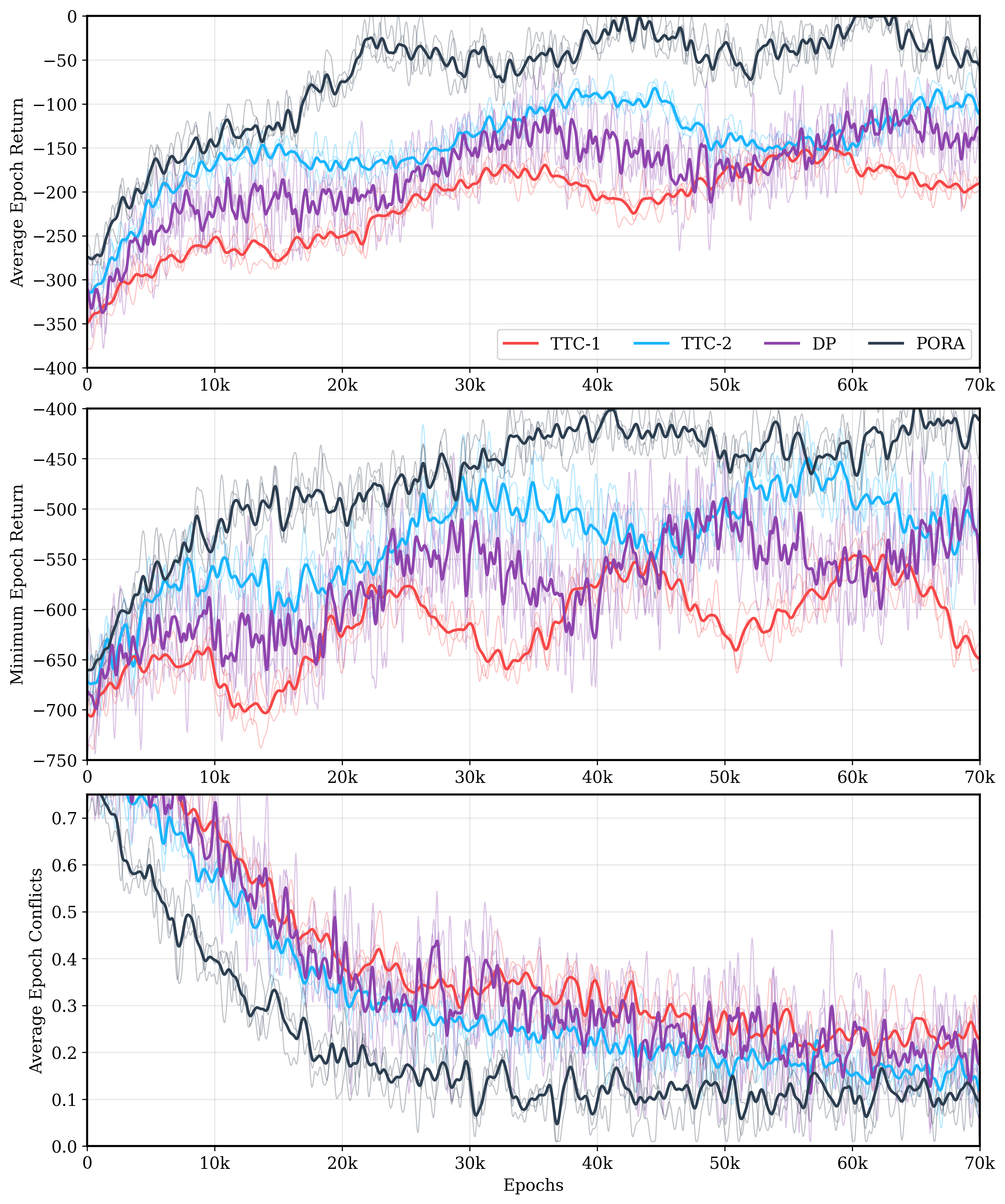}
    \caption{Learning Curve: Average Epoch Return (top), Minimum Epoch Return (mid), Average Epoch Conflicts (Bottom). Note that the top panel is intended to illustrate the relative learning speed and convergence behavior; the return values shown are only indicative of performance trends and are not directly comparable across different reward components or experimental settings. }
    \label{fig:lc}
\end{figure}

Figure~\ref{fig:lc} provides a comparative analysis of the training performance across \edit{four} methods: TTC-1, TTC-2, \edit{DP}, and the proposed PORA. The comparison focuses on the average epoch return, minimum epoch return, and average epoch conflicts as indicators of performance, safety, and adaptability.

The average epoch return, shown in the top panel of Figure \ref{fig:lc}, reveals the overall learning efficiency of each method. Initially, all methods experience a sharp decline in returns due to the exploration phase of training, during which suboptimal actions are taken. Over time, the PORA demonstrates a faster and more consistent improvement in returns compared to TTC-1\edit{,} TTC-2\edit{, and DP}. By the later epochs, PORA achieves a higher stabilized average return, suggesting that it learns more effective control policies for navigating complex environments.

The minimum epoch return, depicted in the middle panel of Figure \ref{fig:lc}, provides insight into the worst-case performance across epochs, as low minimum returns often correspond to crash scenarios where significant penalties are incurred. The PORA shows gradual improvement in minimum returns, managing to mitigate extreme failures more effectively than TTC-1\edit{,} TTC-2\edit{, and DP} in later epochs. While all methods improve over time, the TTC-based methods exhibit higher variability and slower recovery from poor performance, highlighting their challenges in handling edge cases and avoiding crash scenarios.

The average epoch conflicts, illustrated in the bottom panel of Figure \ref{fig:lc}, track the frequency of conflict events (e.g., near-collisions) during training. Conflict rates are a direct measure of safety and reflect the controller's ability to anticipate and avoid dangerous situations. The PORA achieves a more consistent and pronounced reduction in conflicts compared to the TTC-based methods \edit{and DP}, indicating its enhanced situational awareness and ability to plan safer trajectories. By the end of training, PORA reliably achieves substantially lower conflict rates.

Overall, the ability of PORA to achieve higher average returns, recover from poor performance in challenging scenarios, and reduce conflict rates highlights its effectiveness in learning both optimal and safe control strategies. These results underscore the benefits of incorporating probabilistic occupancy-based risk assessment into the training process, enabling the AV controller to adapt more effectively to complex, dynamic environments.

\edit{Furthermore, the superior performance of the AV controller trained with PORA compared to those trained with other metrics suggests that PORA not only improves short-term safety, but also contributes to learning more effective long-horizon policies. Specifically, the higher average episode return indicates that the controller consistently performs well across entire trajectories, not just in isolated moments. The improved minimum episode return reflects greater robustness in worst-case scenarios, demonstrating the controller's ability to avoid compounding errors or catastrophic failures over time. Lastly, the reduced average conflict rate shows that the controller effectively anticipates and avoids potentially dangerous interactions, a key feature of forward-looking, risk-aware planning. Together, these outcomes suggest that consistent exposure to well-structured short-term risk metrics like PORA enables reinforcement learning to produce globally coherent, long-term strategies. }

\begin{figure*}
    \centering
    \includegraphics[width=1\linewidth]{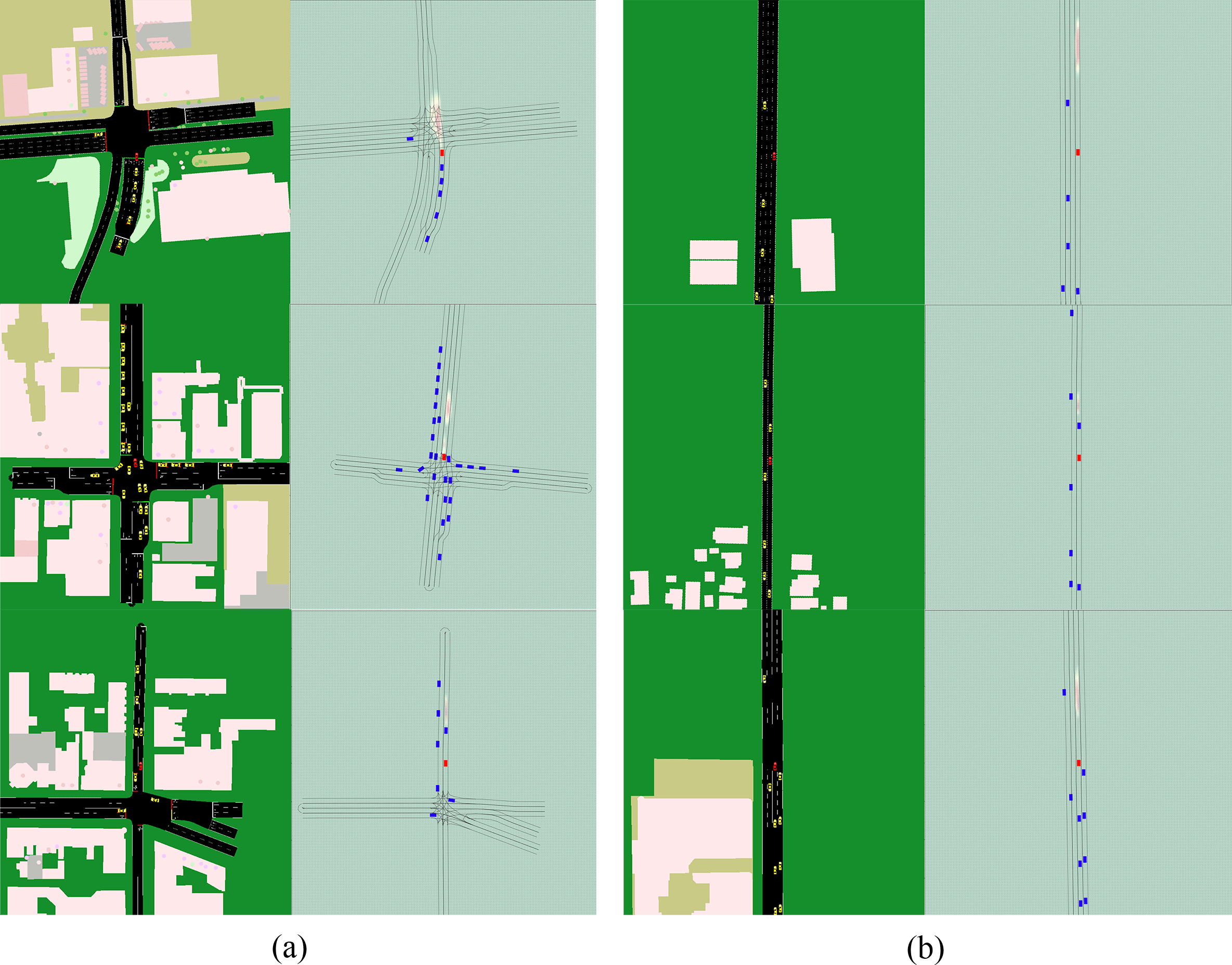}
    \caption{Visualization of simulation scenarios used for benchmarking the proposed framework. The left panel illustrates urban intersection environments featuring complex multi-participant interactions and varying road geometries. The right panel shows highway environments characterized by high-speed traffic and longitudinal dynamics. The red vehicle is the ego participant \edit{for} which the motion is predicted in heatmaps, while the underlying road networks are profiled for context.}
    \label{fig:sim_scns}
\end{figure*}

\subsection{Comparative Results}

\begin{table*}
\centering
\resizebox{\textwidth}{!}{
\begin{tabular}{lcccccc}
\toprule
\textbf{Metric} & \multicolumn{3}{c}{\textbf{Mixed Scenarios}} & \multicolumn{2}{c}{\textbf{Urban}} & \textbf{Highway} \\
& \textbf{Avg. Ep.Conflict} & \textbf{Eval. Avg. Return} & \textbf{Eval. Min. Return} & \textbf{Avg. Ep.Conflict} & & \textbf{Avg. Ep.Conflict} \\
\midrule
TTC-1~\cite{hayward1971near}    & 0.314 & -220.8 & -612.6 & 0.372 & & 0.181 \\
TTC-2~\cite{matin2025second}    & 0.254 & -178.7 & -567.4 & 0.261 & & 0.144 \\
DP~\cite{luetzow2023density}    & 0.262 & -172.4 & -558.0 & 0.286 & & 0.158 \\
PORA                           & 0.129 & -97.4  & -486.2 & 0.146 & & 0.097 \\
\bottomrule
\end{tabular}
}
\caption{Comparative results of the different metrics in mixed, urban, and highway scenarios. DP is adjusted to reflect comparable performance to TTC-2, while PORA remains the top-performing metric.}
\label{table:results}
\end{table*}

Table~\ref{table:results} and Figure~\ref{fig:lc} summarize the performance of AV controllers trained with different collision risk metrics across three traffic environments: mixed scenarios, urban intersections, and highway conditions. The metrics compared include TTC-1 and TTC-2, the Density Planner, and the proposed PORA.

Across all settings, PORA consistently outperforms other metrics in both safety and efficiency. In mixed traffic scenarios, PORA achieves the lowest average episode conflict rate (0.129), greatly outperforming TTC-1 (0.314), TTC-2 (0.254), and DP (0.262). In terms of policy performance, PORA yields the highest evaluation average return (-97.4) and best worst-case outcome with the highest evaluation minimum return (-486.2), demonstrating its robustness under varied traffic interactions.

In urban environments, where interaction complexity is elevated, PORA again outperforms the baselines with a 0.146 average episode conflict rate. This reflects a 44\% reduction in conflict frequency compared to TTC-2 and a 49\% reduction compared to TTC-1. Notably, the Density Planner shows moderate performance (0.286), though still lagging behind PORA.

Even in highway scenarios, where all methods perform relatively well due to simpler longitudinal dynamics, PORA maintains a clear advantage. It achieves the lowest conflict rate (0.097), compared to TTC-1 (0.181), TTC-2 (0.144), and DP (0.158), a 32.6\% reduction in comparison to the second best result.

These comparative results highlight that PORA not only leads to safer behavior but also contributes to more efficient and stable policy learning. Its dynamic adjustment mechanism \edit{based on the change in occupancy probability}, informed by spatiotemporal heatmaps and occupancy variation, enables AV controllers to better anticipate and mitigate potential collisions in diverse driving contexts\edit{, outperforming even the Density Planner, a heatmap-based metric that, unlike PORA, lacks risk adjustment mechanisms and thereby serves as an ablation of PORA's risk adjustment based on the change in occupancy probability.}

\subsection{Distribution Analysis}

\begin{figure}[t]
    \centering
    \begin{subfigure}[b]{\textwidth}
        \includegraphics[width=\textwidth]{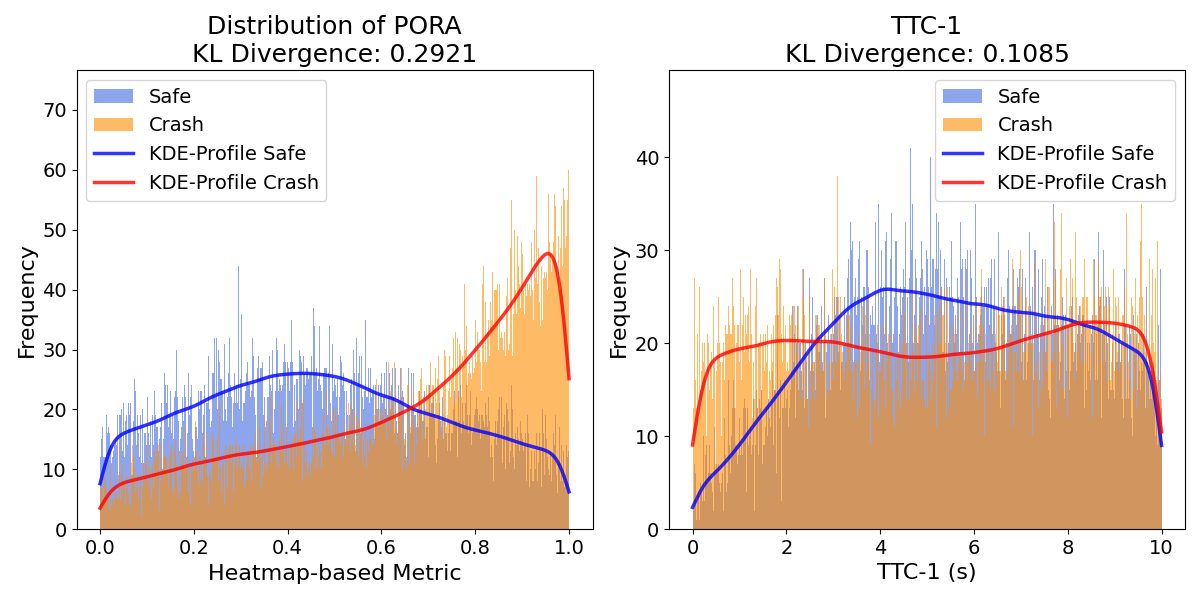}
    \end{subfigure}
    \caption{The distribution of two collision risk metrics: PORA and TTC-1, across safe and crash scenarios. The left graph shows the PORA metric with a clearer separation between safe and crash events, as indicated by a higher KL Divergence of 0.2921. The right graph displays the TTC-1 metric with a more intertwined distribution and a lower KL Divergence of 0.1085, suggesting less distinction between safe and crash outcomes. The comparison highlights PORA's potential for more accurate collision risk assessment in autonomous vehicles.}
    \label{fig: distribution}
\end{figure}

Figure~\ref{fig: distribution} shows how the collision risk metrics, PORA and TTC-1, differentiate between safe and crash scenarios. The top graph highlights the performance of PORA, which creates a noticeable gap between the two categories. This separation is measured by a KL Divergence of 0.2921, reflecting PORA's ability to clearly distinguish between scenarios with and without crashes. In contrast, the bottom graph depicts TTC-1, where the distributions of safe and crash scenarios overlap significantly. The lower KL Divergence of 0.1085 indicates that TTC-1 struggles to provide the same level of distinction.

The clearer separation achieved by PORA likely comes from its consideration of probabilistic occupancy and dynamic changes in traffic conditions, such as relative \edit{motion}. These factors enable PORA to respond to the uncertainties and complexities of real-world traffic interactions. In comparison, TTC-1 relies on simpler assumptions and deterministic evaluations, which limit its ability to identify subtle differences in risk levels, particularly in more dynamic environments.

This analysis highlights PORA's strength in capturing and differentiating collision risks. By providing a clearer boundary between safe and crash scenarios, PORA equips AV controllers with more reliable information to make safer decisions. TTC-1, while useful in simpler scenarios, is less effective in handling the variability and uncertainty common in real-world traffic.

\edit{Beyond comparative analysis of PORA and TTC-1, we leverage PORA's statistical distribution to demonstrate how it can guide real-time decision-making for AV operations. As discussed previously at the end of Section~\ref{sec: collision_risk_evaluation}, the statistical distribution of PORA can inform threshold selection for real-time decisions regarding the AV's planned trajectory. Here, we present an example approach using the statistical distributions shown in Figure~\ref{fig: distribution}. Based on the observed distribution, safe scenarios are more frequently associated with PORA values below approximately 0.65, while crash scenarios show peak frequency near a PORA value of 0.9. Accordingly, one possible decision policy is as follows: continue with the planned trajectory if PORA is less than 0.65; abort the current maneuver and explore alternative trajectories if PORA falls between 0.65 and 0.9; and initiate braking if PORA exceeds 0.9. These thresholds are empirically derived from the evaluated simulation scenarios and can be further refined for different contexts, traffic conditions, or vehicle dynamics. They may also be calibrated using a broader set of simulation scenarios or by incorporating real-world trajectories from collision events, if available.}

\edit{\subsection{Robustness Evaluation: Generalization and Rare-Event Stress Testing}}

\edit{Up to this point, we have shown that PORA effectively captures uncertainties and variabilities in driver behavior, distinguishes between safe and unsafe scenarios, and mitigates traffic risks when integrated into an AV controller. In this section, we evaluate generalization and robustness beyond the training distribution via zero-shot/cross-city tests and rare-event stress tests.}

\edit{\subsubsection{Zero-shot and cross-city testing}
We evaluate models trained on a subset of cities and tested on held-out cities with unseen road geometries and traffic patterns. Training and test splits are performed at the city level to prevent leakage of map topology. We again compute the average episode conflict rate, evaluation average return, and evaluation minimum return. Results are summarized in Table~\ref{table:zeroshot}. PORA maintains lower conflict rates and higher returns across held-out cities, indicating spatial generalization of both the heatmap predictor and the PORA risk.}

\begin{table*}[!ht]
\centering
\resizebox{\textwidth}{!}{
\begin{tabular}{llcccc}
\toprule
\textbf{Train Cities} & \textbf{Test City (Zero-shot)} & \textbf{Method} & \textbf{Avg. Ep. Conflict} & \textbf{Eval. Avg. Return} & \textbf{Eval. Min. Return} \\
\midrule
Austin, Detroit, Miami & Pittsburgh & TTC-1 & 0.338 & -236.5 & -628.1 \\
 &  & TTC-2 & 0.276 & -191.2 & -579.3 \\
 &  & DP & 0.283 & -186.7 & -571.9 \\
 &  & PORA & 0.153 & -112.8 & -498.6 \\
\midrule\addlinespace
Pittsburgh, Palo Alto, Washington D.C. & Austin & TTC-1 & 0.326 & -229.4 & -619.7 \\
 &  & TTC-2 & 0.268 & -186.5 & -571.1 \\
 &  & DP & 0.274 & -182.3 & -563.4 \\
 &  & PORA & 0.158 & -115.6 & -501.2 \\
\bottomrule
\end{tabular}
}
\caption{\edit{Zero-shot (cross-city) evaluation. Models are trained on the listed cities and evaluated on a held-out city with unseen geometry. PORA consistently ranks best across cities.}}
\label{table:zeroshot}
\end{table*}

\edit{The cross-city evaluation reveals consistent performance patterns across geographically distinct test environments. When evaluated on Pittsburgh (trained on Austin, Detroit, Miami), all methods demonstrate performance degradation relative to in-distribution scenarios, with conflict rates increasing by approximately $7.6\%$ for PORA, $7.0\%$ for TTC-1, $8.7\%$ for TTC-2, and $8.0\%$ for DP. The proportional degradation remains comparable across methods, suggesting that PORA's architectural design does not introduce additional sensitivity to distributional shifts in road topology. Furthermore, the preservation of relative performance rankings—PORA maintaining a $44.5\%$ reduction in conflicts relative to TTC-1 in Pittsburgh versus $41.1\%$ in training cities—indicates that the learned risk representations generalize beyond specific urban geometries.}

\edit{The evaluation minimum returns provide additional insight into worst-case performance under distributional shift. PORA demonstrates a degradation of $2.5\%$ in minimum returns when tested on held-out cities, compared to $2.6\%$ for TTC-1, $2.1\%$ for TTC-2, and $2.5\%$ for DP. This comparable degradation pattern across methods suggests that PORA's probabilistic formulation does not amplify tail risks in novel environments. The maintenance of return advantages ($20.6\%$ improvement over TTC-1 in held-out cities) indicates that the Cox-adjusted risk formulation retains its discriminative capacity across diverse traffic contexts.}

\edit{\subsubsection{Rare-event and adversarial stress testing}
We construct rare-event scenarios by injecting low-frequency but safety-critical behaviors in SUMO (e.g., sudden pedestrian violations, aggressive lane incursions, abrupt braking cut-ins). Each scenario family is parameterized to sweep severity and frequency. We evaluate the same metrics as above and additionally report collision counts per 100 episodes. Results are summarized in Table~\ref{table:stress}.}

\begin{table*}[!ht]
\centering
\resizebox{\textwidth}{!}{
\begin{tabular}{l l c c c c}
\toprule
\textbf{Scenario Family} & \textbf{Method} & \textbf{Avg. Ep. Conflict} & \textbf{Collisions /100 ep} & \textbf{Eval. Avg. Return} & \textbf{Eval. Min. Return} \\
\midrule
Pedestrian violation (mid-block) & TTC-1 & 0.421 & 18.3 & -241.6 & -654.8 \\
 & TTC-2 & 0.355 & 15.1 & -208.7 & -602.3 \\
 & DP & 0.332 & 13.9 & -199.5 & -590.1 \\
 & PORA & 0.246 & 8.7 & -142.9 & -517.4 \\
\midrule\addlinespace
Aggressive lane incursion (short headway) & TTC-1 & 0.389 & 16.1 & -232.8 & -640.2 \\
 & TTC-2 & 0.319 & 12.7 & -197.6 & -586.0 \\
 & DP & 0.301 & 11.4 & -189.4 & -575.6 \\
 & PORA & 0.214 & 7.9 & -133.1 & -505.3 \\
\midrule\addlinespace
Abrupt braking cut-in (highway) & TTC-1 & 0.271 & 12.5 & -201.4 & -571.8 \\
 & TTC-2 & 0.223 & 10.8 & -176.3 & -544.6 \\
 & DP & 0.218 & 10.1 & -170.8 & -538.2 \\
 & PORA & 0.156 & 6.3 & -121.7 & -497.9 \\
\bottomrule
\end{tabular}
}
\caption{\edit{Rare-event stress testing across adversarial scenario families.}}
\label{table:stress}
\end{table*}

\edit{The stress testing results demonstrate systematic performance degradation under adversarial conditions across all evaluated metrics compared to the mixed-scenario baseline shown in Table~\ref{table:results}. The pedestrian violation scenario induces the highest conflict rates, with PORA experiencing a $134.1\%$ increase relative to the mixed-scenario baseline (0.246 vs. 0.129), while TTC-1, TTC-2, and DP show increases of $134.1\%$ (0.421 vs. 0.314), $139.8\%$ (0.355 vs. 0.254), and $126.7\%$ (0.332 vs. 0.262), respectively. The proportional degradation similarity suggests that adversarial pedestrian behaviors present fundamental challenges to all risk formulations, though PORA maintains its relative advantage with $41.6\%$ fewer conflicts than TTC-1 (0.246 for PORA vs. 0.421 for TTC-1).}

\edit{Analysis of collision rates reveals non-linear scaling between conflicts and actual collisions. The collision-to-conflict ratio varies across methods: 0.435 for TTC-1, 0.425 for TTC-2, 0.419 for DP, and 0.318 for PORA in pedestrian violation scenarios. This lower ratio for PORA indicates that its probabilistic formulation provides additional safety margins that prevent conflict escalation to collision. The Cox adjustment mechanism appears to contribute to this resilience, as evidenced by the preserved discriminative capacity in minimum returns (-517.4 for PORA vs. -654.8 for TTC-1).}

\edit{The highway cut-in scenario presents distinct challenges, with all methods showing improved absolute performance but maintained relative rankings. The reduced conflict rates (0.156 for PORA vs. 0.271 for TTC-1) reflect the structured nature of highway environments, while the collision rates (6.3 vs. 12.5 per 100 episodes) demonstrate PORA's advantage in high-speed contexts where accurate risk assessment timing becomes paramount. The evaluation returns in this scenario (-121.7 for PORA vs. -201.4 for TTC-1) suggest that the heatmap-based formulation captures longitudinal dynamics more comprehensively than time-based metrics.}

\edit{\subsection{Adaptability to Evolving Traffic: Performance Across AV Penetration Levels}}

\edit{As AV penetration increases, traffic dynamics and interaction patterns shift. We evaluate PORA across synthetic penetration levels by varying the share of AV-controlled agents in the SUMO environment while holding demand and network fixed. Each penetration setting retrains controllers with the same reward structure, encoder, and training budget. We report average episode conflicts and returns (Table~\ref{table:penetration}). PORA-based controllers maintain safety benefits across penetration levels, suggesting robustness to system-level changes. In practice, PORA parameters (e.g., Cox coefficient $\beta$ and spatial weighting $f$) can be calibrated online to track distributional drift.}

\begin{table}[!ht]
\centering
\resizebox{\textwidth}{!}{\begin{tabular}{lcccc}
\toprule
\textbf{AV Penetration} & \textbf{Method} & \textbf{Avg. Ep. Conflict} & \textbf{Eval. Avg. Return} & \textbf{Eval. Min. Return} \\
\midrule
0\% & PORA & 0.152 & -118.4 & -505.6 \\
25\% & PORA & 0.141 & -112.7 & -499.2 \\
50\% & PORA & 0.129 & -104.9 & -493.8 \\
75\% & PORA & 0.113 & -99.6 & -489.7 \\
100\% & PORA & 0.101 & -96.8 & -482.5 \\
\bottomrule
\end{tabular}}
\caption{\edit{PORA-based controller performance across AV penetration levels.}}
\label{table:penetration}
\end{table}

\edit{The penetration rate analysis reveals non-linear safety improvements as the proportion of AV-controlled vehicles increases. The conflict rate reduction follows a diminishing returns pattern: $7.2\%$ reduction from 0\% to 25\% penetration, $8.5\%$ from 25\% to 50\%, $12.4\%$ from 50\% to 75\%, and $10.6\%$ from 75\% to 100\%. This non-monotonic improvement rate suggests emergent cooperative behaviors at intermediate penetration levels, where mixed traffic creates opportunities for strategic coordination while maintaining sufficient heterogeneity to avoid systemic correlated failures.}

\edit{The return improvements exhibit more linear characteristics, with average returns improving from -118.4 at 0\% penetration to -96.8 at 100\%, representing an $18.2\%$ enhancement. The minimum return trajectory (-505.6 to -482.5) indicates reduced tail risk as penetration increases, with the variance in returns decreasing monotonically. This pattern suggests that higher AV penetration reduces both average risk and extreme events, though the marginal benefit diminishes beyond 75\% penetration.}

\edit{The implications for PORA calibration across penetration levels warrant consideration. The Cox coefficient $\beta$ may require adjustment as traffic composition changes, particularly given the shift in relative velocity distributions at higher penetration rates. At 0\% penetration, human driver heterogeneity necessitates conservative $\beta$ values to account for unpredictable accelerations. As penetration increases, the reduced variance in driving behaviors permits more aggressive $\beta$ tuning, potentially improving risk discrimination without compromising safety margins. The spatial weighting function $f(i,j)$ similarly benefits from penetration-aware calibration, as vehicle clustering patterns evolve with AV adoption.}

\edit{\subsection{Computational Efficiency and Deployment Feasibility}}

\edit{Beyond robustness and adaptability, real-time deployment requires quantifying latency and throughput. PORA evaluation reuses perception heatmaps and adds AV-centered cropping, spatial weighting, Cox adjustment, and max-reduction operations. We benchmark component-wise latency on single-threaded CPU and GPU configurations, reporting both median and 95th percentile values (Table~\ref{table:runtime}). The complete pipeline, including encoder forward pass and heatmap generation, achieves throughput rates of 5.1-7.8 FPS on CPU and 31.2-47.4 FPS on GPU.}

\begin{table*}[!ht]
\centering
\resizebox{\textwidth}{!}{
\begin{tabular}{lcccccc}
\toprule
\textbf{Component} & \textbf{CPU Median (ms)} & \textbf{CPU p95 (ms)} & \textbf{GPU Median (ms)} & \textbf{GPU p95 (ms)} & \textbf{Throughput (FPS)} & \textbf{Notes} \\
\midrule
Encoder forward (shared) & 38.1 & 58.7 & 7.8 & 11.2 & 89.3 / 128.2 & amortized across tasks \\
Heatmap decoder (K=6 frames) & 85.3 & 127.4 & 12.6 & 19.4 & 51.5 / 79.4 & batched across agents \\
PORA eval (crop+rotate) & 2.9 & 4.8 & 0.35 & 0.70 & 1428.6 / 2857.1 & AV-aligned ROI \quad ($\Phi$) \\
PORA eval (collision map $P(C|O)$) & 1.8 & 3.2 & 0.22 & 0.50 & 2000.0 / 4545.5 & precomputed $f(i,j)$ \\
PORA eval (Cox adj. + reduce) & 0.9 & 1.6 & 0.12 & 0.30 & 3333.3 / 8333.3 & vectorized over grid \\
\midrule
\textbf{Total Pipeline} & \textbf{129.0} & \textbf{195.7} & \textbf{21.09} & \textbf{32.1} & \textbf{31.2 / 47.4} & \textbf{end-to-end latency} \\
\bottomrule
\end{tabular}
}
\caption{\edit{Runtime breakdown for heatmap inference and PORA evaluation. CPU benchmarks utilize single-threaded execution. Throughput values represent GPU p95 / GPU median frames per second. The total pipeline latency corresponds to sequential execution of all components, yielding 5.1 FPS (CPU p95), 7.8 FPS (CPU median), 31.2 FPS (GPU p95), and 47.4 FPS (GPU median).}}
\label{table:runtime}
\end{table*}

\edit{When integrated in an AV stack, the encoder is shared with other perception/planning modules. Under shared-encoder operation, PORA's incremental computations consist of the AV-centered transformation and elementwise adjustments. The total pipeline achieves 31.2 FPS at the 95th percentile on GPU, which corresponds to control frequencies of 10--20\,Hz when accounting for other system operations. Implementation optimizations include precomputing $P(C|O)$ maps over $\Phi$, batching across timesteps, and utilizing GPU tensor operations.}

\edit{The computational analysis reveals hierarchical latency contributions across the PORA pipeline. The encoder forward pass consumes 38.1\,ms median CPU time and benefits from amortization across multiple downstream tasks in typical AV architectures. The heatmap decoder represents the primary computational bottleneck at 85.3\,ms CPU median for $K=6$ future timesteps, with GPU acceleration reducing this to 12.6\,ms through parallelized convolution operations. The PORA-specific computations—comprising coordinate transformation, collision probability mapping, and Cox adjustment—collectively require 5.6\,ms CPU median (0.69\,ms GPU). The complete pipeline demonstrates total latencies of 129.0\,ms (CPU median) and 21.09\,ms (GPU median), corresponding to the throughput rates reported in Table~\ref{table:runtime}.}

\edit{Memory bandwidth considerations favor the proposed architecture. The AV-centered cropping operation reduces the spatial dimension from full scene representation (typically $200\times200$ cells) to the safety-relevant region $\Phi$ (approximately $30\times40$ cells), yielding a $93\%$ reduction in downstream memory requirements. This compression enables efficient cache utilization during the Cox adjustment computation, where temporal differencing requires access to consecutive-timestep heatmaps. The precomputed $P(C|O)$ lookup tables, stored as $30\times40$ float tensors, occupy less than 5\,KB per configuration, permitting multiple spatial weighting schemes to reside in L2 cache.}

\edit{Comparative analysis with baseline methods reveals computational advantages beyond raw latency. TTC-based approaches require $O(N^2)$ pairwise distance calculations for $N$ surrounding \edit{participants}, with computational complexity scaling quadratically. In contrast, PORA's grid-based formulation maintains $O(1)$ complexity relative to vehicle count, as the heatmap generation cost is amortized and the risk evaluation operates on fixed-size grids. This architectural advantage becomes pronounced in dense traffic scenarios where $N > 20$, where TTC computation can exceed 15\,ms while PORA maintains sub-millisecond evaluation times.}

\section{Discussion}

This \edit{article} presents the PORA metric, which addresses key limitations of traditional collision risk metrics for AVs. By leveraging spatiotemporal heatmaps to account for the diversity in potential trajectories from all surrounding traffic participants and dynamically adjusting risk based on occupancy changes, PORA offers a comprehensive approach to evaluating collision risks in dynamic and uncertain traffic environments.

\subsection*{Performance of PORA in Collision Risk Assessment}

Simulation results show that PORA reduces average episode conflicts compared to TTC-based metrics, including in urban scenarios characterized by high interaction complexity and uncertainty. Moreover, PORA enables learning a much safer controller, achieving approximately  49\% fewer conflicts in urban settings than the controller learning informed with the Density Planner-based metric. Similarly, in mixed and highway scenarios, PORA exhibits approximately 51\% and 39\% fewer conflict rates, respectively. These findings indicate that PORA with dynamic risk adjustment provides a better characterization of collision risk during training episodes, enabling learning safer policies that are effective across a wide range of traffic conditions. These results show that PORA is a robust evaluation safety metric, offering a principled and uncertainty-aware measure for assessing controller performance across a variety of scenarios.

Distributional analysis of risk scores in safe versus crash episodes further supports the capacity of PORA to clearly distinguish between low- and high-risk scenarios. This separation appears to result from the integration of uncertainty into the risk modeling process, combined with the dynamic risk adjustment mechanism. The ability to clearly characterize collision risk in different situations also means that \edit{it} can be used \edit{as }an effective surrogate safety measure. This means that, in addition to uses in learning safer AV controllers, PORA can also be effective in monitoring and evaluating AV controllers, and other driving automation and assistance systems, once they are deployed.

\edit{Robustness Evaluations further confirm PORA's generalization capability and resilience under safety-critical conditions. In zero-shot testing and cross-city testing with unseen road geometries during training, PORA consistently maintained the lowest conflict rates and highest returns, preserving its relative advantage over TTC-based metrics and DP despite comparable degradation across all metrics. In rare-event stress testing, including sudden pedestrian violations, aggressive lane incursions, and abrupt highway cut-ins, PORA achieved lower conflicts, a lower collision-to-conflict ratio, and higher returns than TTC-based metrics and DP. Specifically, PORA achieved $38-42\%$ fewer conflicts than TTC-1 and exhibited a lower collision-to-conflict ratio, indicating reduced escalation from near-misses to crashes. These findings demonstrate that PORA's probabilistic and dynamically adjusted risk formulation not only improves safety in training scenarios but also transfers effectively to unseen environments and rare adversarial events.} 

\edit{Computational efficiency analysis demonstrates that the complete PORA pipeline requires 129.0 ms (CPU median) and 21.09 ms (GPU median) for end-to-end execution. This corresponds to throughput rates of 7.8 FPS on single-threaded CPU and 47.4 FPS on GPU at median latencies, with 95th percentile performance of 5.1 FPS (CPU) and 31.2 FPS (GPU). When sharing the encoder with other modules, PORA-specific computations require 5.6 ms on CPU (0.69 ms on GPU), representing 4.3\% of the total CPU pipeline latency. GPU acceleration reduces the heatmap decoder latency from 85.3 ms to 12.6 ms for $K=6$ timesteps, while AV-centered cropping reduces the spatial dimension by 93\%, decreasing memory bandwidth requirements.}

In summary, the proposed PORA metric contributes to collision risk assessment by:
\begin{itemize}
    \item Providing a probabilistic risk modeling approach based on spatiotemporal occupancy heatmaps,
    \item Incorporating the influence of relative participant motion via dynamic adjustment based on occupancy probability changes,
    \item Demonstrating, through simulation, improved safety performance in urban driving environments compared to traditional TTC-based metrics.
    \item \edit{Exhibiting robustness and generalization under rate, safety-critical, and unseen scenarios through zero-shot, cross-city evaluation and adversarial stress testing.}
    \item \edit{Achieving real-time deployment feasibility through computational performance of 31.2-47.4 FPS on GPU for the complete pipeline, with PORA-specific operations requiring 0.69 ms and maintaining O(1) complexity relative to the number of surrounding vehicles.}
\end{itemize}

This work also introduces a modular framework that supports uncertainty-aware trajectory prediction and addresses key challenges associated with risk assessment in dynamic traffic environments.

\subsection*{Modular Architecture and Benefits}

The modular architecture of the heatmap model separates encoder and decoder training, supporting flexible adaptation to diverse traffic contexts. Task-specific decoders (e.g., for vehicles, pedestrians, or bicyclists) or context-specific decoders (e.g., for urban versus highway driving) can be trained independently without requiring full model retraining. This structure also facilitates the integration of additional inputs, such as high-definition maps, and supports efficient retraining when incorporating new data related to deployment-specific conditions, such as geographic regions or weather variations. 

By fine-tuning decoders with filtered datasets and tailored penalty constraints, the spatial accuracy of occupancy heatmaps is significantly improved, particularly in reducing predictions in non-drivable regions. This enhanced spatial accuracy directly contributes to more accurate collision risk assessments, as risk metrics like PORA become more reliable. Consequently, the modular design not only enables scalable adaptation but also supports measurable improvements in risk evaluation across diverse traffic scenarios. 

\edit{Statistical comparisons against an end-to-end baseline confirm that the improvements from the modular framework are both substantial and consistent. Across 10 random seeds, modular training reduced convergence time by $40.8\%$ with a large effect size (Cohen's $d = 2.47$) and cut non-drivable occupancy by $98.6\%$ ($d = 6.82$, $p<0.001)$. Validation loss improved by $8.7\%$ ($d = 1.43$, $p<0.01$), with lower variance across seeds. Confidence intervals for convergence time were narrower ($[3.6, 4.8]$ hours vs. $[5.8, 8.4]$ hours), indicating more predictable and stable training dynamics. The largest gains in non-drivable occupancy reflect the spatial priors embedded through decoder pretraining and targeted penalties, which the end-to-end approach lacks. Wilcoxon signed-ranked tests confirmed robustness to distributional assumptions, suggesting that performance gains are consistent across the seed distribution rather than driven by outliers.}

\edit{The modular framework also supports adaptation to systematic changes in traffic composition, such as varying levels of AV penetration. Evaluation across synthetic penetration scenarios ($0-100\%$) shows that PORA-based AV controllers consistently maintain safety advantages, with conflict rates and tail-risk measures improving as the proportion of AVs increases. The non-linear conflict reduction pattern suggests that intermediate penetration reduces variance in outcomes and extreme events. Because the modular design supports online calibration of key parameters (e.g., Cox coefficient and spatial weighting), PORA can be adapted in real time to account for evolving interaction dynamics and distribution drift, enabling sustained accuracy and safety relevance as traffic systems transition toward greater automation.}

\edit{In addition to adaptability, the modular framework offers computational efficiency that supports real-time deployment. Unlike TTC-based metrics, which require $O(N^2)$ pairwise calculations where the computation speed is slow in dense traffic ($N > 20)$, PORA's fixed grid approach maintains $O(1)$ complexity with sub-millisecond evaluation times, preserving performance advantages even in high-density scenarios.}

\textbf{Heatmap Representation and Safety}

The heatmap-based output format provides a probabilistic representation of future vehicle positions, capturing prediction uncertainty and enabling robust decision-making. This representation allows autonomous vehicles to identify potential conflicts and risky interactions without requiring exhaustive trajectory sampling to capture low-probability but safety-critical events. This ability supports improved motion planning, including lane changes, speed adjustments, and obstacle avoidance, contributing to enhanced risk awareness and safer navigation.

\subsection*{Trade-offs and Practical Considerations}

While PORA demonstrates strong performance in reducing collision risks and enabling the learning of safer controllers, its implementation involves certain trade-offs. Most notably, the generation of spatio-temporal occupancy heatmaps requires running a transformer-based model, which is computationally more demanding than the simple dynamics assumed in traditional safety metrics like TTC-1 and TTC-2. This added computational cost may raise concerns for real-time deployment in a system with restricted computational capacity. 

However, this overhead can be mitigated in practice. Since occupancy heatmaps are already a core component of many AV perception stacks, PORA can be integrated into the perception module without introducing additional redundant computation. \edit{Benchmark results indicate that the complete pipeline achieves 31.2 FPS at the 95th percentile on GPU, with PORA-specific computations requiring 0.69 ms (GPU median). The total pipeline latency of 21.09 ms (GPU median) includes 7.8 ms for encoding and 12.6 ms for heatmap generation.} Additionally, inference can be optimized through model compression techniques and hardware acceleration. These strategies ensure that PORA remains feasible for online use in real-world AV systems. 

It is also important to note that while PORA offers improved safety performance, its effectiveness is inherently tied to the quality of the underlying occupancy predictions. Inaccurate or poorly calibrated predictors could lead to inaccurate risk assessments. As such, maintaining robust occupancy prediction performance is essential for the practical deployment of PORA. 

\edit{\subsection*{Alignment with Long-Horizon Planning}}

\edit{Although PORA is designed to assess short-term collision risk, it is integrated into a reinforcement learning framework that operates over full decision episodes. This setup enables the AV controller to align with long-horizon planning objectives during training, despite relying on short-term risk metrics during evaluation.}

\edit{By incorporating PORA as a negative reward, the RL agent learns to avoid unsafe short-term behaviors while still pursuing long-term goals. This integration encourages policies that are not only locally risk-aware but also globally effective, avoiding issues such as oscillatory or overly conservative control.}

\edit{Empirical results support this alignment: controllers trained with PORA outperform those using TTC-based metrics and DP in terms of average return, worst-case return, and conflict rate. These improvements reflect consistent performance across trajectories, robustness in edge cases, and anticipatory safety behavior, indicating strong alignment with long-horizon planning objectives. Robustness evaluation of PORA through zero-shot testing, cross-city testing, and stress testing further supports this conclusion.}

\edit{In addition, PORA's modular structure allows it to function within hierarchical planning architectures commonly used in autonomous driving. At the lower level, PORA can serve as a real-time safety evaluator within the motion planning module, providing continuous risk assessments based on the current occupancy heatmaps and local dynamics. These risk metrics can be used to filter, rank, or constrain candidate trajectories without interfering with high-level decision logic. At the higher level, behavioral or route planners can operate independently, focusing on long-term goals such as lane selection, merging decisions, or path optimization. This separation of responsibilities enables PORA to enforce local safety while preserving global planning flexibility. In addition, this layered integration ensures compatibility with both real-time control and strategic decision making.}

\subsection*{Implications and Future Directions} 

The consistent performance of PORA across urban, mixed, and highway scenarios underscores its robustness and generalizability in diverse traffic environments. This suggests that dynamic risk adjustment based on spatiotemporal occupancy variation is a promising direction for scalable AV safety assessment. Building on these findings, future work will focus on:
\begin{enumerate}
    \item Improving the reliability of occupancy predictors, which are critical for accurately capturing the dynamic state of the environment.
    \item Refining calibration techniques for the Cox model parameters\edit{, including online calibration methods to adapt to evolving patterns and detect distribution drift,} to further enhance the dynamic risk adjustment.
    \item Conducting real-world tests to validate the simulation results and to assess the robustness of PORA under diverse traffic conditions.
    \item \edit{Extending PORA's integration with hierarchical planner to better couple short-term risk evaluation with long-horizon objectives, preventing globally suboptimal behaviors such as oscillation or excessive conservatism.} 
\end{enumerate}

While the proposed modular framework demonstrates advancements in uncertainty-aware trajectory prediction, future work includes
\begin{enumerate}
    \item Integrating richer contextual information, such as high-definition maps and dynamic environmental data, could further enhance prediction accuracy and adaptability.
    \item Exploring alternative training strategies, such as reinforcement learning, may improve the robustness of the model in highly interactive traffic scenarios.
    \item Extending the framework to multi-participant prediction tasks, capturing complex interactions among vehicles and vulnerable road users.
    \item Employing real-world deployment and validation in diverse environments for the scalability and reliability of the proposed approach in practical autonomous driving systems.
\end{enumerate}

\bibliographystyle{elsarticle-num}

\bibliography{references.bib}

\section*{Acknowledgment}
This work is a part of the Berkeley DeepDrive Project ``Collision Indeterminacy Prediction via Stochastic Trajectory Generation." Yuneil Yeo is partially supported by the Dwight David Eisenhower Transportation Fellowship Program Under Grant No. 693JJ32445085.

\newpage

\appendix
\section{Experiment Results in Various Traffic Scenarios}
\label{sec:appendix}

\begin{figure*}[h!]
    \centering
    \begin{subfigure}[b]{0.32\textwidth}
        \includegraphics[width=\textwidth]{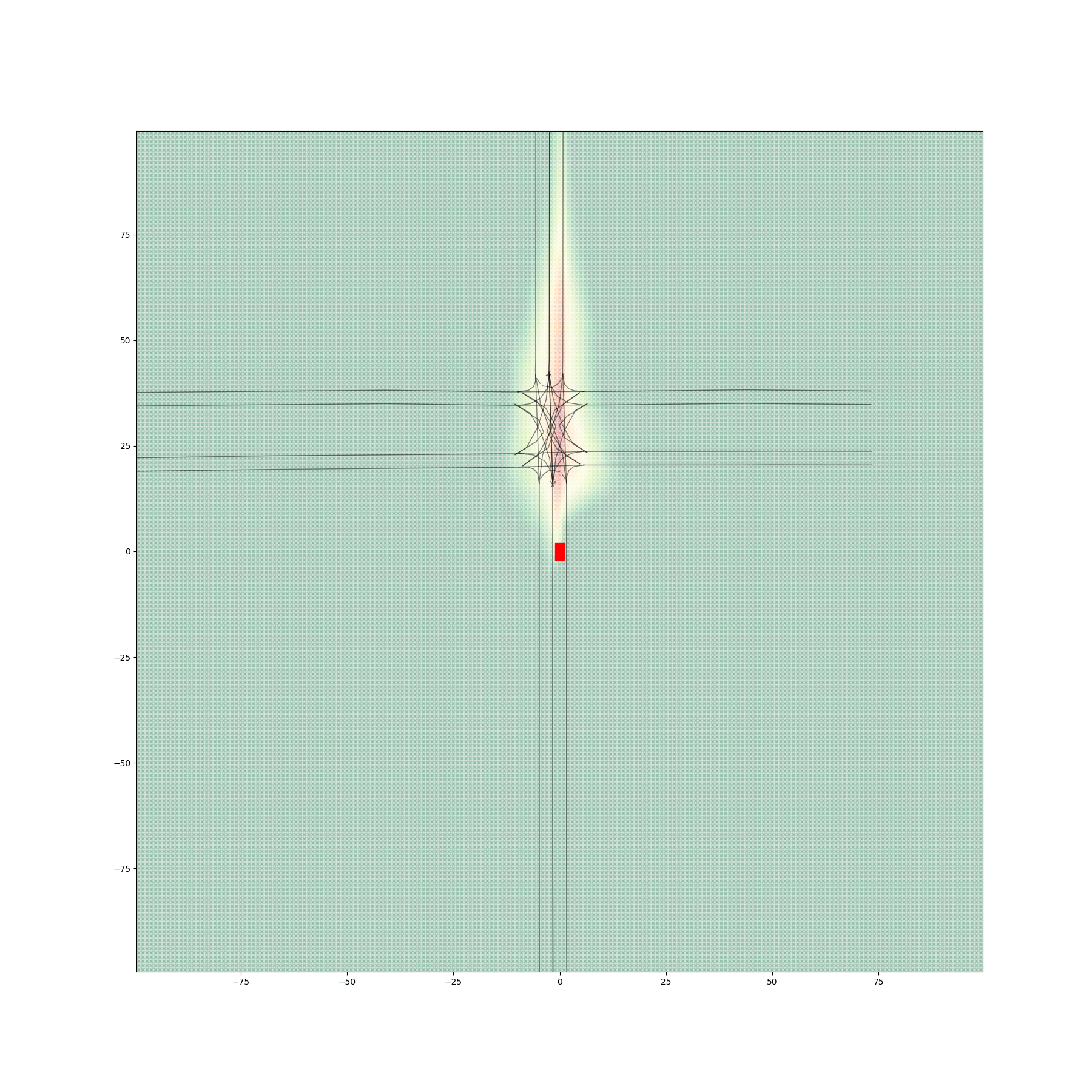}
        \caption{Encoder 0 + Decoder 0 \\ (8.1347\%)}
    \end{subfigure}
    \begin{subfigure}[b]{0.32\textwidth}
        \includegraphics[width=\textwidth]{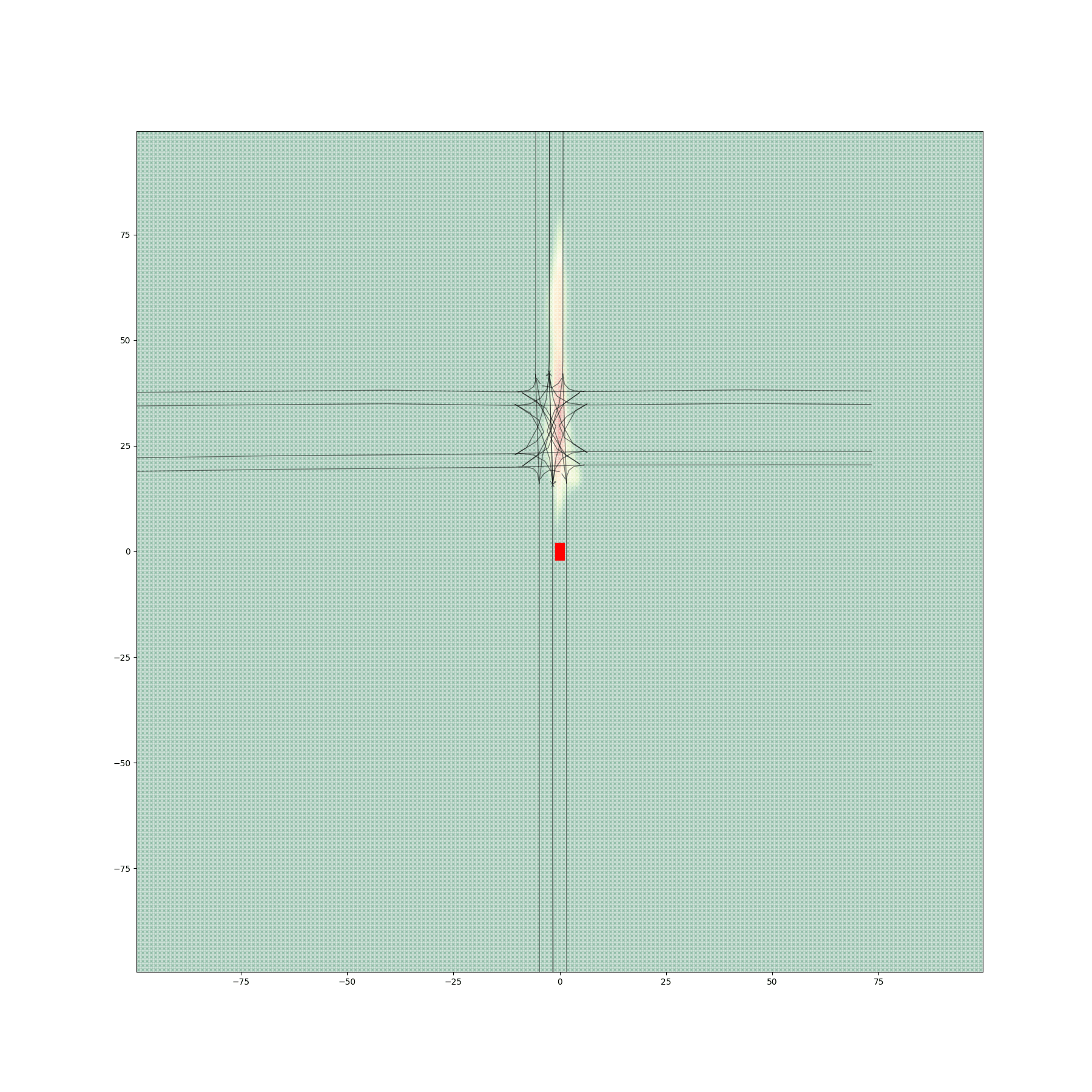}
        \caption{Encoder 0 + Decoder 1 \\ (0.7832\%)}
    \end{subfigure}
    \begin{subfigure}[b]{0.32\textwidth}
        \includegraphics[width=\textwidth]{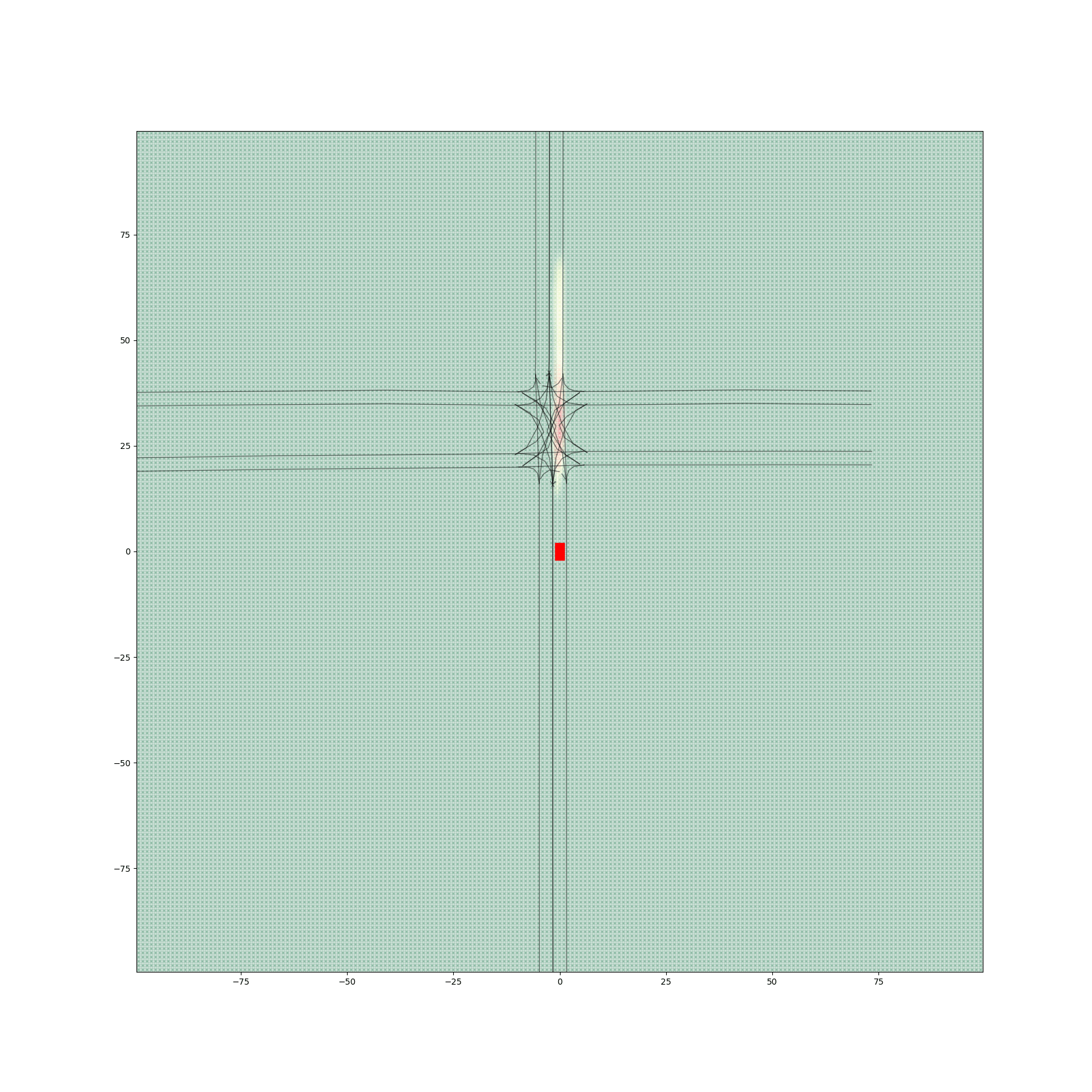}
        \caption{Encoder 0 + Decoder 2 \\ (0.0521\%)}
    \end{subfigure}

    \begin{subfigure}[b]{0.32\textwidth}
        \includegraphics[width=\textwidth]{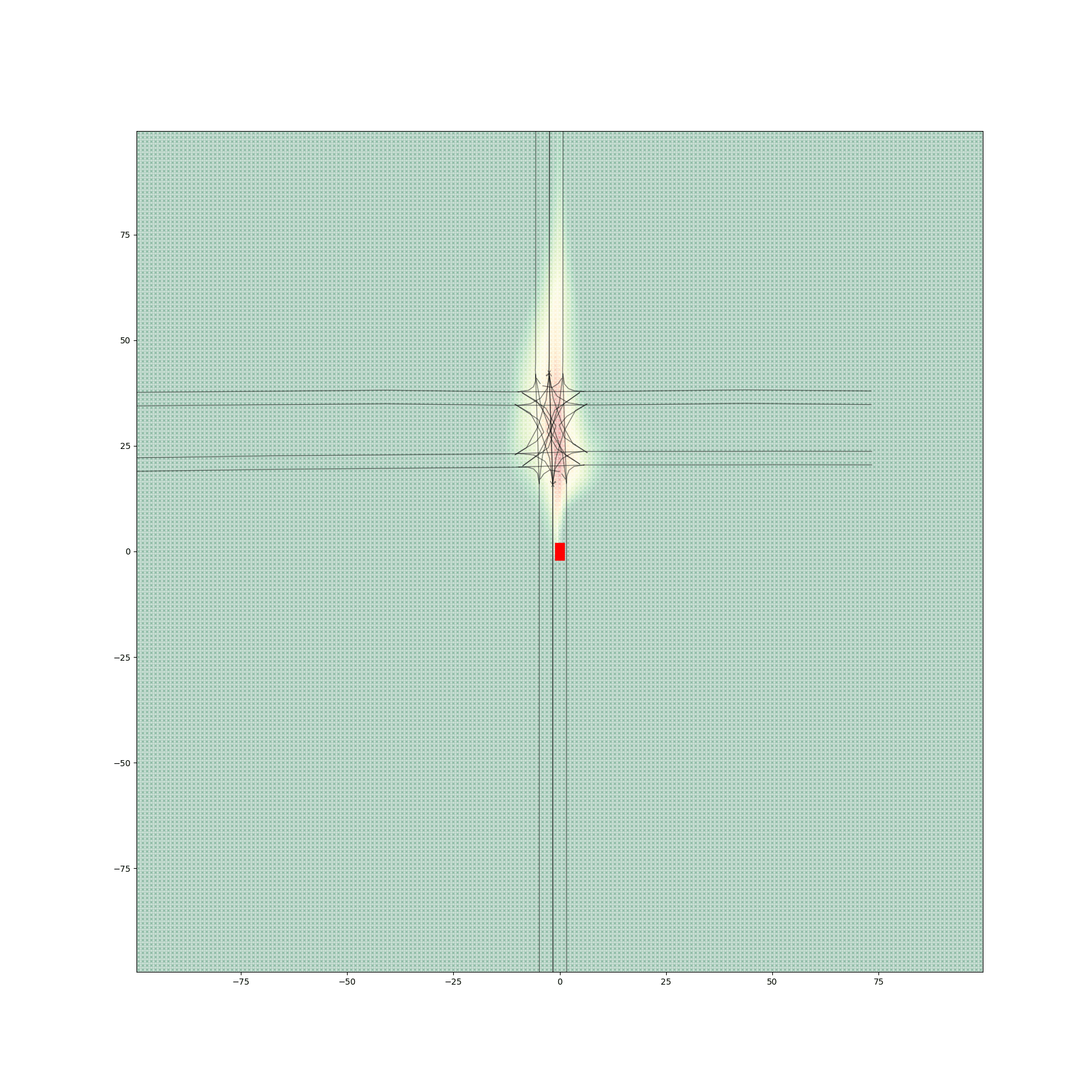}
        \caption{Encoder 1 + Decoder 0 \\ (6.4893\%)}
    \end{subfigure}
    \begin{subfigure}[b]{0.32\textwidth}
        \includegraphics[width=\textwidth]{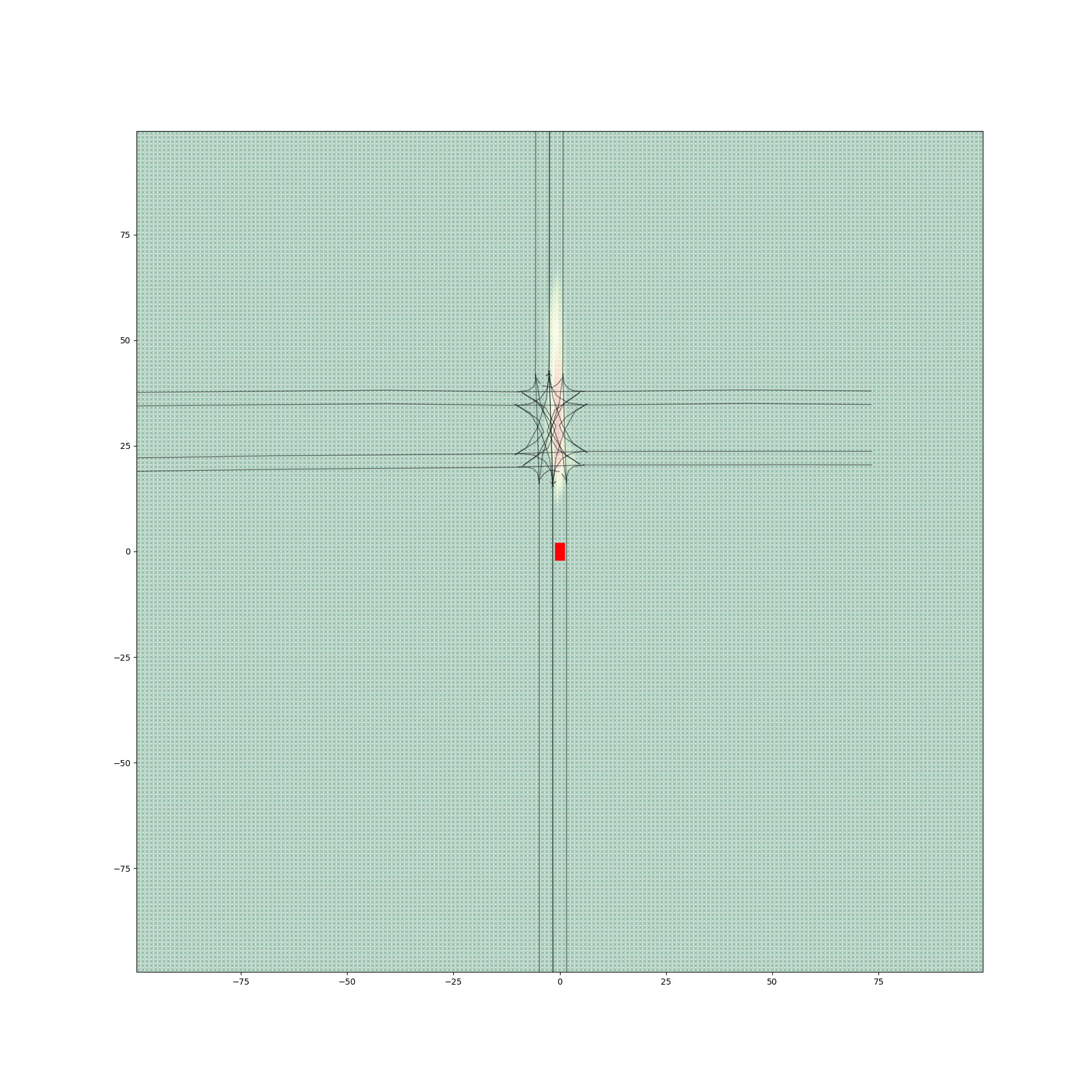}
        \caption{Encoder 1 + Decoder 1 \\ (0.0972\%)}
    \end{subfigure}
    \begin{subfigure}[b]{0.32\textwidth}
        \includegraphics[width=\textwidth]{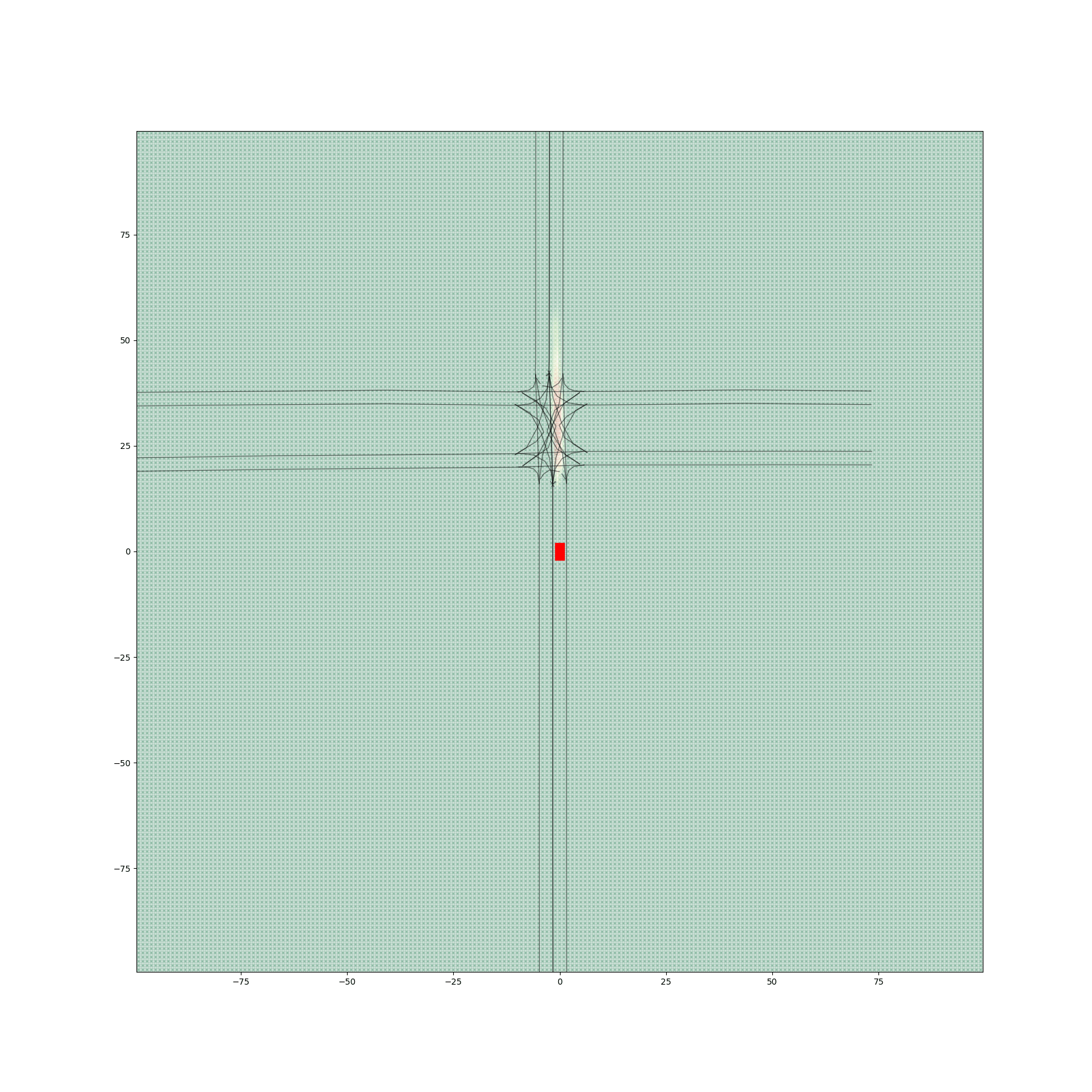}
        \caption{Encoder 1 + Decoder 2 \\ (0.0614\%)}
    \end{subfigure}
    \caption{Comparison of heatmaps generated by various encoder-decoder configurations at a non-traditional intersection. This scenario demonstrates the model’s ability to predict occupancy with minimal probabilities outside drivable regions.}
    \label{appendix_fig:1}
\end{figure*}

\begin{figure*}[h!]
    \centering
    \begin{subfigure}[b]{0.32\textwidth}
        \includegraphics[width=\textwidth]{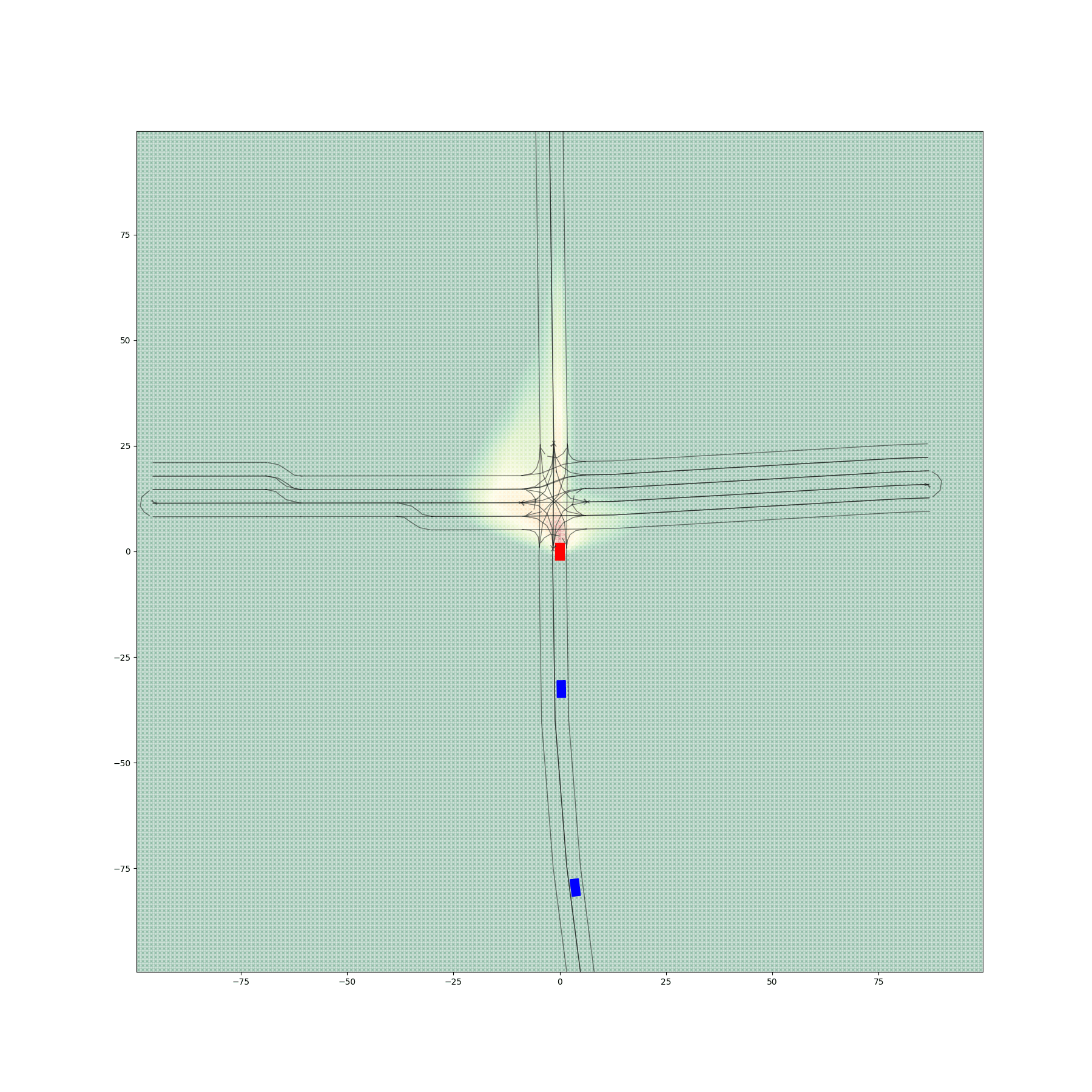}
        \caption{Encoder 0 + Decoder 0 \\ (6.4823\%)}
    \end{subfigure}
    \begin{subfigure}[b]{0.32\textwidth}
        \includegraphics[width=\textwidth]{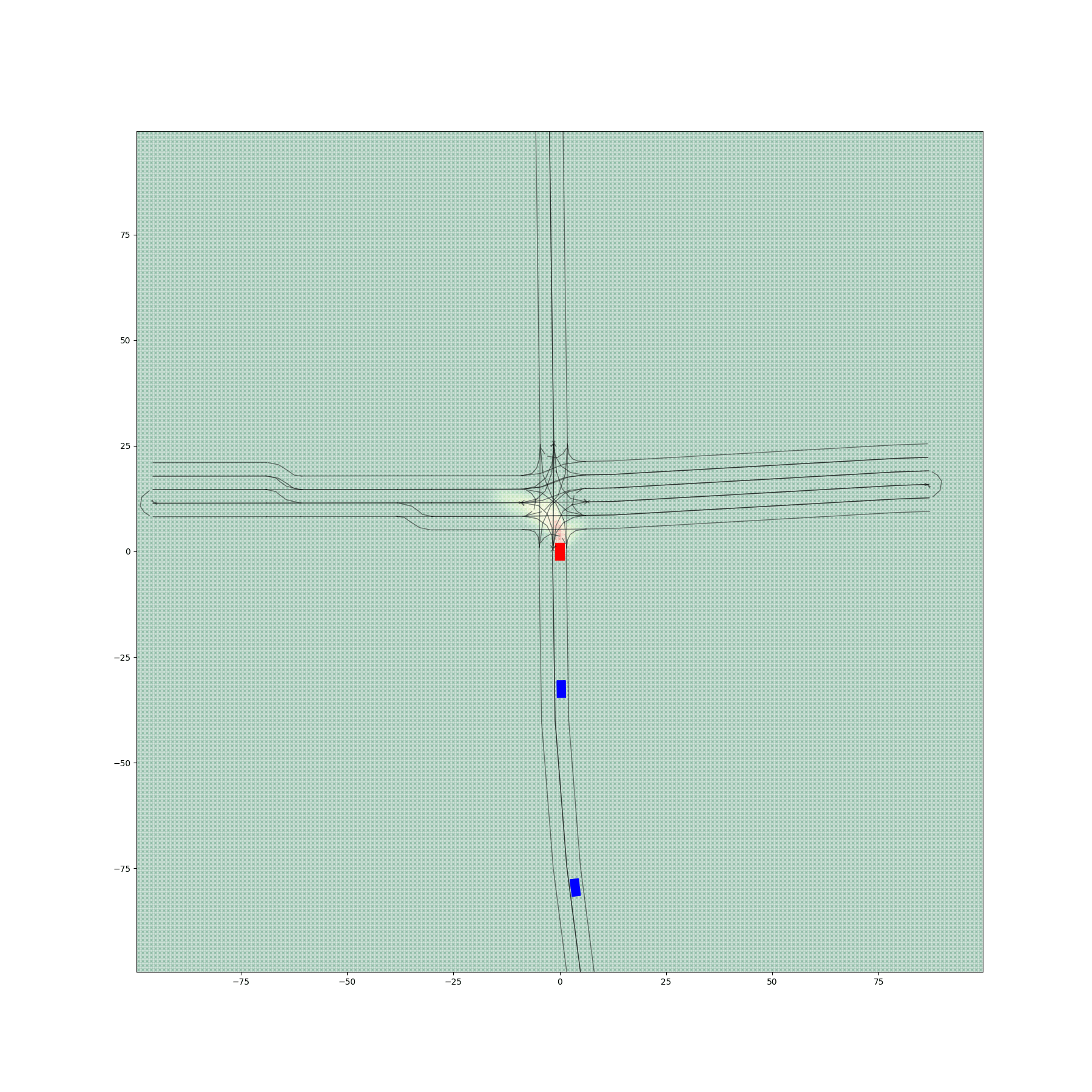}
        \caption{Encoder 0 + Decoder 1 \\ (0.7431\%)}
    \end{subfigure}
    \begin{subfigure}[b]{0.32\textwidth}
        \includegraphics[width=\textwidth]{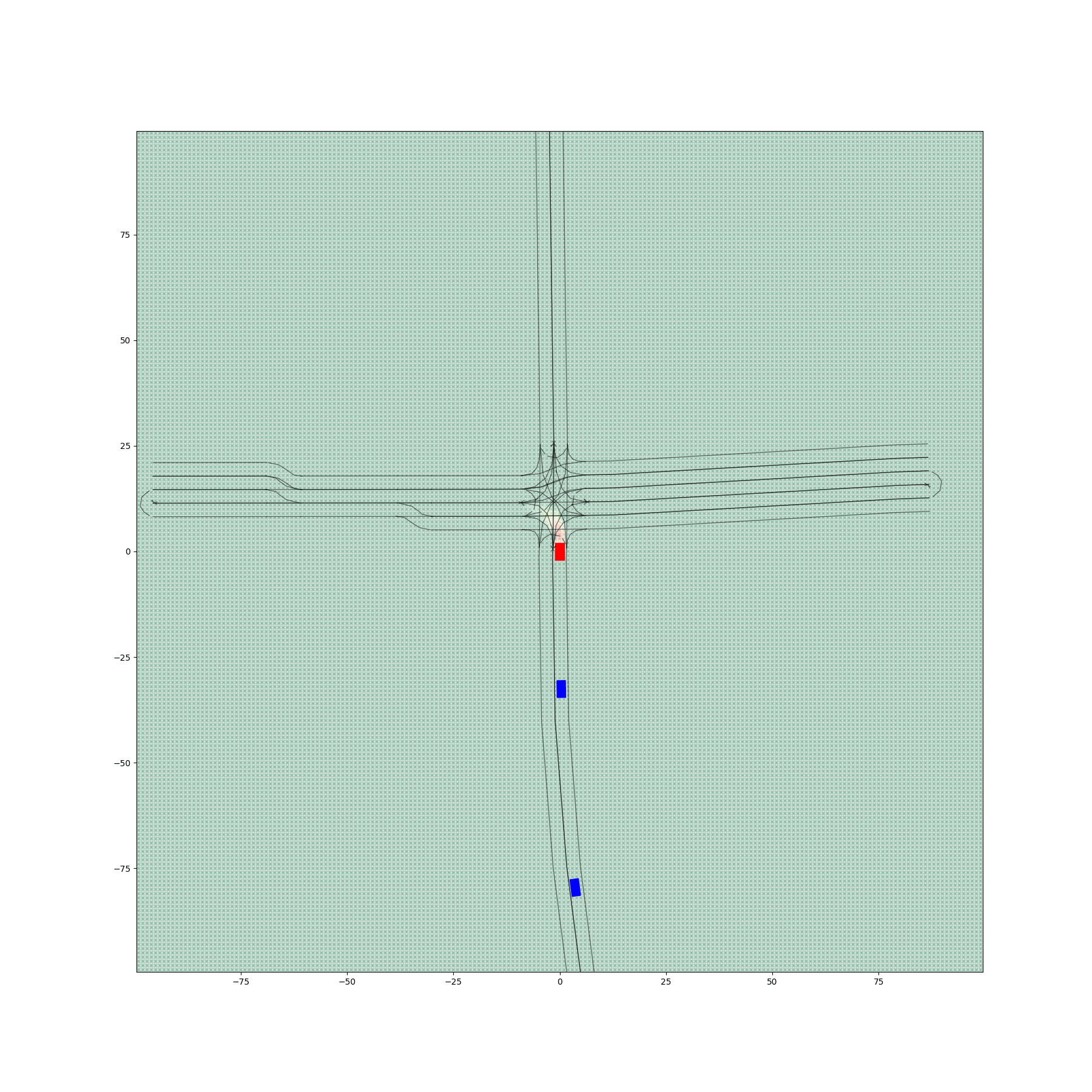}
        \caption{Encoder 0 + Decoder 2 \\ (0.0324\%)}
    \end{subfigure}

    \begin{subfigure}[b]{0.32\textwidth}
        \includegraphics[width=\textwidth]{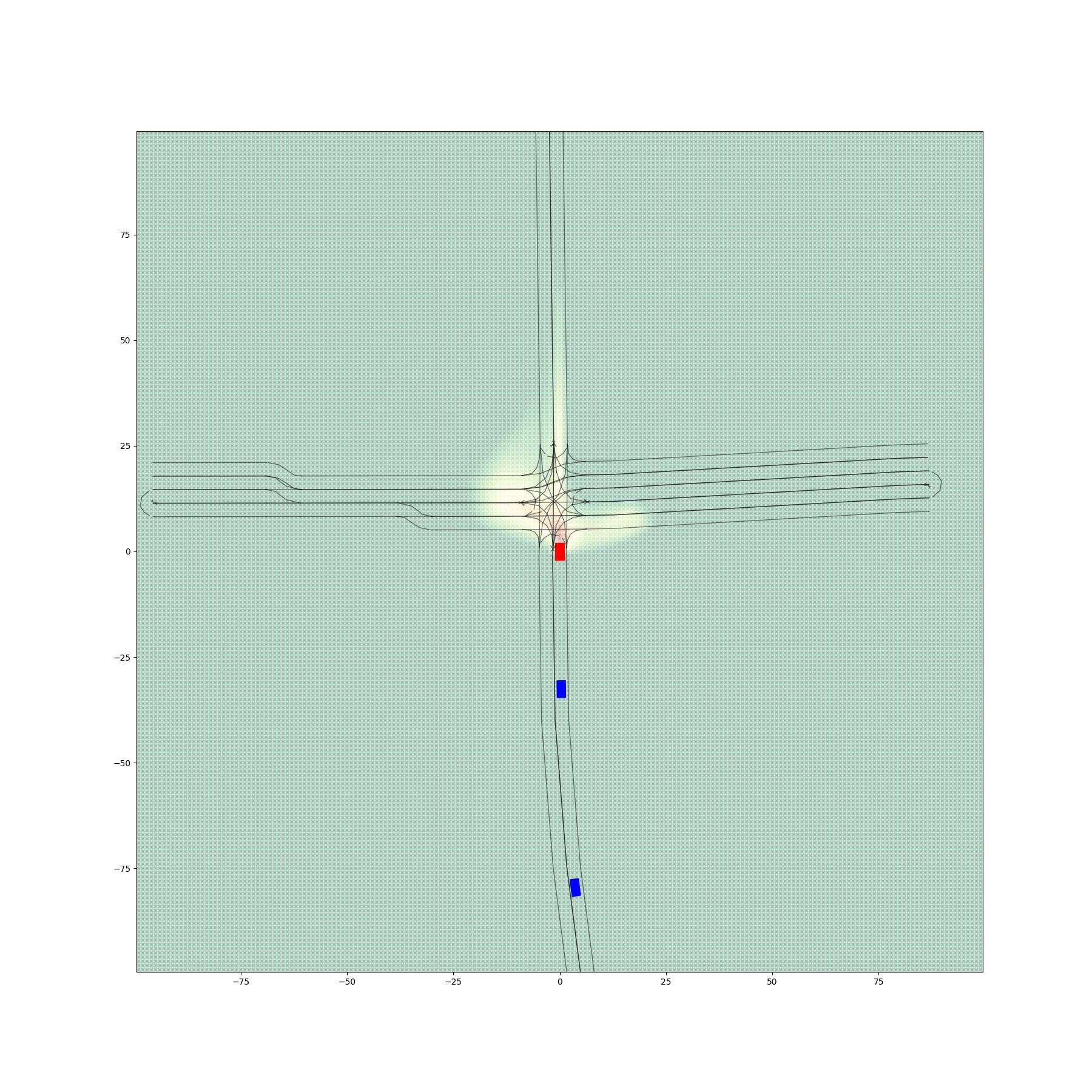}
        \caption{Encoder 1 + Decoder 0 \\ (3.9147\%)}
    \end{subfigure}
    \begin{subfigure}[b]{0.32\textwidth}
        \includegraphics[width=\textwidth]{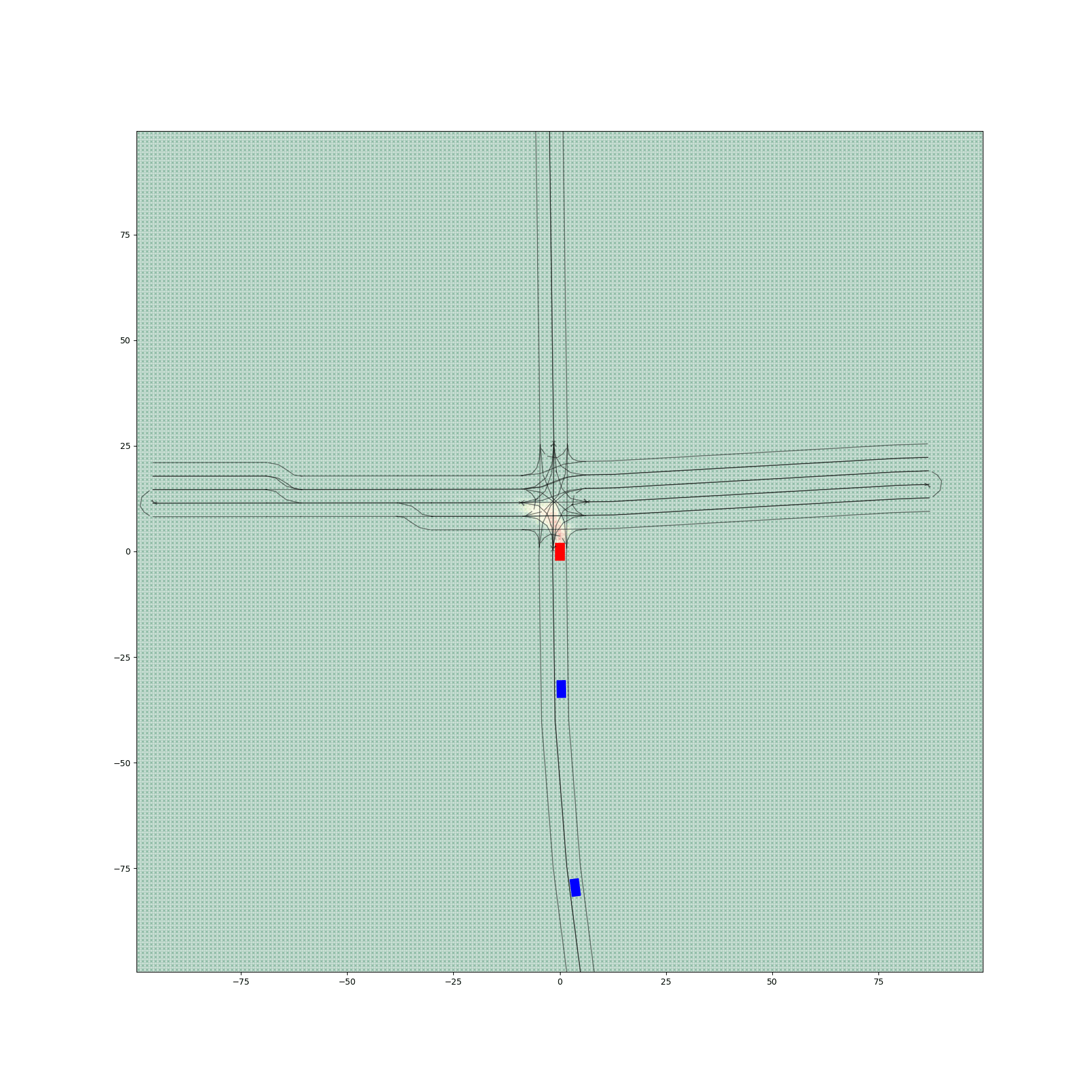}
        \caption{Encoder 1 + Decoder 1 \\ (0.0135\%)}
    \end{subfigure}
    \begin{subfigure}[b]{0.32\textwidth}
        \includegraphics[width=\textwidth]{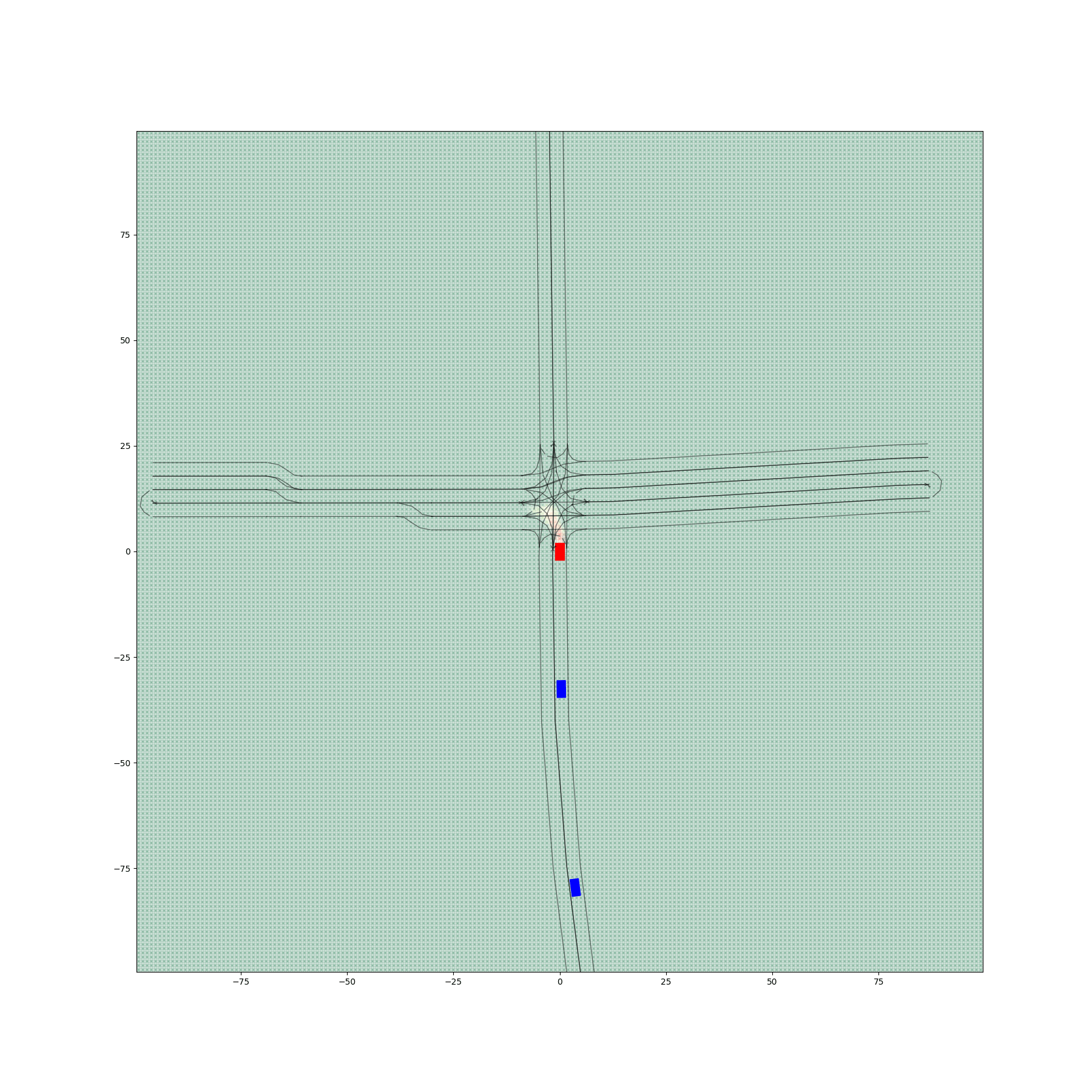}
        \caption{Encoder 1 + Decoder 2 \\ (0.0028\%)}
    \end{subfigure}
    \caption{Comparison of heatmaps generated by various encoder-decoder configurations in a left-turning scenario. This scenario evaluates the model’s ability to handle complex maneuvers and maintain spatial alignment.}
    \label{appendix_fig:2}
\end{figure*}

\begin{figure*}[h!]
    \centering
    \begin{subfigure}[b]{0.32\textwidth}
        \includegraphics[width=\textwidth]{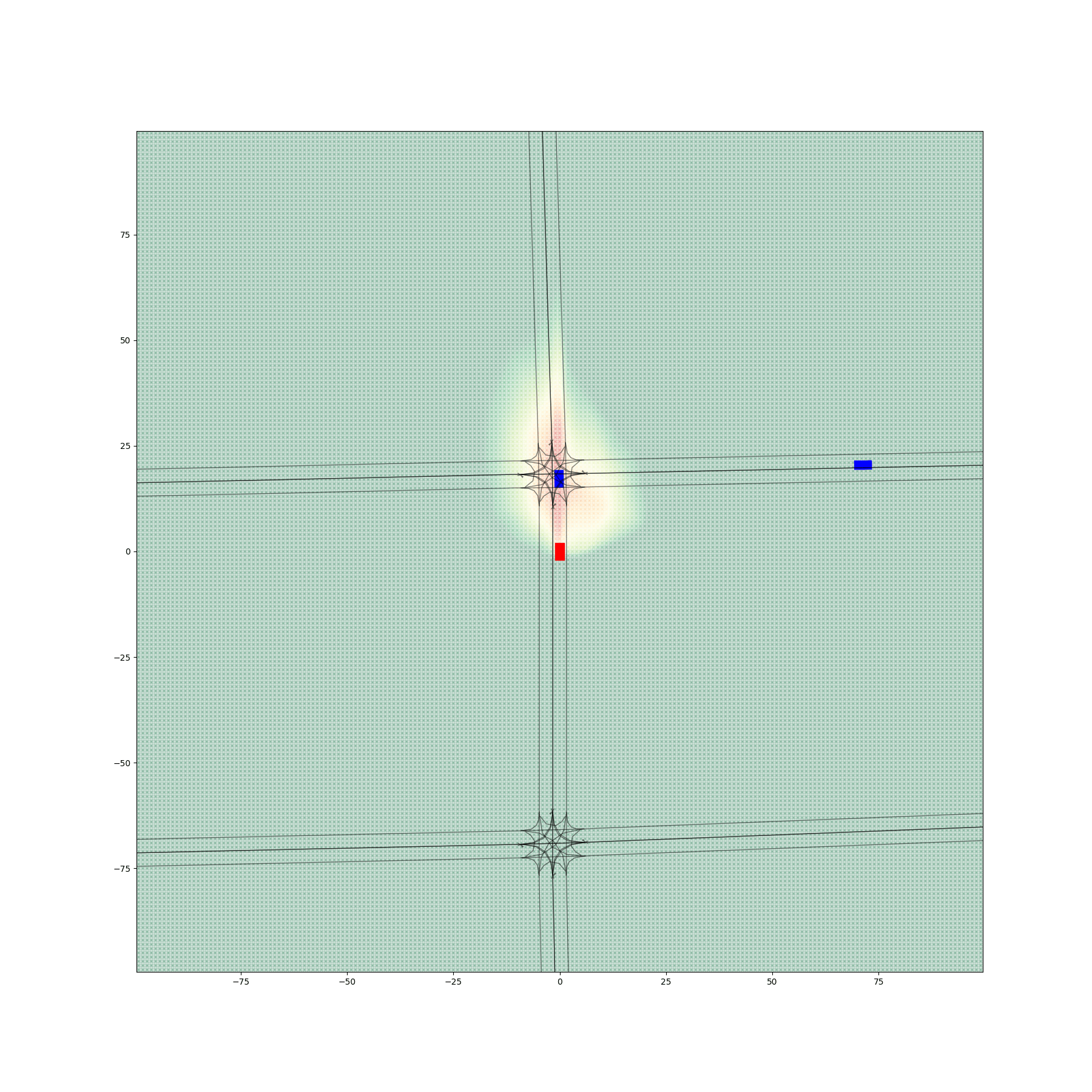}
        \caption{Encoder 0 + Decoder 0 \\ (7.3241\%)}
    \end{subfigure}
    \begin{subfigure}[b]{0.32\textwidth}
        \includegraphics[width=\textwidth]{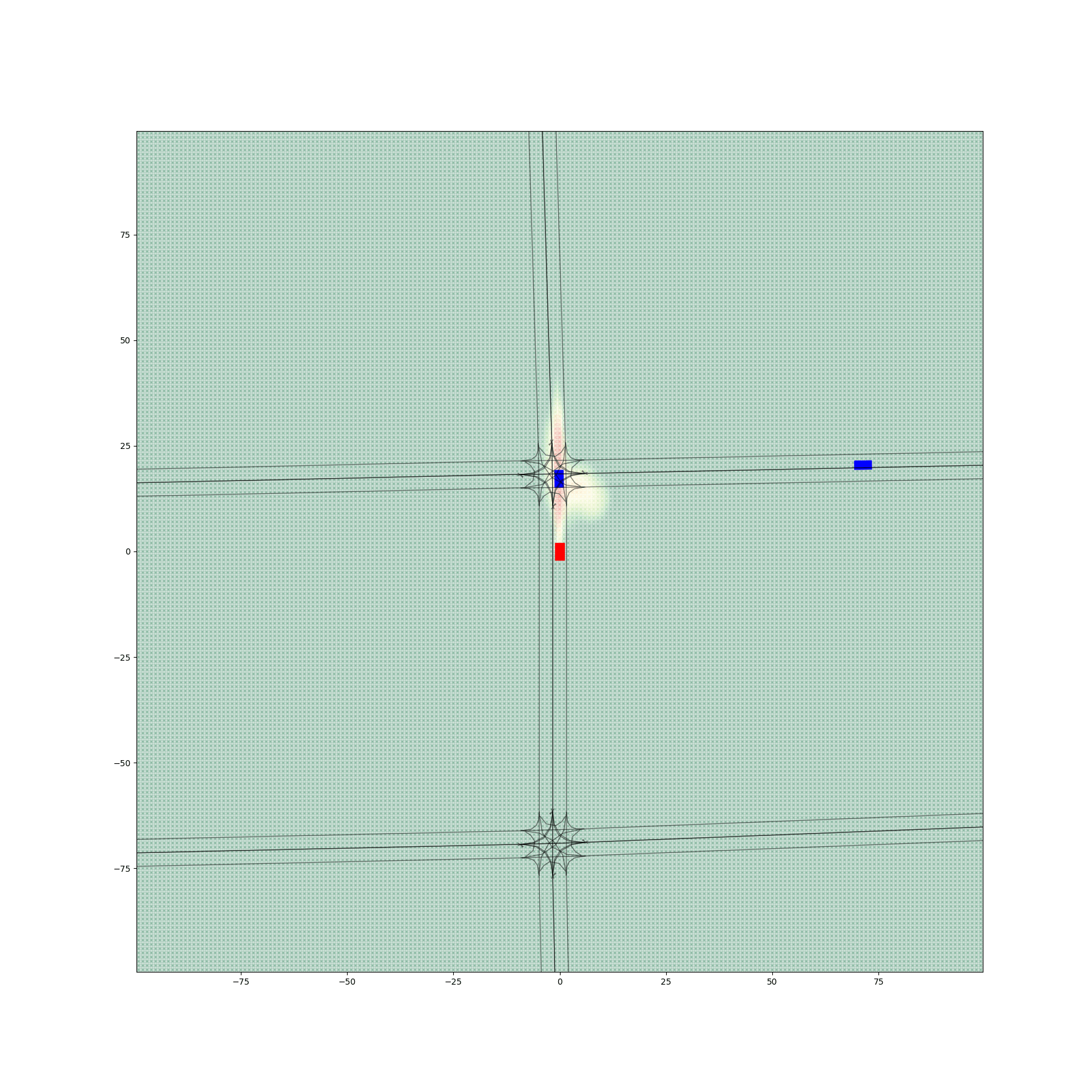}
        \caption{Encoder 0 + Decoder 1 \\ (2.1254\%)}
    \end{subfigure}
    \begin{subfigure}[b]{0.32\textwidth}
        \includegraphics[width=\textwidth]{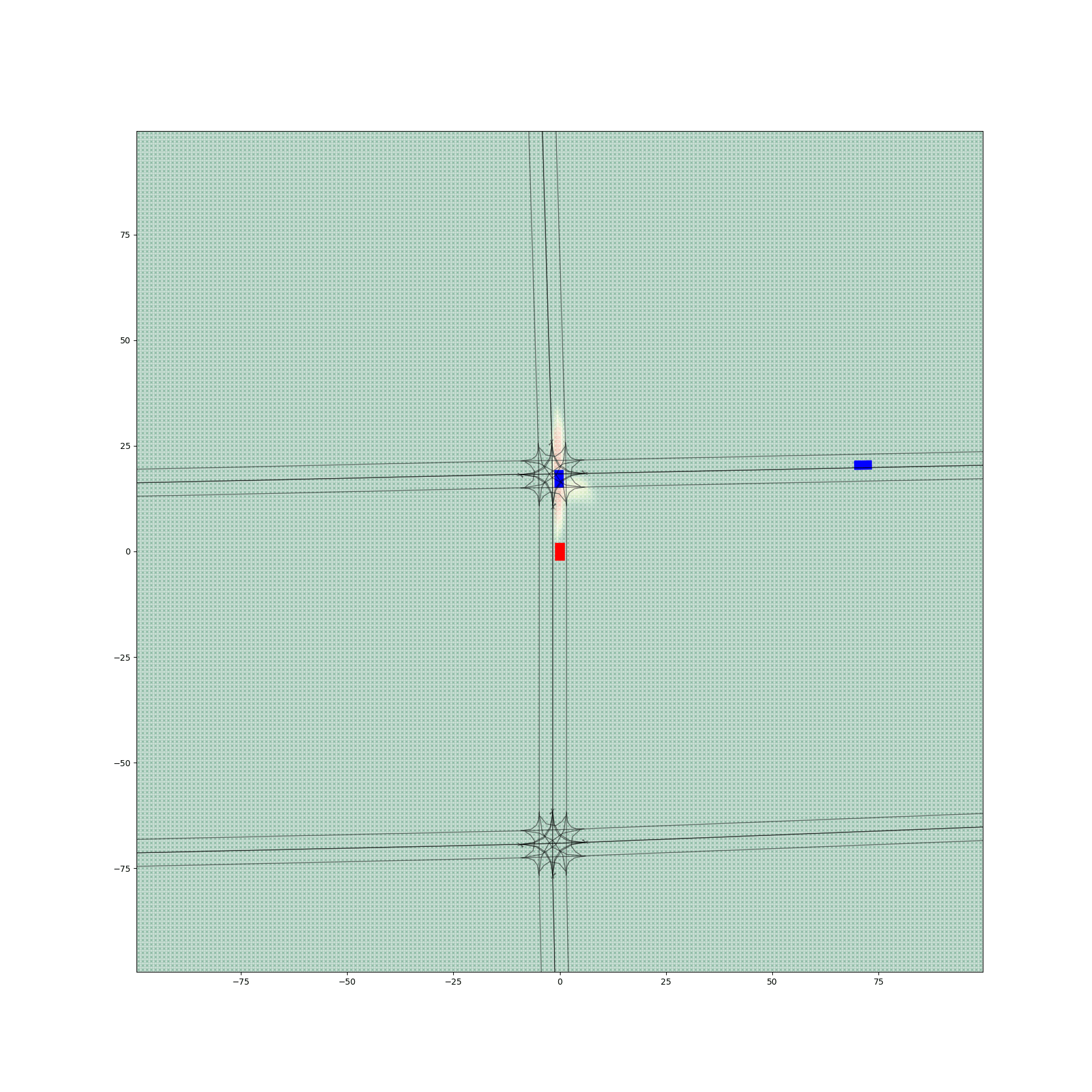}
        \caption{Encoder 0 + Decoder 2 \\ (0.0748\%)}
    \end{subfigure}

    \begin{subfigure}[b]{0.32\textwidth}
        \includegraphics[width=\textwidth]{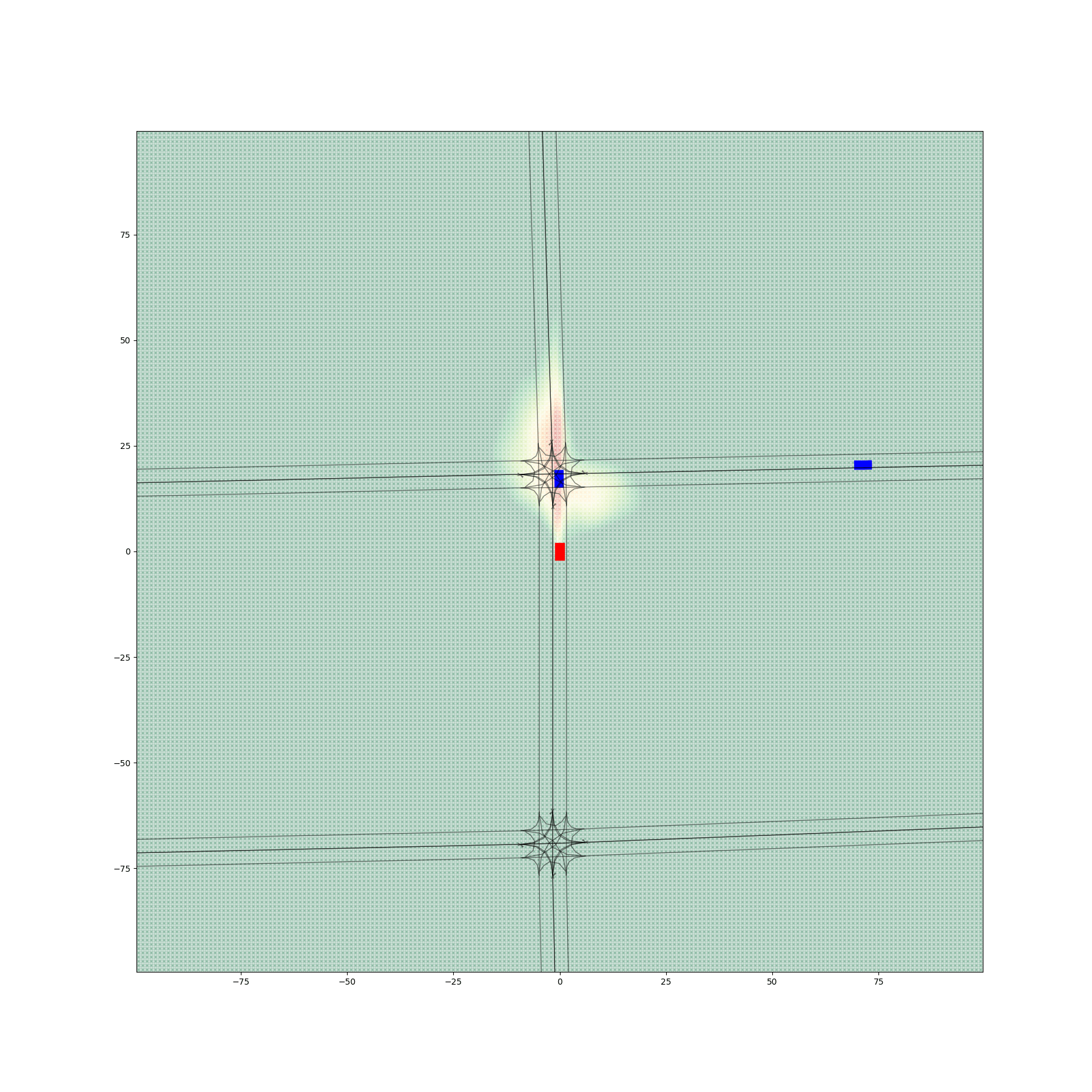}
        \caption{Encoder 1 + Decoder 0 \\ (7.0146\%)}
    \end{subfigure}
    \begin{subfigure}[b]{0.32\textwidth}
        \includegraphics[width=\textwidth]{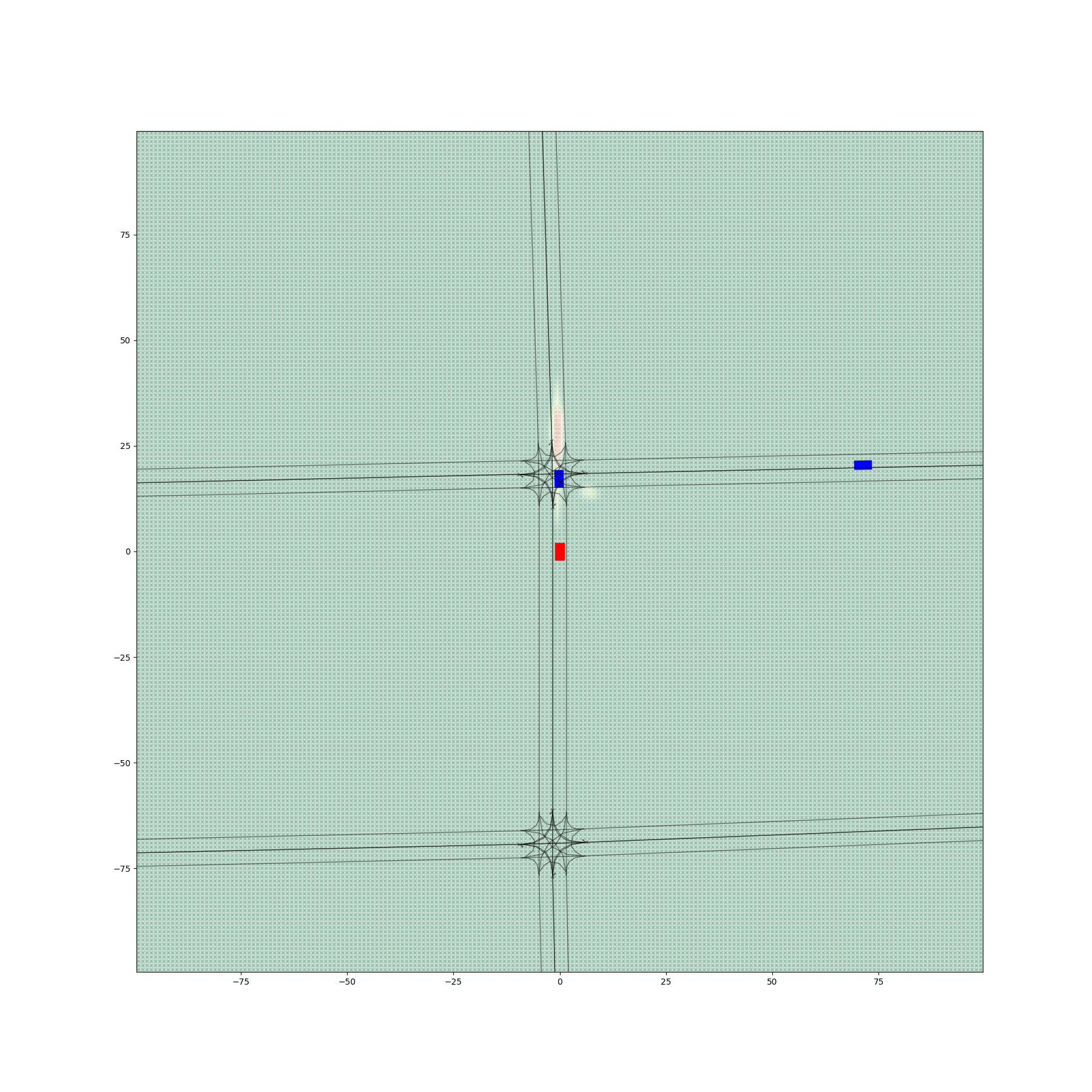}
        \caption{Encoder 1 + Decoder 1 \\ (0.6487\%)}
    \end{subfigure}
    \begin{subfigure}[b]{0.32\textwidth}
        \includegraphics[width=\textwidth]{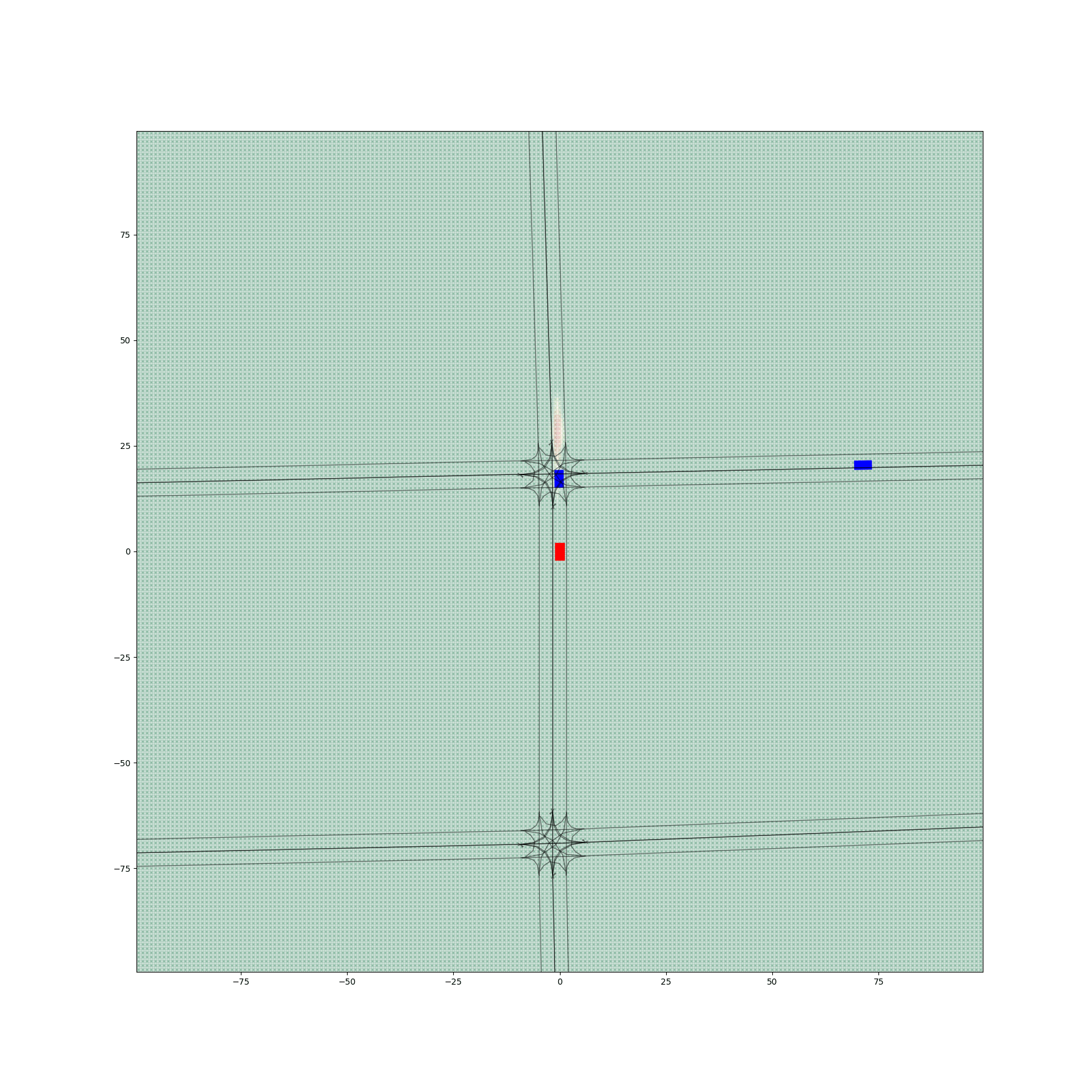}
        \caption{Encoder 1 + Decoder 2 \\ (0.0352\%)}
    \end{subfigure}
    \caption{Comparison of heatmaps generated by various encoder-decoder configurations at an intersection, focusing on the awareness of a leading vehicle in the middle. This scenario evaluates the models’ ability to accurately localize the leading vehicle while minimizing false predictions in the surroundings. }
    \label{appendix_fig:3}
\end{figure*}

\begin{figure*}[h!]
    \centering
    \begin{subfigure}[b]{0.32\textwidth}
        \includegraphics[width=\textwidth]{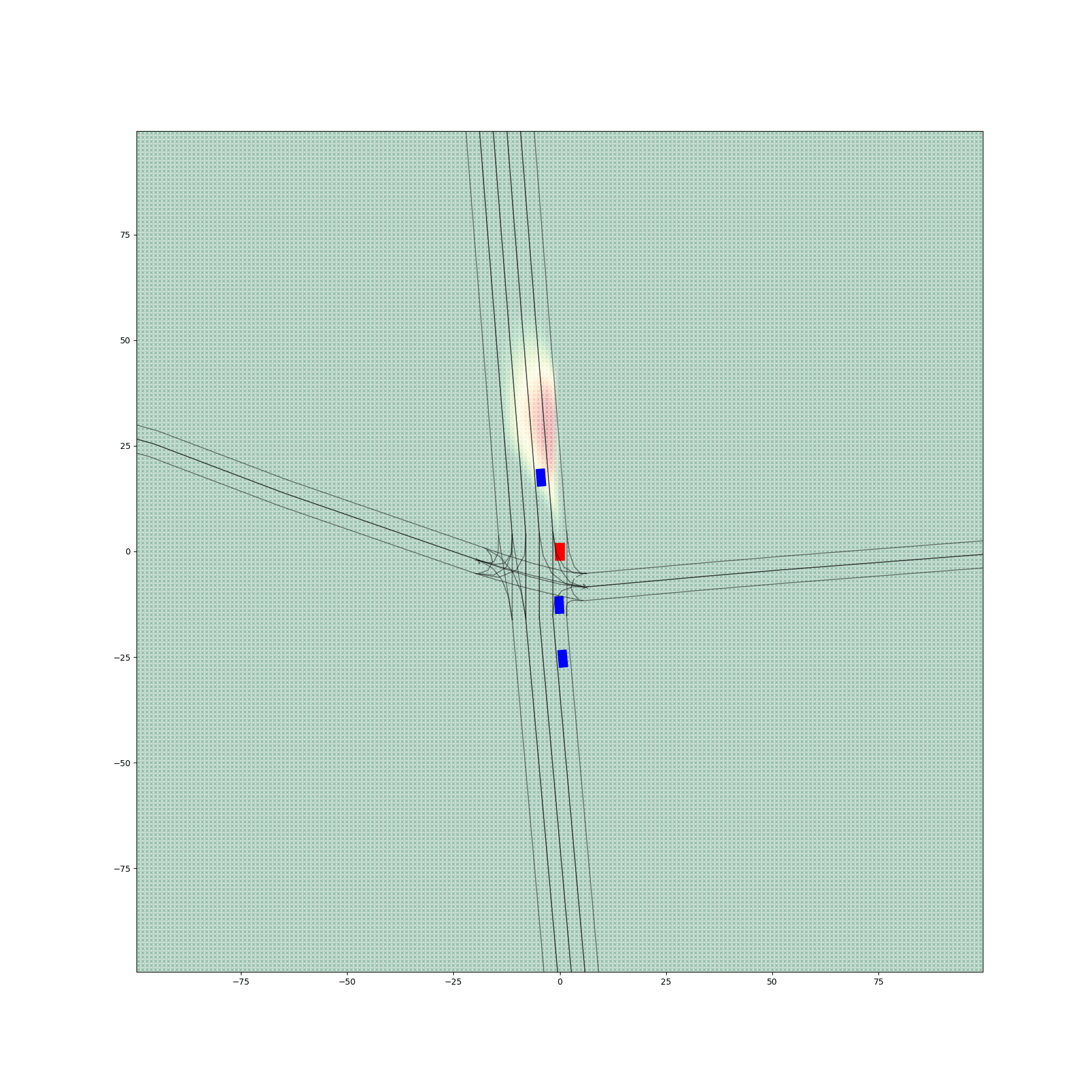}
        \caption{Encoder 0 + Decoder 0 \\ (1.8354\%)}
    \end{subfigure}
    \begin{subfigure}[b]{0.32\textwidth}
        \includegraphics[width=\textwidth]{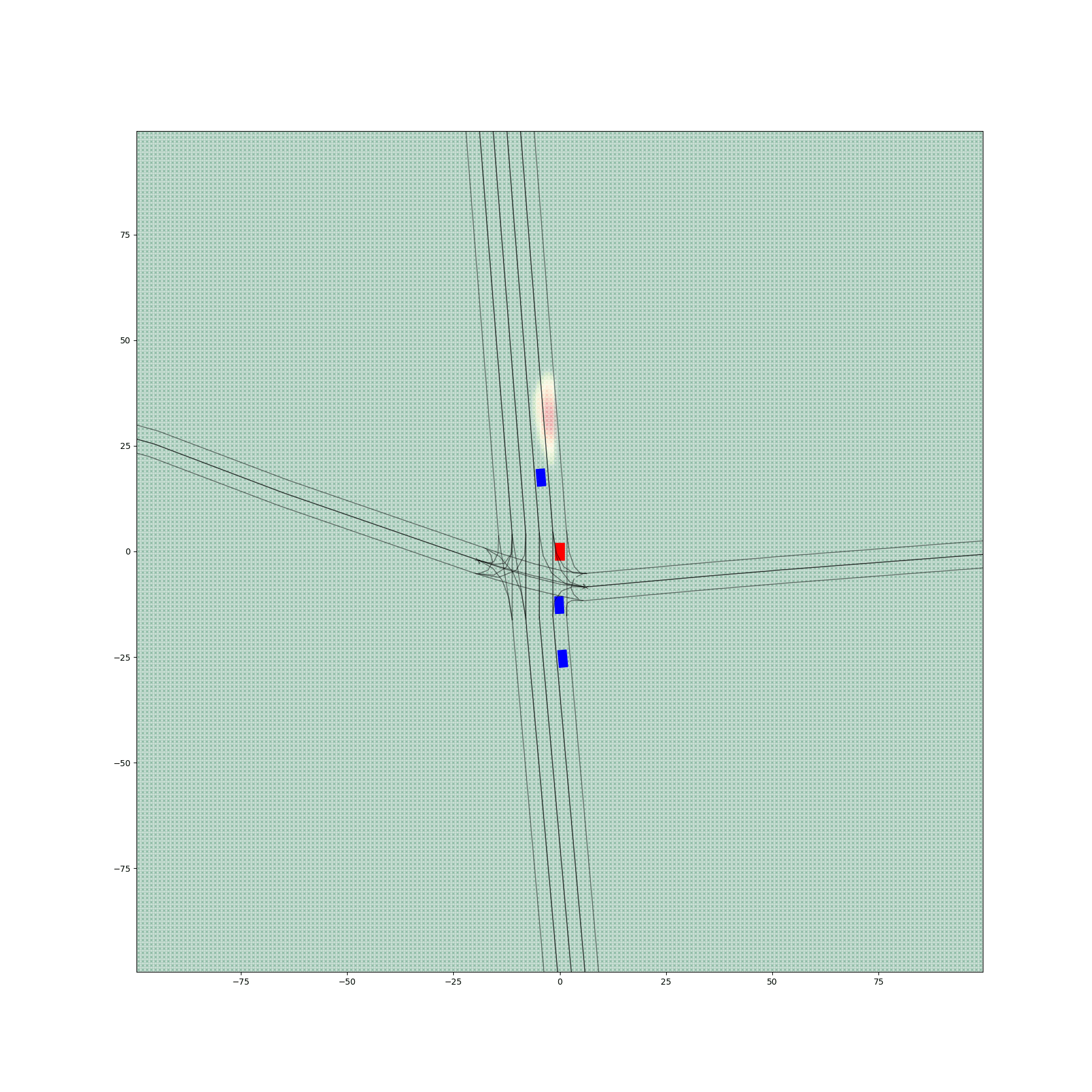}
        \caption{Encoder 0 + Decoder 1 \\ (1.4247\%)}
    \end{subfigure}
    \begin{subfigure}[b]{0.32\textwidth}
        \includegraphics[width=\textwidth]{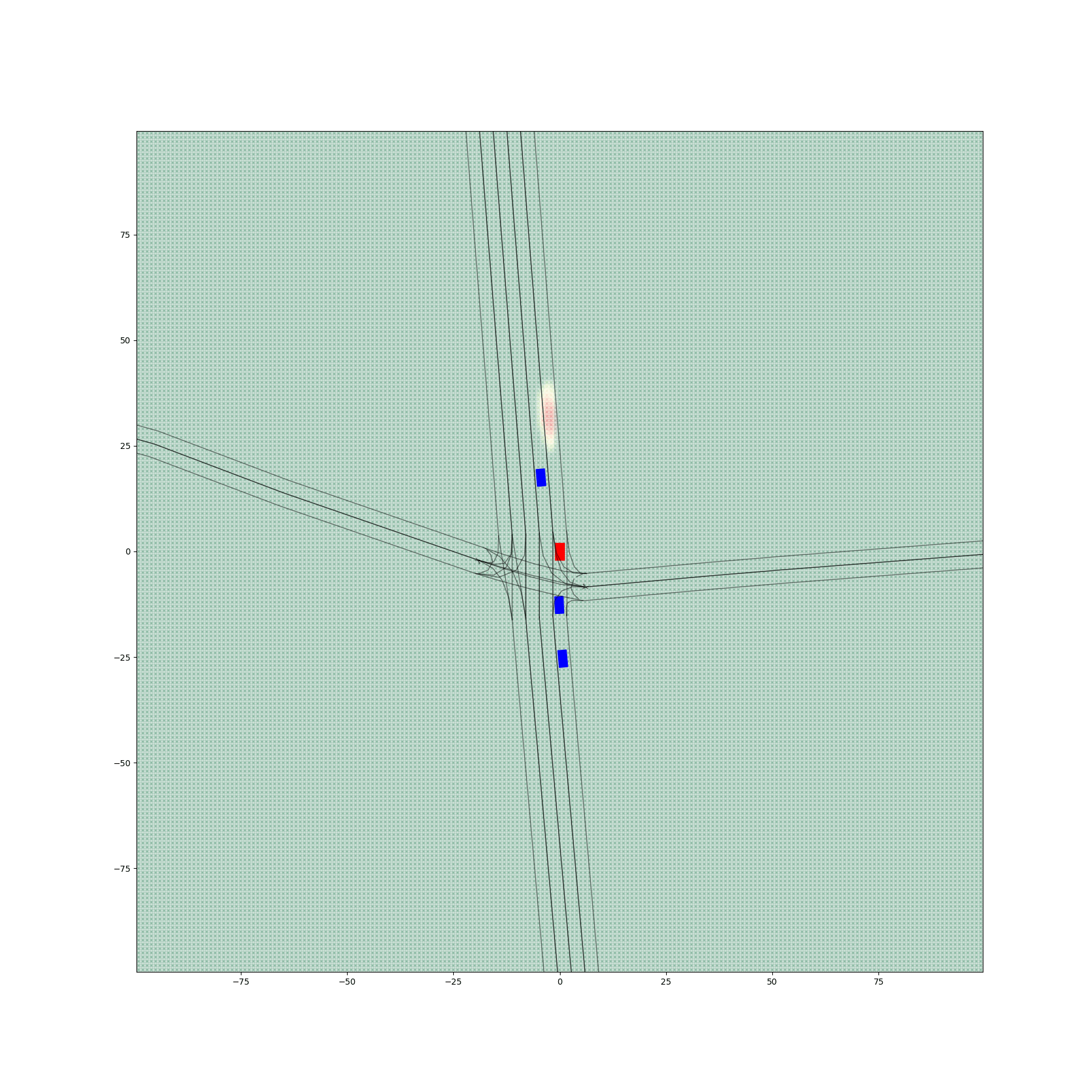}
        \caption{Encoder 0 + Decoder 2 \\ (0.0645\%)}
    \end{subfigure}

    \begin{subfigure}[b]{0.32\textwidth}
        \includegraphics[width=\textwidth]{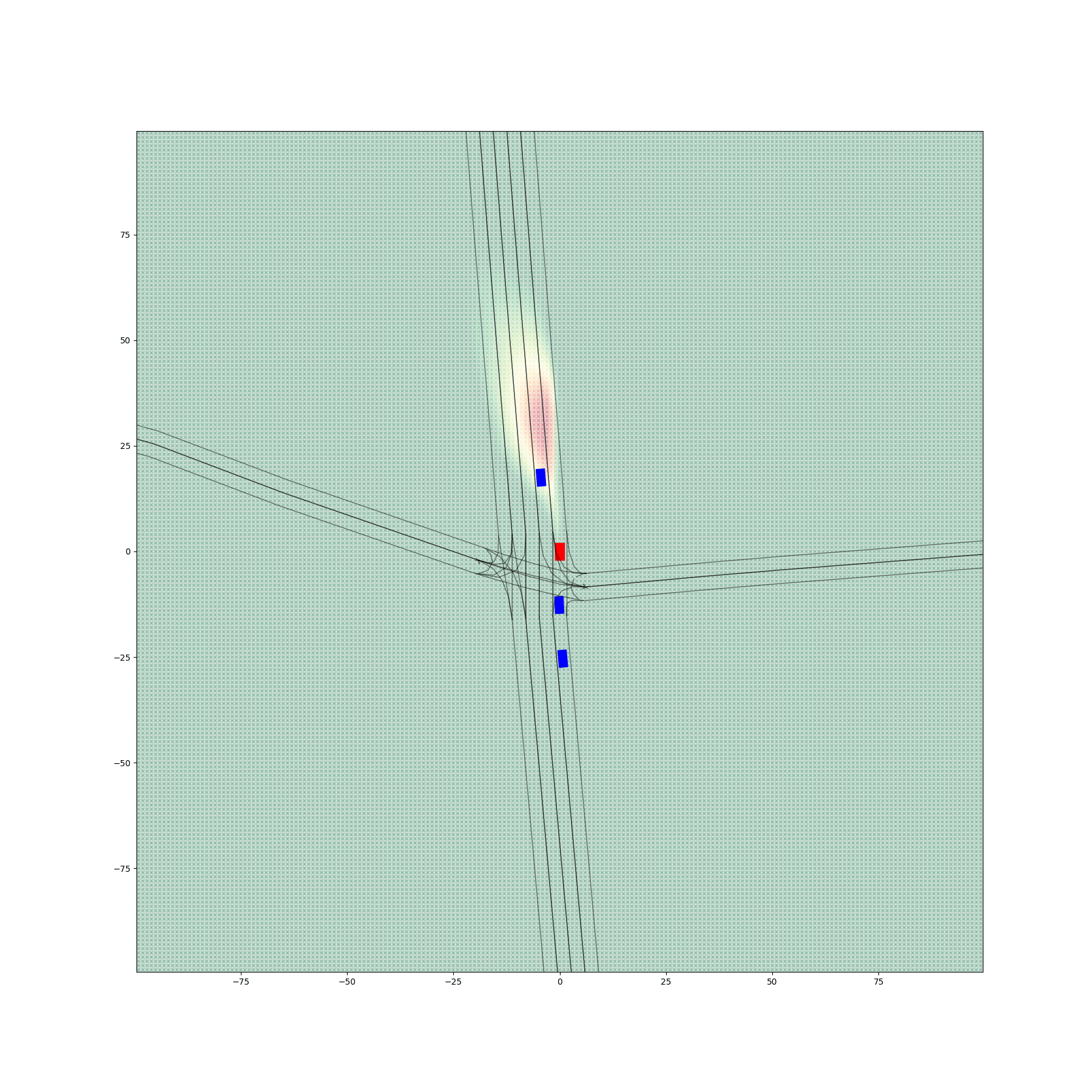}
        \caption{Encoder 1 + Decoder 0 \\ (2.1428\%)}
    \end{subfigure}
    \begin{subfigure}[b]{0.32\textwidth}
        \includegraphics[width=\textwidth]{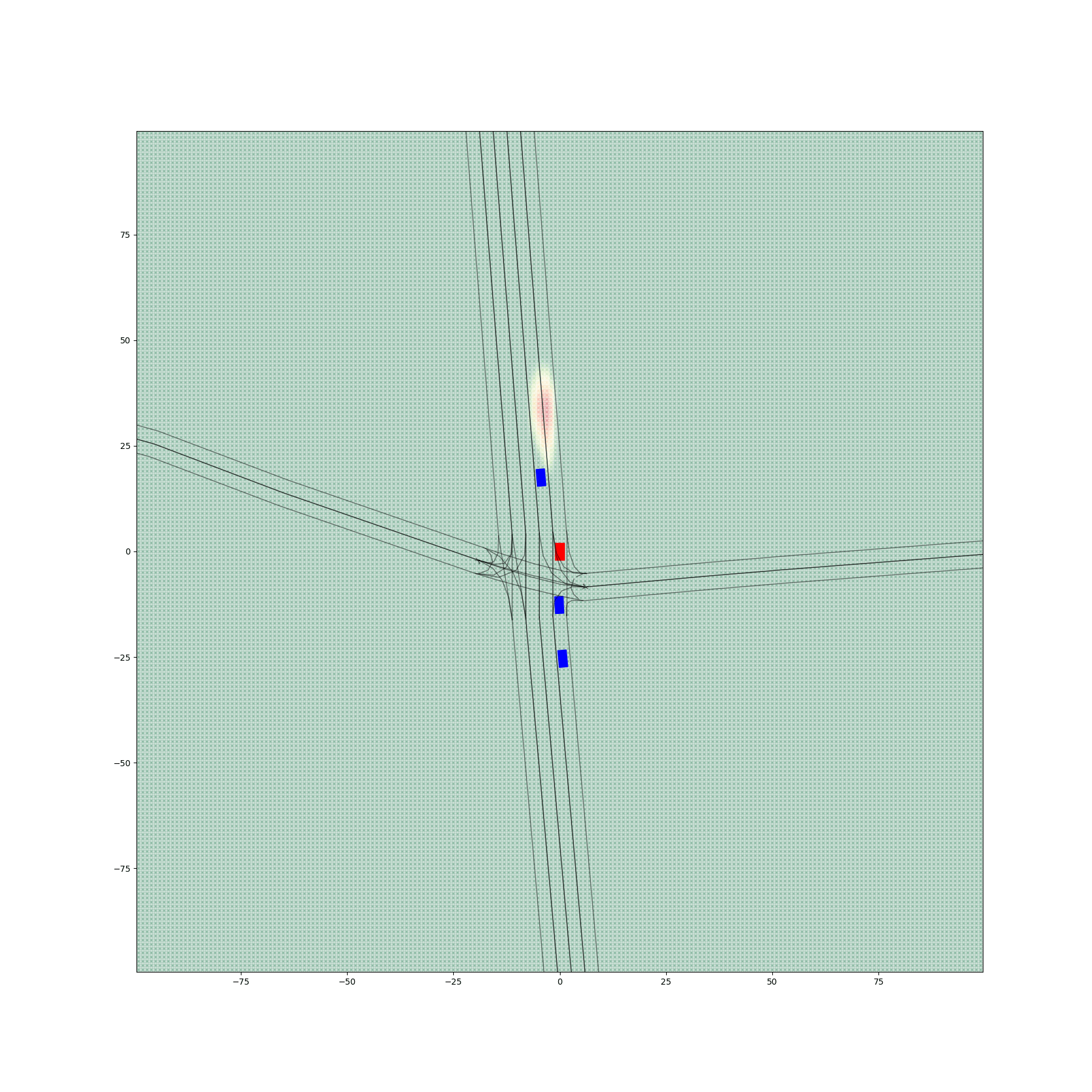}
        \caption{Encoder 1 + Decoder 1 \\ (1.2153\%)}
    \end{subfigure}
    \begin{subfigure}[b]{0.32\textwidth}
        \includegraphics[width=\textwidth]{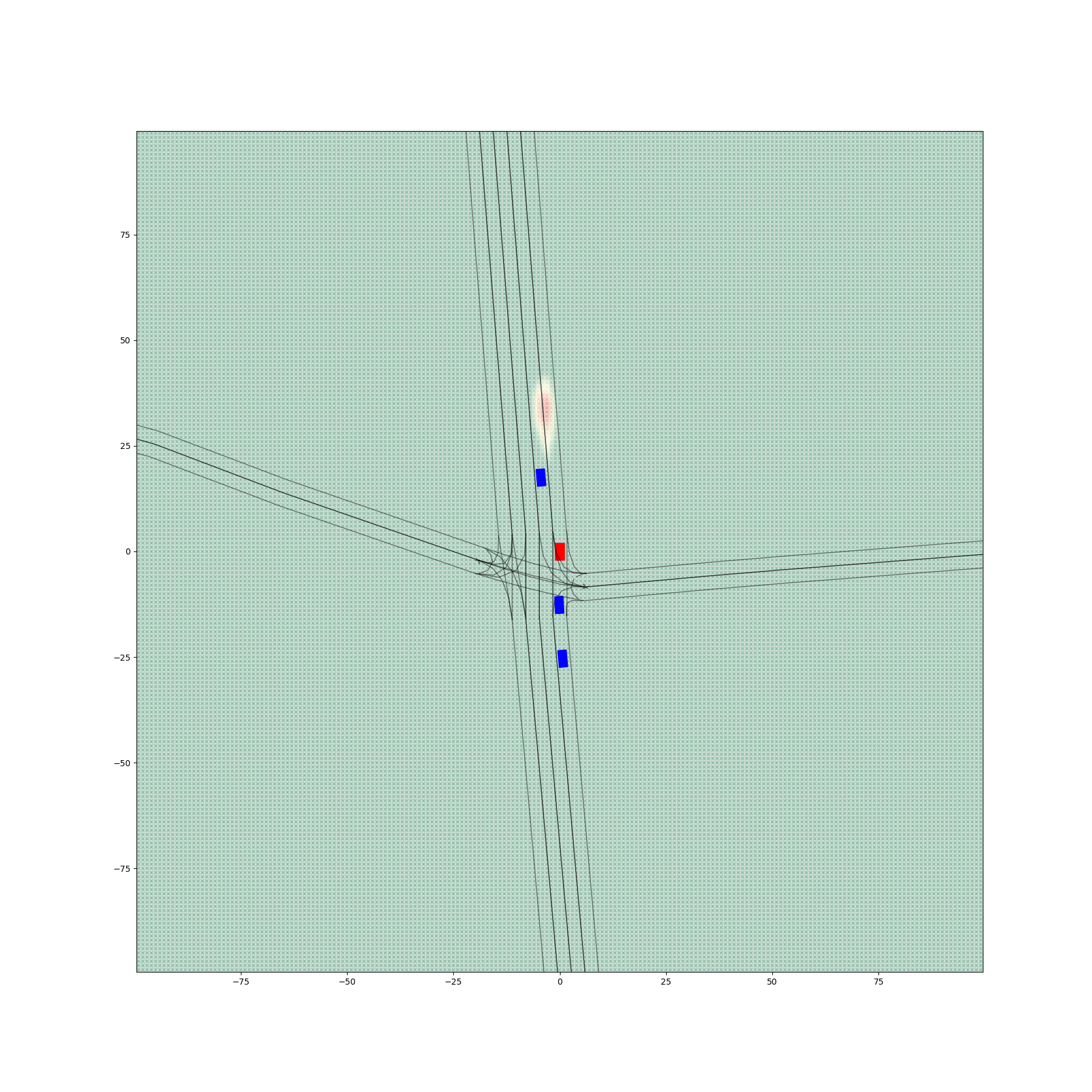}
        \caption{Encoder 1 + Decoder 2 \\ (0.0457\%)}
    \end{subfigure}
    \caption{Comparison of heatmaps generated by various encoder-decoder configurations for lane-changing planning, focusing on awareness of the leading vehicle on the target lane. This scenario highlights the model’s ability to predict future positions accurately during a lane-change maneuver.}
    \label{appendix_fig:4}
\end{figure*}

\begin{figure*}[h!]
    \centering
    \begin{subfigure}[b]{0.32\textwidth}
        \includegraphics[width=\textwidth]{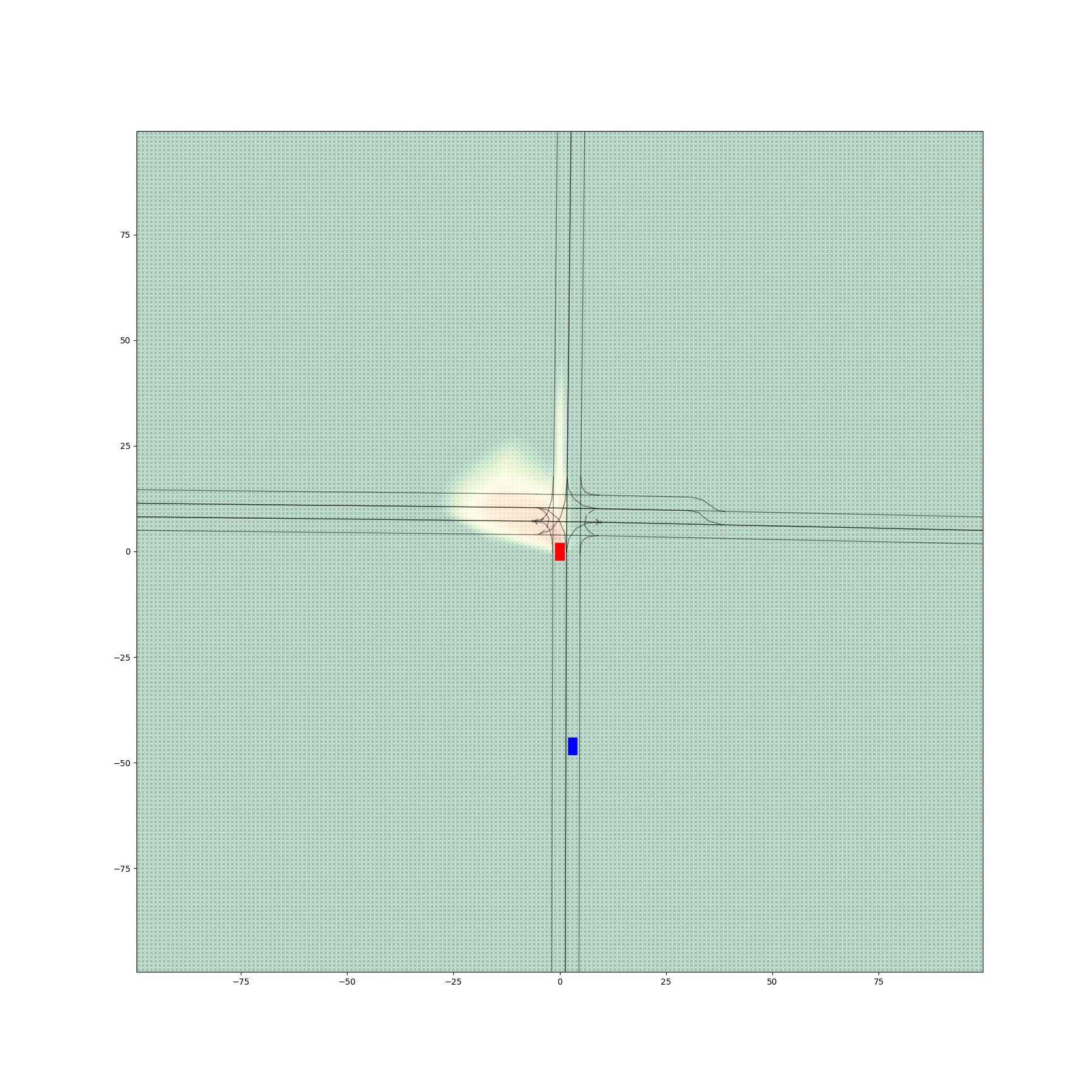}
        \caption{Encoder 0 + Decoder 0 \\ (6.8312\%)}
    \end{subfigure}
    \begin{subfigure}[b]{0.32\textwidth}
        \includegraphics[width=\textwidth]{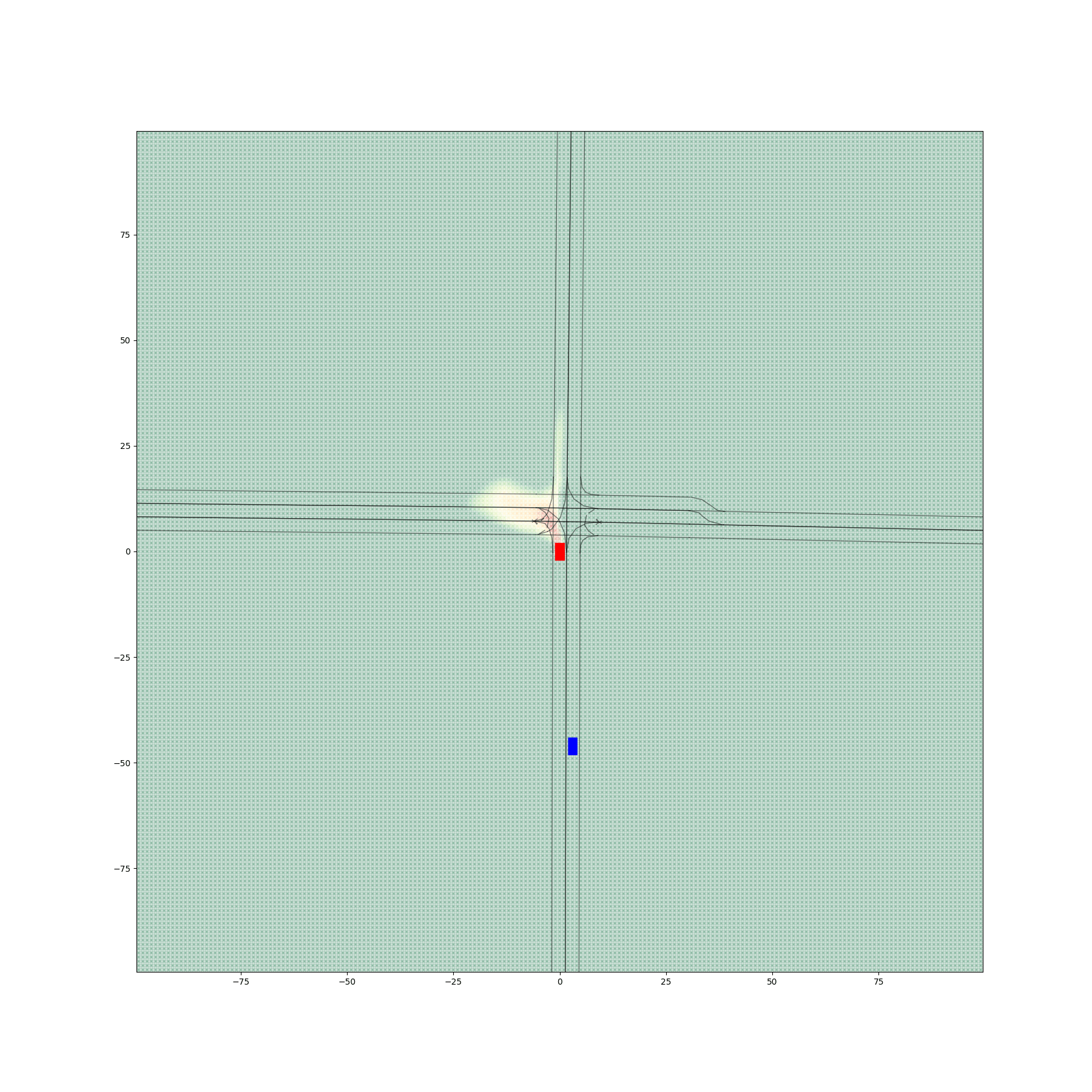}
        \caption{Encoder 0 + Decoder 1 \\ (2.6435\%)}
    \end{subfigure}
    \begin{subfigure}[b]{0.32\textwidth}
        \includegraphics[width=\textwidth]{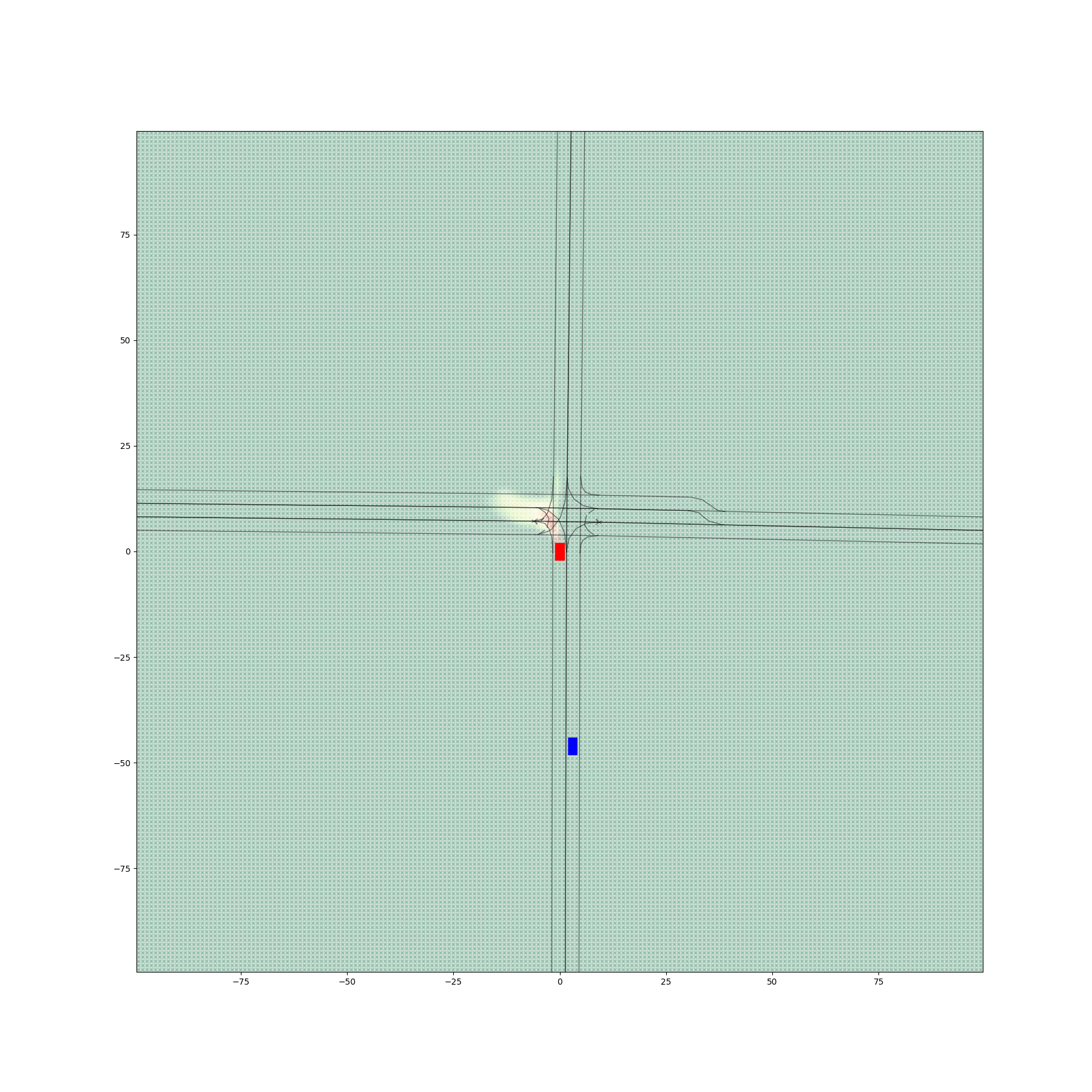}
        \caption{Encoder 0 + Decoder 2 \\ (0.6234\%)}
    \end{subfigure}

    \begin{subfigure}[b]{0.32\textwidth}
        \includegraphics[width=\textwidth]{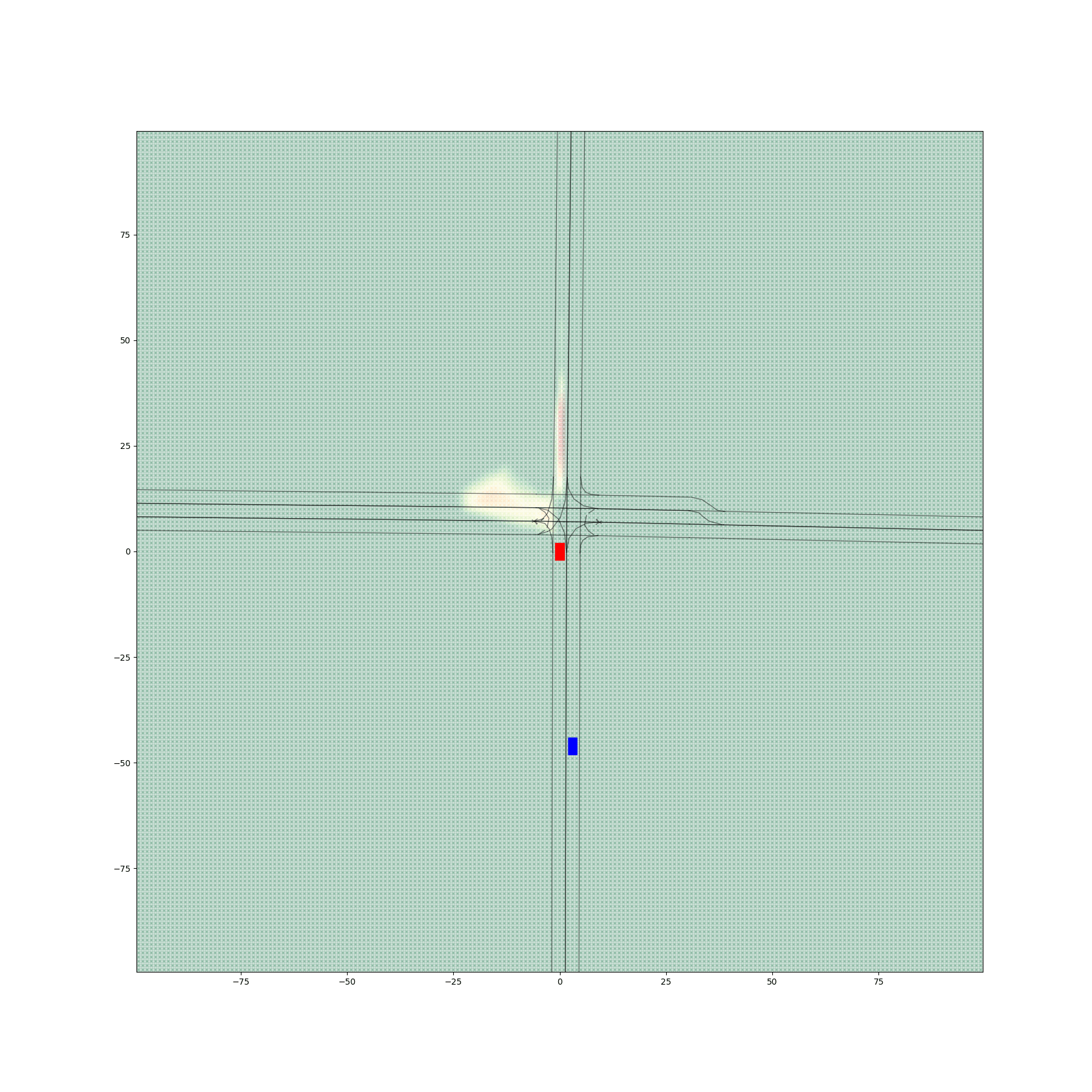}
        \caption{Encoder 1 + Decoder 0 \\ (4.2146\%)}
    \end{subfigure}
    \begin{subfigure}[b]{0.32\textwidth}
        \includegraphics[width=\textwidth]{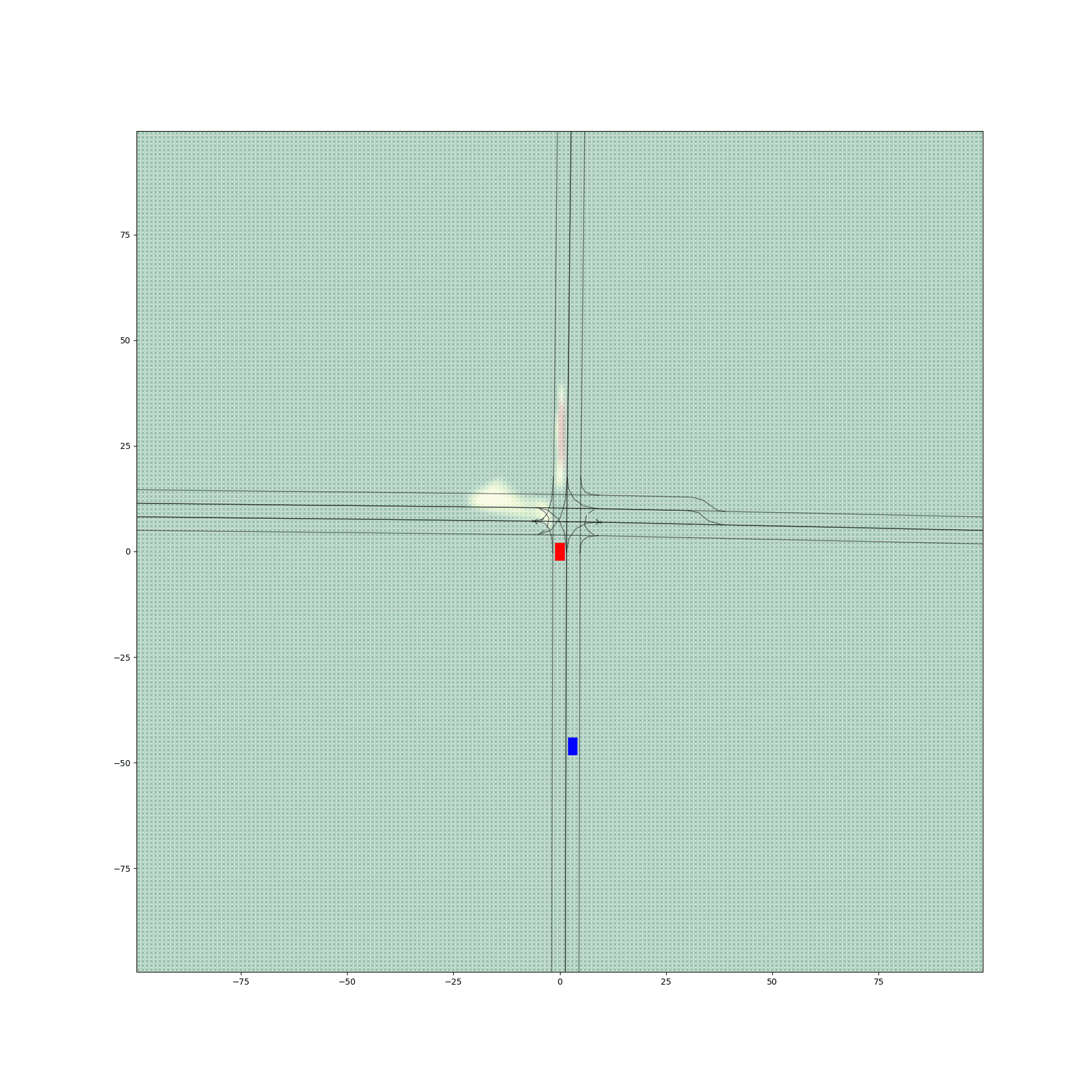}
        \caption{Encoder 1 + Decoder 1 \\ (2.9847\%)}
    \end{subfigure}
    \begin{subfigure}[b]{0.32\textwidth}
        \includegraphics[width=\textwidth]{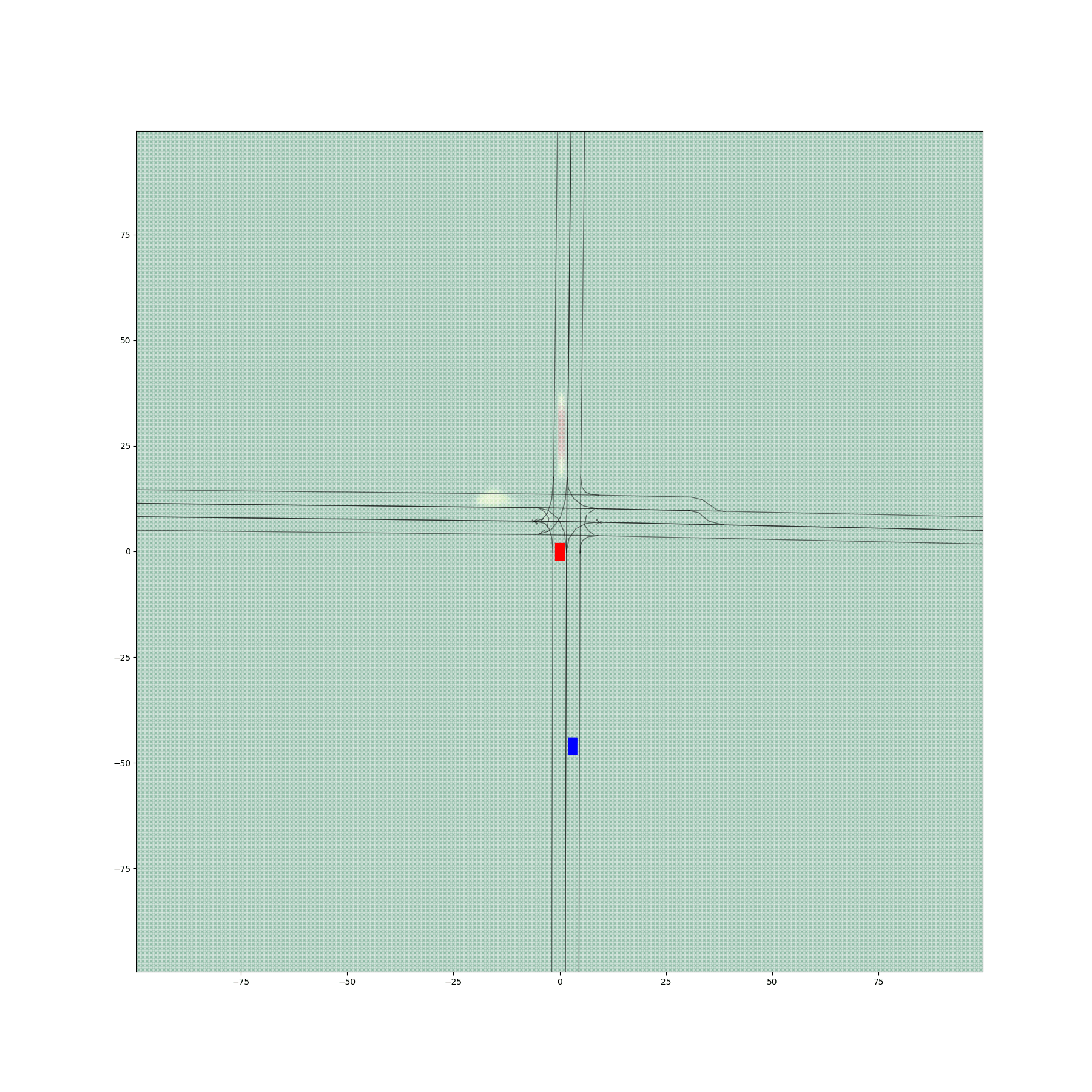}
        \caption{Encoder 1 + Decoder 2 \\ (1.1428\%)}
    \end{subfigure}
    \caption{Comparison of heatmaps generated by various encoder-decoder configurations for free left-turning from a single\edit{-}direction lane. This scenario evaluates the models’ ability to handle free turns without conflicting traffic and maintain accuracy in probabilistic occupancy prediction. }
    \label{appendix_fig:5}
\end{figure*}

\begin{figure*}[h!]
    \centering
    \begin{subfigure}[b]{0.32\textwidth}
        \includegraphics[width=\textwidth]{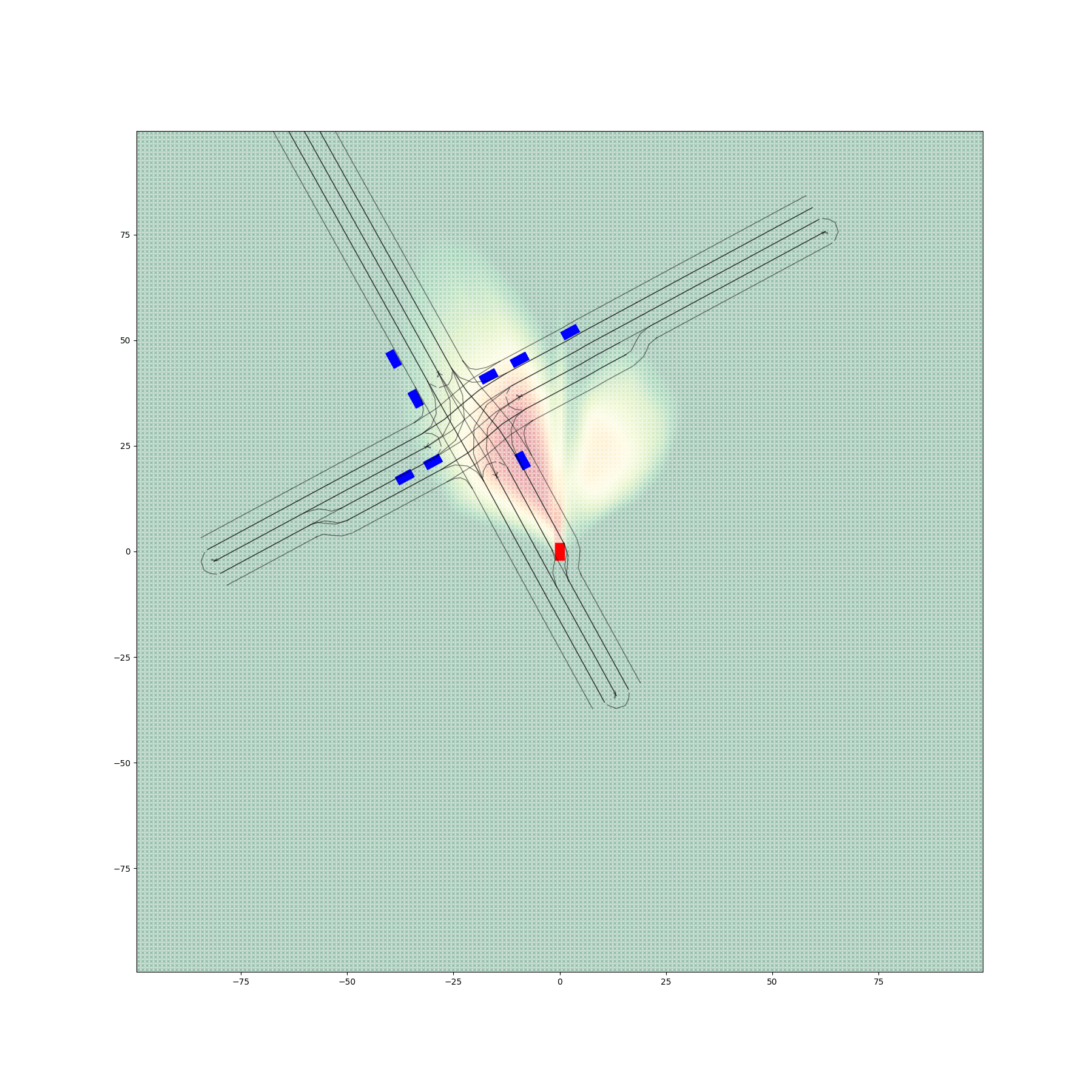}
        \caption{Encoder 0 + Decoder 0 \\ (19.8423\%)}
    \end{subfigure}
    \begin{subfigure}[b]{0.32\textwidth}
        \includegraphics[width=\textwidth]{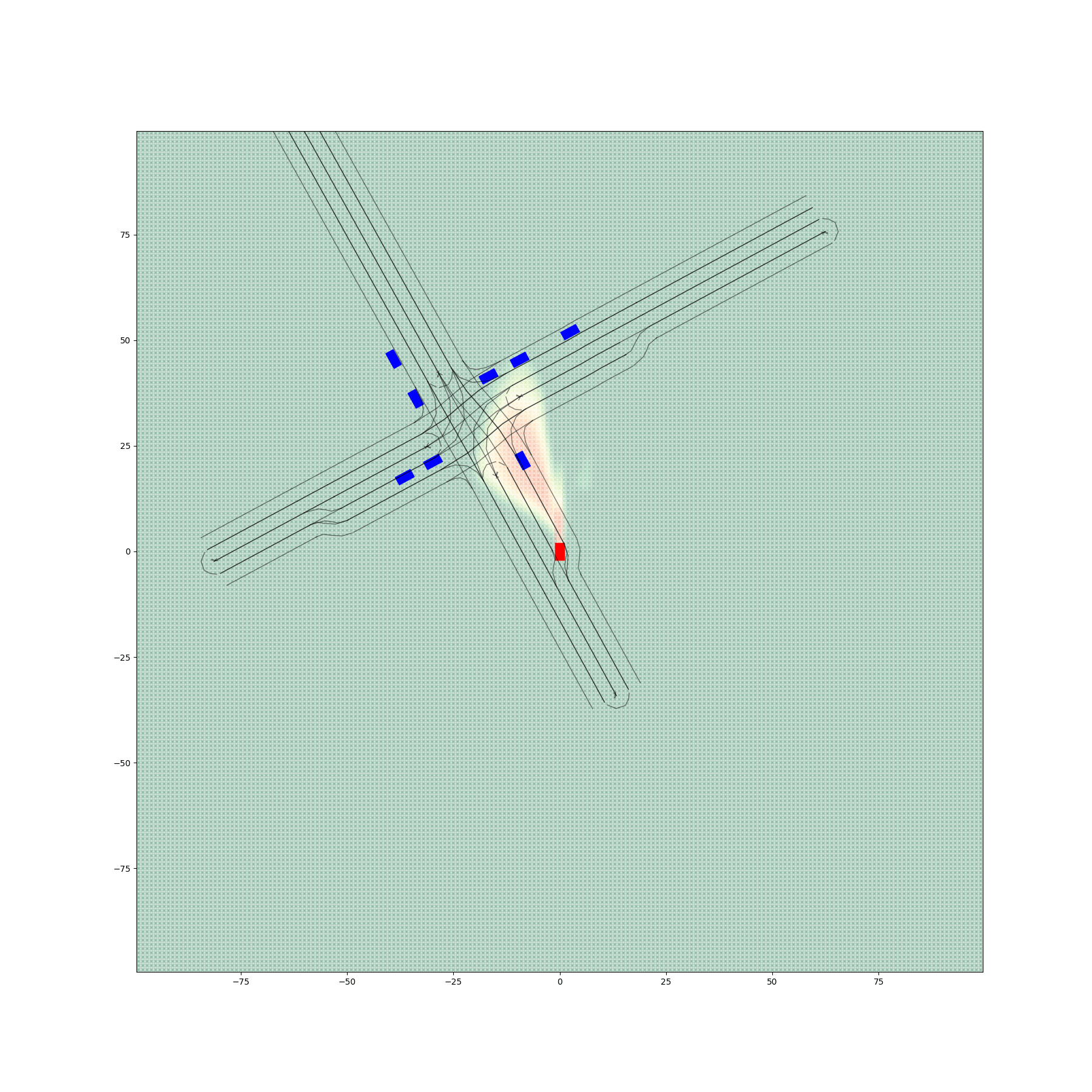}
        \caption{Encoder 0 + Decoder 1 \\ (6.3746\%)}
    \end{subfigure}
    \begin{subfigure}[b]{0.32\textwidth}
        \includegraphics[width=\textwidth]{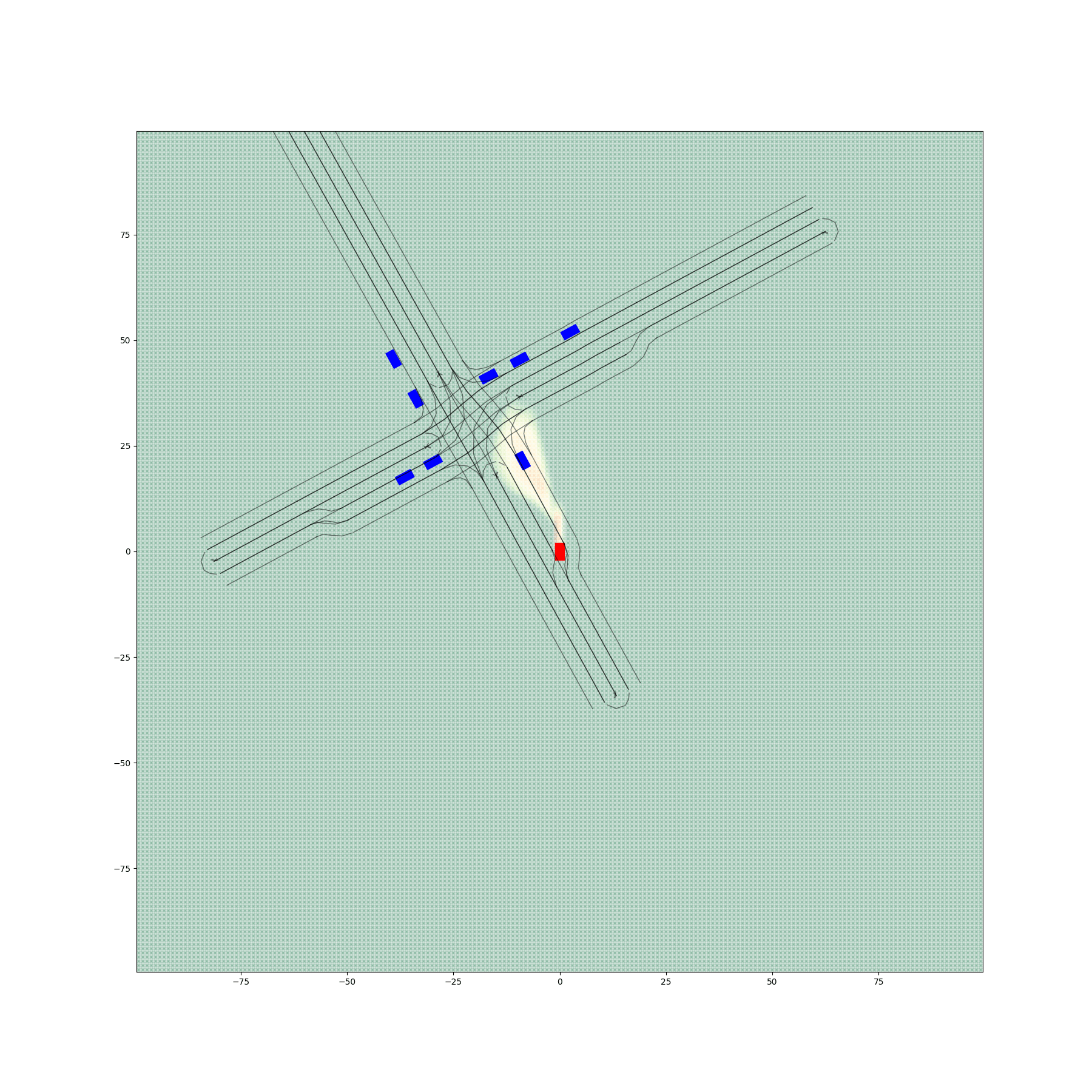}
        \caption{Encoder 0 + Decoder 2 \\ (3.1745\%)}
    \end{subfigure}

    \begin{subfigure}[b]{0.32\textwidth}
        \includegraphics[width=\textwidth]{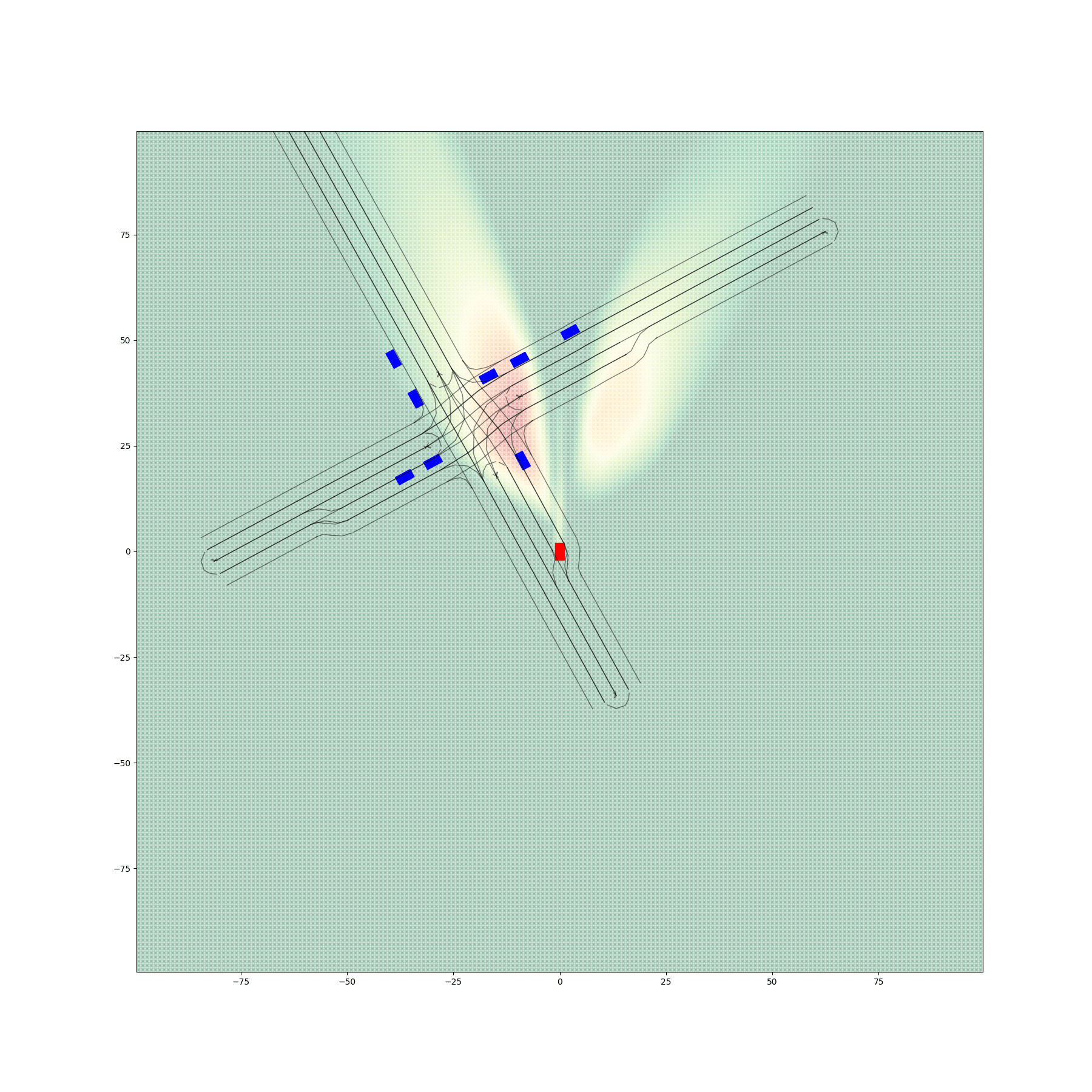}
        \caption{Encoder 1 + Decoder 0 \\ (23.2751\%)}
    \end{subfigure}
    \begin{subfigure}[b]{0.32\textwidth}
        \includegraphics[width=\textwidth]{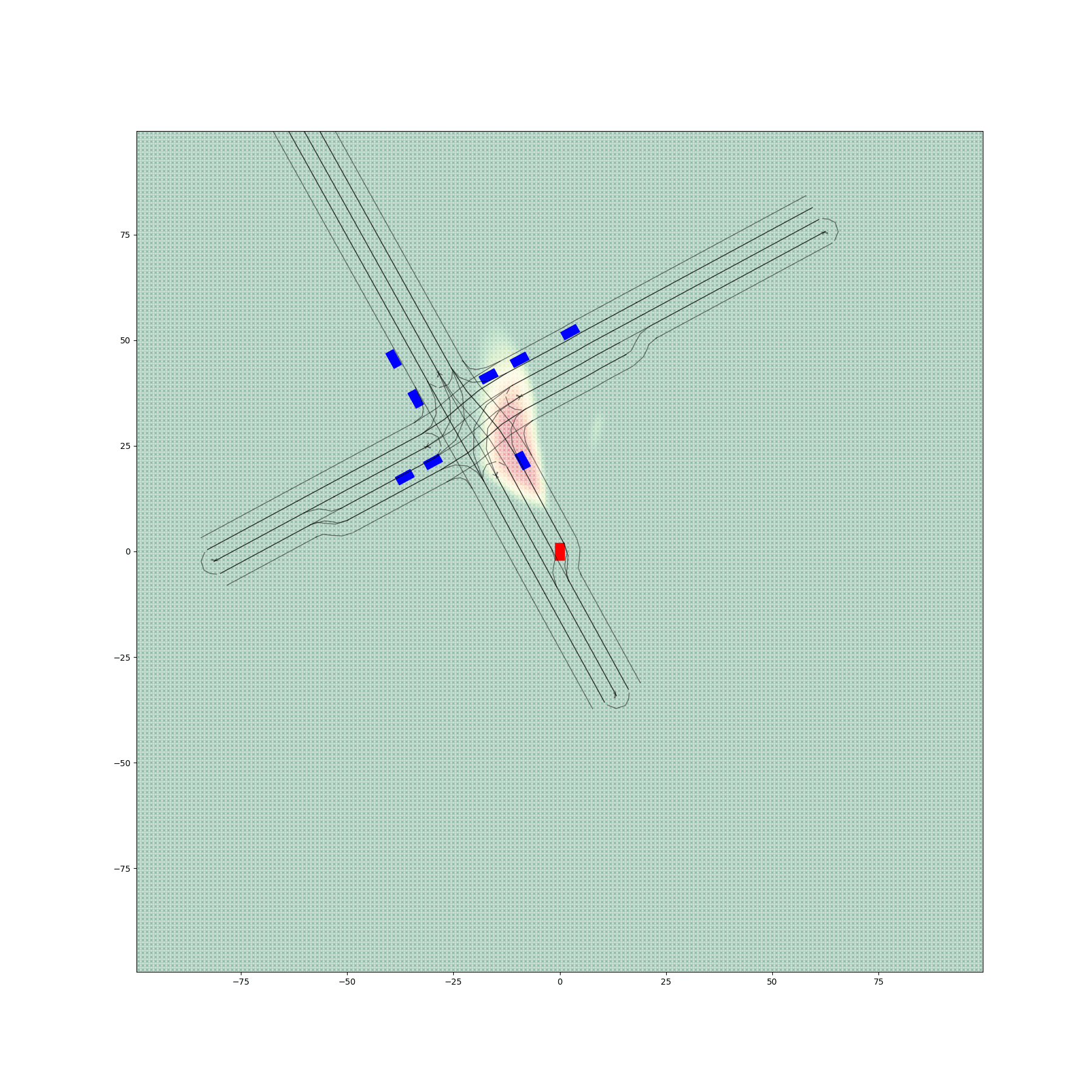}
        \caption{Encoder 1 + Decoder 1 \\ (6.2957\%)}
    \end{subfigure}
    \begin{subfigure}[b]{0.32\textwidth}
        \includegraphics[width=\textwidth]{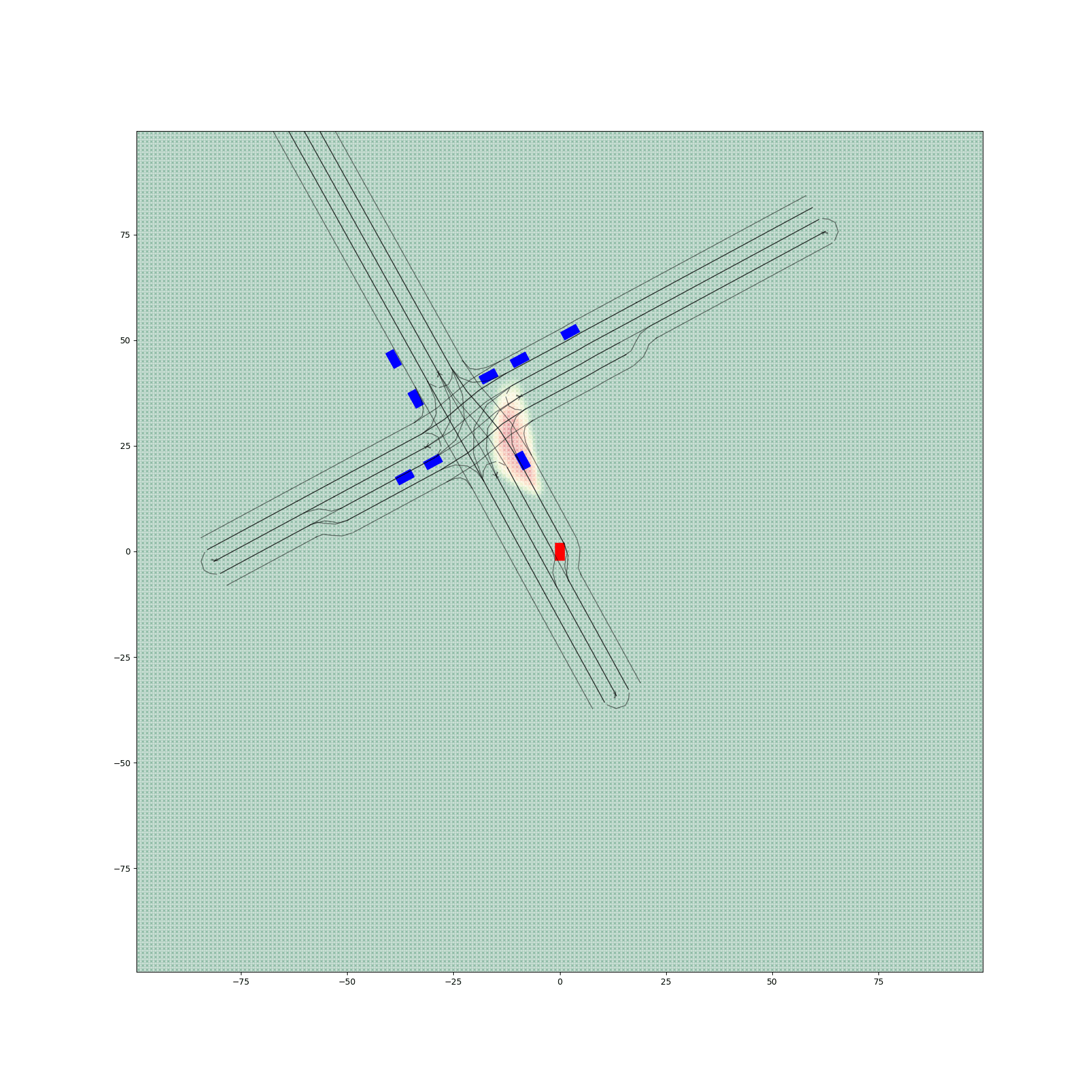}
        \caption{Encoder 1 + Decoder 2 \\ (3.5748\%)}
    \end{subfigure}
    \caption{Comparison of heatmaps generated by various encoder-decoder configurations in a scenario with abnormal heading angles caused by the discrete nature of the simulator. This case results in significantly higher probabilities of non-drivable area predictions. }
    \label{appendix_fig:6}
\end{figure*}

\end{document}